\documentclass{article}
\usepackage{perpage} 
\MakePerPage{footnote} 

\PassOptionsToPackage{numbers, compress}{natbib}

\usepackage[final]{neurips_2021}

\usepackage[utf8]{inputenc} 
\usepackage[T1]{fontenc}    
\usepackage{url}            
\usepackage{booktabs}       
\usepackage{amsfonts}       
\usepackage{nicefrac}       
\usepackage{microtype}      
\usepackage{xcolor}         
\usepackage{adjustbox}
\usepackage{mmstyle}
\usepackage{multirow}
\usepackage{tabularx, makecell}
\usepackage{subfig}
\usepackage{rotating}
\usepackage{comment}
\usepackage{tablefootnote}
\usepackage{CJKutf8}
\usepackage{ulem}

\usepackage{xcolor}
\usepackage{hyperref}
\hypersetup{colorlinks=true,
    breaklinks=true,
    urlcolor=magenta,
    linkcolor=red,
    bookmarksopen=false,
    filecolor=black,
    citecolor=green,
    linkbordercolor=red
}

\usepackage{threeparttable}

\def\etc{etc\@\xspace}

\newcommand{\datanamecls}{GV-D$_\text{c}$-36M}
\newcommand{\datanamedet}{GV-D$_\text{d}$-3M}
\newcommand{\datanameseg}{GV-D$_\text{s}$-143K}
\newcommand{\mtb}{MetaNet}
\newcommand{\mtbone}{MN-B1}
\newcommand{\mtbfour}{MN-B4}
\newcommand{\mtbseven}{MN-B7}
\newcommand{\mtbfifteen}{MN-B15}
\newcommand{\stagetwo}{Up-E stage}      

\def\amateur{{\em amateur}}
\newcommand{\expertcls}{Up-E (C)}
\newcommand{\expertdet}{Up-E (D)}
\newcommand{\expertseg}{Up-E (S)}

\title{INTERN: A New Learning Paradigm Towards General Vision}

\author{
Jing Shao$^*$
\And
Siyu Chen$^*$
\And
Yangguang Li$^*$
\And
Kun Wang$^*$
\And
Zhenfei Yin$^*$
\And
Yinan He$^*$
\And
Jianing Teng$^*$
\And
Qinghong Sun$^*$
\And
Mengya Gao$^*$
\And
Jihao Liu$^*$
\And
Gengshi Huang\thanks{Equal Contribution}
\And
Guanglu Song
\And
Yichao Wu
\And
Yuming Huang
\And
Fenggang Liu
\And
Huan Peng
\And
Shuo Qin
\And
Chengyu Wang
\And
Yujie Wang
\And
Conghui He
\And
Ding Liang
\And
Yu Liu
\And
Fengwei Yu
\And
Junjie Yan
\And
Dahua Lin
\And
Xiaogang Wang
\And
Yu Qiao\thanks{Corresponding Author: qiaoyu@pjlab.org.cn}
\And
~\\
Shanghai AI Laboratory
\quad
SenseTime Research \\
The Chinese University of Hong Kong 
\quad
Shanghai Jiao Tong University \\
~\\
\url{https://opengvlab.shlab.org.cn}
}

\begin{document}

\maketitle

\begin{abstract}

Enormous waves of technological innovations over the past several years, marked by the advances in AI technologies, are profoundly reshaping the industry and the society.
However, down the road, a key challenge awaits us, that is, our capability of meeting rapidly-growing scenario-specific demands is severely limited by the cost of acquiring the commensurate amount of training data.
This difficult situation is in essence due to limitations of the mainstream learning paradigm: we need to train a new model for each new scenario, based on a large quantity of well-annotated data and commonly from scratch.
In tackling this fundamental problem, we move beyond and develop a new learning paradigm named INTERN. By learning with supervisory signals from multiple sources in multiple stages, the model being trained will develop strong generalizability.
We evaluate our model on 26 well-known datasets that cover four categories of tasks in computer vision.
In most cases, our models, adapted with only 10\% of the training data in the target domain, outperform the counterparts trained with the full set of data, often by a significant margin.
This is an important step towards a promising prospect where such a model with general vision capability can dramatically reduce our reliance on data, thus expediting the adoption of AI technologies.
Furthermore, revolving around our new paradigm, we also introduce a new data system, a new architecture, and a new benchmark, which, together, form a general vision ecosystem to support its future development in an open and inclusive manner.

\end{abstract}

\section{Introduction}
\label{sec:introduction}

Current state-of-the-art AI models typically over-specialize on a single task, despite remarkable progress in recent years. The consequence is that we develop thousands of models for thousands of tasks or circumstances individually. Each new task requires collecting and annotating a large amount of data, and consumes a giant scale of computational resources. 
From~\cite{bommasani2021opportunities,brown2020language}, this presents itself as a significant hurdle in front of AI researches and applications, considering thousands of long-tailed tasks in industries.
Alternatively, the artificial general intelligence approach, taking "general intelligence" as a fundamentally distinct property~\cite{goertzel2014artificial}, focuses directly on the \emph{generality}, \emph{adaptability}, and \emph{flexibility} of AI models.

Vision and language are two indispensable modalities for artificial general intelligence. For language, impressive progress has been achieved towards general language model (GLM).
Recent advances of large-scale pretrained language models such as BERT~\cite{devlin2018bert}, T5~\cite{raffel2019exploring} and GPT-3~\cite{brown2020language} have shown potential in developing GLMs that substantially benefit a wide range of language-related downstream tasks by allowing economical task-specific adaptations.
Moreover, with the advent of task-agnostic training objectives~\cite{yang2019xlnet,devlin2018bert}, performance gains from pretraining can be steadily improved by scaling up web-crawled data and the model capacity together with computational budgets.

\begin{figure*}[t]
  \centering
   \includegraphics[width=1.0\linewidth]{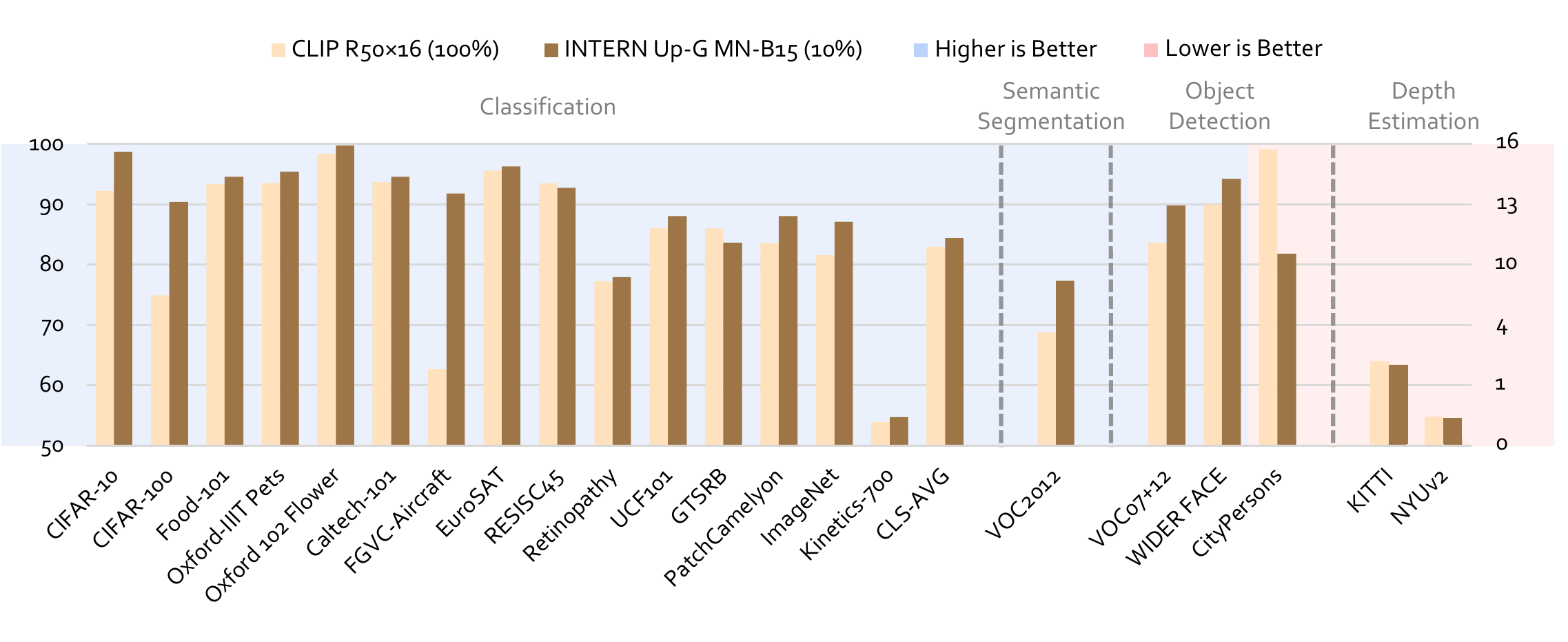}
   \vspace{-30pt}
    \caption{\textbf{Comparison of transfer learning performance on diverse tasks.} Our largest pretrained model, Up-G MN-B15, with 90\% fewer downstream data, surpasses the best publicly available pretrained model (CLIP-R50$\times$16) on most tasks. All results are obtained with backbone parameters fixed during downstream training. Note that the last three datasets in the pink background use the $y$-axis on the right, and the lower bar means the better performance.}
    \label{fig:leaderboard}
\end{figure*}

The success of GLMs has inspired new directions for general vision model (GVM) learning.
Pioneers working on large-scale supervised~\cite{zhai2021scaling,riquelme2021scaling,dai2021coatnet,JFT300M,deng2009imagenet}, self-supervised~\cite{chen2021exploring,grill2020bootstrap,chen2020simple,chen2021empirical,bao2021beit,caron2021emerging}, and cross-modal~\cite{huo2021wenlan,CLIP} pretraining have shown certain generality on a limited scope of downstream vision tasks. Nevertheless, it is still challenging to design reliable approaches towards GVMs. 
Most previous works mainly utilize one source of supervisory signal, \eg,~ViT-G/14~\cite{zhai2021scaling} uses categorical supervision,  SEER~\cite{goyal2021self} applies contrastive learning between different augmentation views, and CLIP~\cite{CLIP} makes use of paired language descriptions.
Pretraining monotonically with an individual supervision is able to produce models performing well in selected scenarios, but cannot offer sufficient competence if we aim at obtaining a "true" GVM that is \emph{generalizable} to a vast set of downstream tasks, even unseen ones.
To achieve generality with respect to diverse vision tasks, it is favorable to learn abundant information from miscellaneous types of supervisory signals, including image-level categories, bounding boxes, pixel-wise labels, quantities, as well as natural language.

\begin{figure*}[t]
  \centering
   \includegraphics[width=1.0\linewidth]{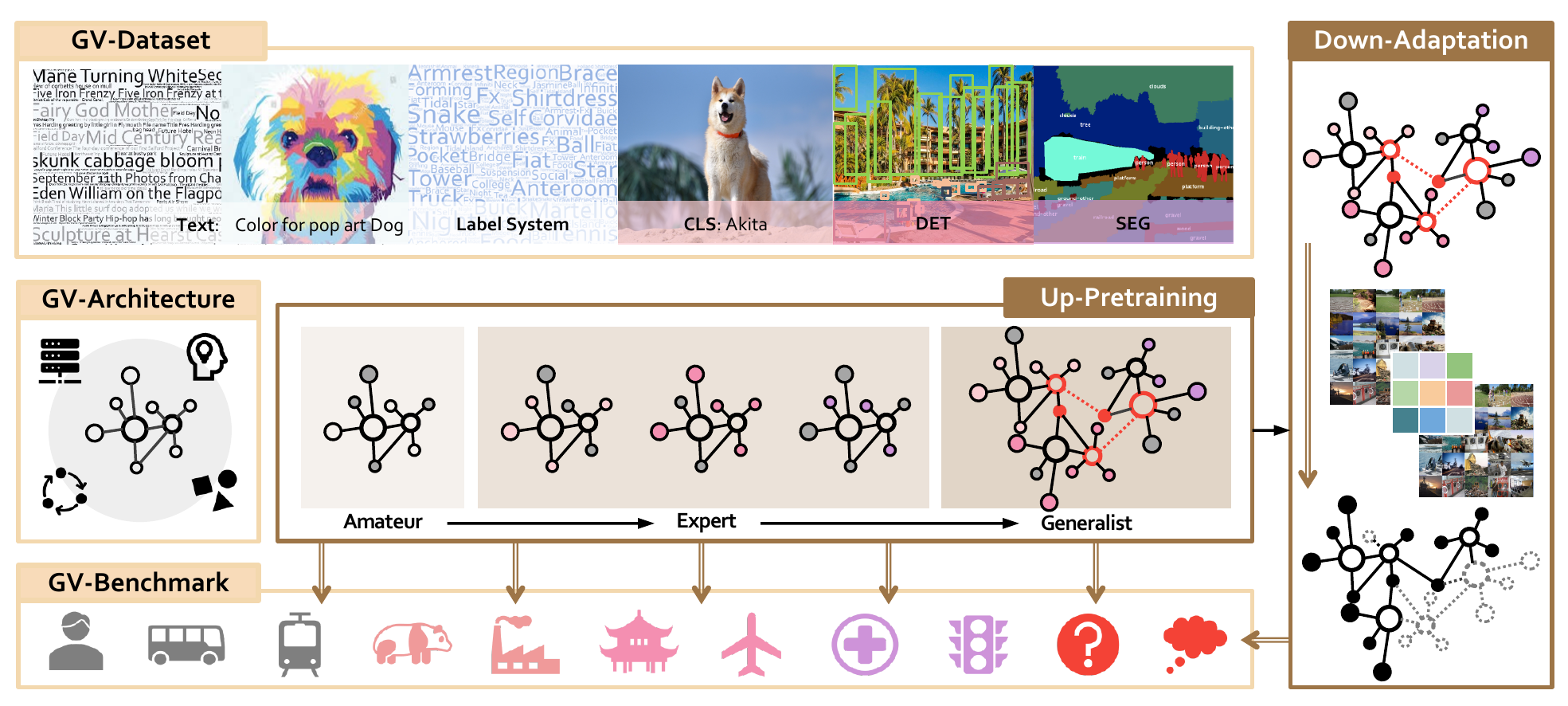}
    \caption{\textbf{Overview of INTERN.}
    A complete flow of learning and evaluating a general vision model consists of three fundamental bases (\ie~GV-Dataset, GV-Architecture, and GV-Benchmark), a three-stage upstream pretraining scheme (\ie~Amateur, Expert, and Generalist), and a downstream adaptation algorithm that transfers the up-pretrained models to various downstream tasks in the benchmark. It shows that a general model (\eg~Generalist) with a gradual-learning process behaves stronger generalizability even with unseen tasks (shown in a red question mark).
    }
   \label{fig:overview}
\end{figure*}

In this work, we propose a new learning paradigm named INTERN, further pushing forward to successful general vision modeling. 
Specifically, INTERN introduces a \textit{continuous learning} scheme (see Fig.~\ref{fig:overview}), including a highly extensible upstream pretraining pipeline leveraging large-scale data and various supervisory signals, as well as a flexible downstream adaptation towards diversified tasks.

For an analogy of the upstream pretraining procedure, one may look no further than a typical learning process of an "intern" in real life, which can be roughly divided into the three subsequent stages based on the level of expertise:
\begin{itemize}
\vspace{-0.1cm}\item[I:] An \emph{amateur} with fundamental skill sets who can superficially address encountered problems;
\vspace{-0.1cm}\item[II:] An \emph{expert} who has additionally mastered one particular task with careful supervision;
\vspace{-0.1cm}\item[III:] A \emph{generalist} who is knowledgeable about all known tasks, and adapts fast to unseen tasks.
\end{itemize}
\vspace{-0.1cm}
We demonstrate that it is beneficial to imitate this amateur-to-generalist learning process of an ``intern'' to ease the non-trivial training of the desired GVM.
The resulting multi-stage pretraining scheme not only efficiently absorbs knowledge from broad sources of supervisions, but also is easily scalable with the presence of more data or tasks. 
Based on this pipeline, our final generalists prove to possess sufficient generality towards a wide range of downstream tasks, and outperform (see Fig.~\ref{fig:leaderboard}) the previous state-of-the-art (CLIP~\cite{CLIP}) while only using an order of magnitude fewer (10\%) downstream data.

In addition to upstream pretraining, we also introduce a downstream adaptation step.
The key challenge is to form downstream task-specific models while largely preserving the merits of an upstream GVM.
We show that designing a proxy for flexible knowledge transfer onto various downstream tasks is a promising attempt towards this goal.
Moreover, establishing a comprehensive benchmark is essential to propel GVM development. 
A GVM is expected to not only generalize to different tasks but also have lower requirements on downstream data, achieving few-shot adaptation.
To this end, our general vision benchmark in INTERN contains a wide range of common downstream tasks, and systematically examines important factors relating to GVMs, especially generality and data efficiency.

\subsection{Overview of INTERN}

As shown in Fig.~\ref{fig:overview}, INTERN consists of seven key components. 
Three of them serve as fundamental bases: General Vision Dataset is constructed as the database for the upstream step-wise learning process of INTERN. General Vision Architecture is the backbone of INTERN models. General Vision Benchmark consists of a broad range of downstream datasets and evaluation settings to assess the generalization ability of vision models thoroughly.
The other four refer to our upstream pretraining scheme with three stages (\ie~Upstream-Amateur, Upstream-Expert, and Upstream-Generalist) and Downstream-Adaptation, which provides a refined solution to adapting learned upstream general models towards various types of downstream tasks.
\begin{itemize}
\vspace{-0.2cm}\item \textbf{General Vision Data (GV-D)} 
is a super-scale collection of vision datasets with 10 billion samples and various supervisory signals. It presents an extensive label system with 115K visual concepts, covering numerous realms in nature and almost all labels currently studied in computer vision. 
Guided by this comprehensive label system, we conduct four high-quality general vision datasets actively and continuously. Precisely, GV-D consists of GV-D-10B with multi-modal data, as well as GV-D$_\text{c}$-36M, GV-D$_\text{d}$-3M and GV-D$_\text{s}$-143K with classification, detection, and segmentation annotations, respectively.

\item \textbf{General Vision Architecture (GV-A)} introduces a set of network architectures with higher modeling capacity, which is constructed from a unified search space with both convolution and transformer operators. 
We name this automatically assembled and high-performing vision network family as \textit{MetaNet}.
\item \textbf{General Vision Benchmark (GV-B)} collects 26 downstream tasks consisting of 4 task types, on which models produced by our INTERN paradigm are benchmarked along with publicly released pretrained models for comparison. 
In addition, GV-B introduces a \textit{percentage-shot} setting where the amount of training data of downstream tasks is shrunk by only taking a portion of the full dataset, such as 10$\%$, 20$\%$, \etc. Compared to traditional few-shot settings, our \textit{percentage-shot} setup can well preserve properties like long-tailed distribution of the original dataset and alleviate the sensitivity to sample selection.
\item \textbf{Upstream-Amateur (Up-A)}
is a multi-modal pretraining stage for acquiring the amateur model, which simultaneously uses rich supervisory signals from image-text, image-image, and text-text pairs to train task-agnostic models serving as initialization for the next stage.
\item \textbf{Upstream-Expert (Up-E)} is the following stage in our pretraining scheme for achieving the expert model, which accumulates specialized knowledge with multi-source supervisions within one of the task types. Each expert only pays attention to its own specialty without interfering with the learning of others.
\item \textbf{Upstream-Generalist (Up-G)} is a combinational pretraining stage resulting in the generalist model, which integrates knowledge of experts and produces the final form of general representation capable of handling any known or unknown task.
\item \textbf{Downstream-Adaptation (Down-A)} introduces a transfer learning scheme aimed at enhancing knowledge transfer onto various downstream task types, which can be applied to any upstream pretrained model. It effectively improves the results of adaptation, especially in the low-data regime.
\end{itemize} 

\vspace{-0.2cm}

\section{Core Results}
\label{sec:core_results}

We first highlight the high performance of models learned with our multi-stage pretraining scheme at downstream transfer learning, especially in the data-efficient (10\%) setting.
We then evaluate the generality of our paradigm and its extensibility in the face of new tasks.
Experiments are conducted on ResNet~\cite{he2016deep} and the proposed MetaNet. MetaNet-B4 (MN-B4) shares a similar parameter complexity with ResNet-50 (R50), and MetaNet-B15 (MN-B15) is a considerably larger model with $\sim$1B parameters.
Transfer learning performances of these models are assessed on GV-B. During the evaluation, we fix pretrained backbone weights and only fine-tune the output head. This setting is previously known as \textit{linear probe} for classification tasks, and we extend this term to other tasks in our benchmark.

\subsection{Superior Transfer Learning Performance with Only 10\% Training Data}

\label{sec:superior_transfer}

Our goal is to train a general vision model capable of effectively transferring to various downstream tasks with significantly fewer data and annotations.
As displayed in Tab.~\ref{tab:whole_results}, we assess the transferability of our proposed models on a wide range of tasks, covering general and fine-grained classification, object detection, semantic segmentation, and depth estimation.
Both low-data (10\%) and full-data (100\%) settings are considered for appraising data efficiency.
We also make comparisons with a list of popular large-scale pretraining approaches on our benchmark, including supervised~\cite{bit,IN-1B}, self-supervised~\cite{chen2020improved,caron2020unsupervised,caron2018deep,detco} and cross-modal~\cite{CLIP} approaches with a variety of network architectures~\cite{he2016deep,xie2017aggregated,dosovitskiy2020image}.

From Tab.~\ref{tab:key_results}, we observe that generalist models from our multi-stage pretraining achieve state-of-the-art results on most tasks even with only 10\% downstream training data.
Take ResNet-50~\cite{he2016deep} as an example, in the low-data (10\%) regime, Up-G's performances surpass those of the ImageNet~\cite{russakovsky2015imagenet} pretrained ResNet-50 by large margins, with +11.5\% average accuracy on the classification suite, +18.2\% AP on VOC detection, and +8.2\% mIoU on semantic segmentation.
It is notable that while the pretrained ResNet-50 from CLIP~\cite{CLIP} shows a gain of +4.1\% on classification, its performance degrades on VOC detection and segmentation tasks. The results of ResNet-50 checkpoints from MoCo v2~\cite{chen2020improved} and SwAV~\cite{caron2020unsupervised} are roughly similar, both showing unsteady performances across tasks compared to the ImageNet baseline. 
Therefore, our approach is the only pretraining scheme that leads to superior transfer capabilities on all evaluated tasks.
In addition, it is worth mentioning that our ResNet-50 in the 10\% setting also steadily outperforms the ImageNet pretrained one with 100\% data in terms of all metrics. 
This further demonstrates the great downstream data efficiency of our models.

For our proposed MetaNet, MN-B4 consistently outperforms its similar-complexity rival, ResNet-50. This observation also reflects INTERN's good compatibility with different types of backbone networks.
Furthermore, our largest generalist model, MN-B15, achieves state-of-the-art performance on our GV-B benchmark in most dimensions, surpassing the best publicly available pretrained model, CLIP-R50$\times$16~\cite{CLIP}. It is also worth noting that compared to CLIP-R50$\times$16~\cite{CLIP} in the low-data regime, our approach relatively decreases the mean error rate by 40.2\% on the classification suite, 52.8\% on VOC detection, 45.0\% on CityPersons (R), 44.2\% on WIDER FACE (M), 34.8\% on VOC segmentation, 6.9\% on KITTI and 11.9\% on NYUv2, respectively.

\begin{table*}[!t]
\centering
\begin{small}
\resizebox{\textwidth}{!}{
\begin{tabular}{l|c|c|ccc|c|cc}
\toprule  
Model           &  Data Setting    & CLS-AVG $\uparrow$    & VOC-DET $\uparrow$ & CP (R) $\downarrow$ & WF (M) $\uparrow$ & VOC-SEG $\uparrow$  &KITTI $\downarrow$ & NYUv2 $\downarrow$ \\
\midrule
\multirow{2}{*}{ImageNet~\cite{russakovsky2015imagenet} R50}& 10\% & 62.8      & 69.5      & 29.6      & 84.2      & 58.0      & 3.26  & 0.48  \\
                        & 100\%     & 73.0      & 79.5      & 22.7      & 86.8      & 66.0      & 3.09  & 0.43  \\
\midrule
MoCo v2~\cite{chen2020improved} R50             & 10\%      & 61.1      & 70.2      & 23.2      & 86.3      & 60.1      & 3.13  & 0.46  \\
SwAV~\cite{caron2020unsupervised} R50                & 10\%      & 64.3      & 69.2      & 23.2      & 86.4      & 56.8      & 2.95  & 0.46  \\
\midrule
CLIP~\cite{CLIP} R50                & 10\%      & 66.9      & 68.6      & 21.3      & 87.1      & 55.8      & 3.15  & 0.48  \\
\multirow{2}{*}{CLIP~\cite{CLIP} R50$\times$16}      & 10\%      & 75.0      & 78.4      & 19.1      & 89.6      & 65.2      & 2.91  & 0.42  \\
                        & 100\%     & 82.9      & 83.6      & 16.2      & 89.9      & 68.7      & 2.83  & 0.39  \\
\midrule
Up-G R50                & 10\%      & 74.3      & 87.7      & 14.7      & 92.2      & 66.2      & 2.84  & 0.39  \\
Up-G MN-B4              & 10\%      & 78.6      & 89.1      & 12.0      & 92.8      & 71.4      & 2.94  & 0.40  \\
Up-G MN-B15             & 10\%      & \textbf{84.4}      &\textbf{89.8}      & \textbf{10.5}      & \textbf{94.2}     & \textbf{77.3}      & \textbf{2.71}  & \textbf{0.37}  \\
\bottomrule
\end{tabular}
}
\caption{\textbf{Linear probe performance of selected pretrained models.} CLS-AVG denotes average accuracy on 20 classification datasets, CP (R) denotes MR$^{-2}$ of the \textit{reasonable} setup in CityPersons and WF (M) denotes AP50 of the \textit{medium} subset in WIDER FACE. Our Up-G R50 outperforms the ImageNet pretrained version on classification, object detection, semantic segmentation and depth estimation tasks while trained with 90\% fewer downstream data.
}
\label{tab:key_results}
\end{small}
\end{table*}

\begin{table*}[t]
\centering

\begin{small}
{
  \begin{tabular}{l|c|cc}
  \toprule
    Pretrain        & Data Setting & CLS-AVG $\uparrow$            & VOC-DET $\uparrow$  \\ 
    \midrule
    ImageNet        & $100\%$       & 73.0                          & 79.5   \\
    \midrule
    Up-E (C)        & $10\%$        & {\color{red}{73.7}}           & 72.2   \\   
    Up-E (D)        & $10\%$        & 53.9                          & {\color{red}{87.7}}    \\ 
    \midrule
    Up-G (C-D)      & $10\%$        & {\color{red}{\textbf{74.3}}}  & {\color{red}{\textbf{87.7}}}    \\
  \bottomrule
  \end{tabular}
  }
  \caption{\textbf{Effectiveness of diverse knowledge integration.} Our classification and detection expert models, denoted by Up-E (C) and Up-E (D) respectively, with only $10\%$ training data, both surpass the ImageNet supervised baseline with full data on their corresponding tasks. Transferability of these two experts are successfully preserved by our generalist, Up-G (C-D), achieving $+0.6\%$ accuracy on classification. The {\color{red}{red}} color denotes that the model has been pretrained on the corresponding task.}
  \label{tab:consolidation}
  \end{small}

\end{table*}

\begin{table*}[t]
  \begin{small}
  \centering
  \begin{tabular}{l|ccc}
  \toprule
    Pretrain               & CLS-AVG $\uparrow$                 & VOC-DET $\uparrow$            & VOC-SEG $\uparrow$ \\
    \midrule
    Up-E (C)              & {\color{red}{73.7}}                 & 72.2                          & 57.7   \\
    Up-E (D)              & 53.9                                & {\color{red}{87.7}}           & 62.3   \\
    Up-E (S)              & 47.5                                & 75.0                          & {\color{red}{71.9}}  \\
    \midrule
    Up-G (C-D)            & {\color{red}{74.3}}                 & {\color{red}{87.7}}           & 66.2  \\
    Up-G (C-D-S)          & {\color{red}{\textbf{{74.3}}}}      & {\color{red}{\textbf{87.7}}}  & {\color{red}{\textbf{73.7}}}   \\
  \bottomrule
  \end{tabular}
  \caption{\textbf{Extensibility of generalist.} Based on Up-G (C-D), Up-G (C-D-S) is easily obtained by additionally linking to Up-E (S). With the new expert added, original performances are not harmed.}
  \label{tab:extensibility}
  \end{small}
\end{table*}

\begin{table*}[htb]
  \begin{small}
  \centering
  \begin{tabular}{l|c|cc}
  \toprule
    Pretrain                  & Data Setting           & VOC-SEG $\uparrow$                  & KITTI $\downarrow$  \\
    \midrule
    ImageNet                  & 100\%                   & 66.0                                & 3.09  \\
    \midrule
    Up-E (C)                  & 10\%                    & 57.7                                & 3.21 \\
    Up-E (D)                  & 10\%                    & 62.3                                & 3.09 \\
    \midrule
    Up-G (C-D)                & 10\%                    & 66.2                                & 2.84  \\
    Up-G (C-D-S)              & 10\%                    & {\color{red}{\textbf{{73.7}}}}      & \textbf{2.80}  \\
  \bottomrule
  \end{tabular}
  \caption{\textbf{Generalizability to unseen tasks.} Up-G (C-D) outperforms the ImageNet pretrained ResNet-50 on VOC segmentation and KITTI depth estimation with $90\%$ fewer training data. Up-G (C-D-S) also performs well on the depth estimation task. Up-E (C) or Up-E (D) alone fails to match the performances of the generalist Up-G (C-D).}
  \label{tab:generalization}
  \end{small}
\end{table*}

\subsection{Easy Extensibility and Great Generalizability}
\label{subsec:core_results_generalizability}

INTERN is a general-purpose pretraining paradigm suitable for any task or model architecture. In this section, we select the ResNet-50 backbone to demonstrate the extensibility and generalizability of our approach for a fair comparison with previous works.

\emph{Effective integration of diverse knowledge.} The role of Up-E, \ie~expert training stage, is to learn representation for each specialized task type such as image classification, object detection, and semantic segmentation. This design naturally mitigates the learning difficulty caused by task conflicts~\cite{crawshaw2020multi} in multi-task learning setups.
The representation learned for each task type performs better than simple ImageNet pretraining when tested on other tasks of the same type~\cite{mensink2021factors} due to strengthened knowledge specifically for that task type.
Benefiting from GV-D's large-scale data with rich multi-domain information, Up-E's quality is improved to a greater extent and it becomes worthy of its name, expert. For example, in Tab.~\ref{tab:consolidation}, downstream performances on corresponding tasks with only $10\%$ training data of the experts for classification and detection (referred to as Up-E (C) and Up-E (D) respectively) beat those of the ImageNet baseline trained with full data, showing excellent few-shot capabilities. Next, we consider how to obtain a general representation across different task types. In Up-G, we introduce a simple yet effective \emph{knowledge transfer module} for gluing the experts together. This module lies between any two experts, learning complementary information for the target expert from the source expert.
Such a knowledge integration mechanism over all experts results in a universal representation instead of specific ones. In Tab.~\ref{tab:consolidation}, Up-G (C-D) successfully preserves the capabilities of both the classification and detection experts, and we observe an extra gain of $+0.6\%$ on average classification accuracy. The results imply that our pretraining approach is able to effectively consolidate diverse knowledge from different task types.

\emph{Easily extensible with more experts.} To build a robust pretraining system, we always keep extensibility in mind. 
Our Up-G is a simple multi-task learning approach where experts of new tasks can be easily added.
We already show that Up-G (C-D) matches or excels its two experts in terms of transferability on both tasks.
When introduced with a semantic segmentation expert Up-E (S), we simply extend Up-G (C-D) 
to Up-G (C-D-S)\textcolor{red}{,} which further acquires the few-shot ability from the new expert without dropping performances on classification and detection. Moreover, we observe a gain of $+1.8\%$ on segmentation compared to Up-E (S), showing again the value of multiple experts in our pretraining system.

\emph{Robust representation for generalization to unseen tasks.} As a generalist, it is required to be proficient at all known tasks and has a fast adaptation ability to unseen tasks with few samples.
We show that pretrained models based on our pipeline meet the goal of possessing sufficient generality towards unknown downstream tasks.
As displayed in Tab.~\ref{tab:generalization}, with only $10\%$ training data, our Up-G (C-D) achieves $66.2\%$ mIoU on VOC segmentation and $2.84$ RMSE on KITTI depth estimation. Both metrics on the two unseen tasks are better than those of the ImageNet supervised model with $100\%$ data. As for Up-G (C-D-S), besides the significant gain on segmentation compared to Up-G (C-D), we see slightly stronger performance on the unseen depth estimation as well. This generalizability is not achieved by any of the single expert models. Specifically, either Up-E (C) or Up-E (D) alone considerably lags behind Up-G (C-D) in terms of VOC segmentation and KITTI depth estimation.
We attribute this improvement achieved by our generalists to the diverse supervisory signals applied during our multi-stage pretraining. Our pipeline effectively exploits information embedded in multiple tasks, leading to a more robust and generalizable model. Our experiments may further shed light on learning complementary information from a multi-task setting.

\subsection{Factors Contributing to General Vision Intelligence}

\begin{table*}[t]
\begin{small}
\centering

\begin{tabular}{lllll}
\toprule
Datasets & Concepts & Images & Labels        & Open Source \\ \midrule
YFCC-100M~\cite{thomee2016yfcc100m}     & -            & 99M &  99M Texts  & Yes        \\
WIT~\cite{CLIP}           & 500K Queries & 400M & 400M Texts & No        \\
ALIGN~\cite{jia2021scaling}           & -     & 1.8B  & 1.8B Texts    & No        \\
\textbf{GV-D-10B}    & \textbf{1.65M Queries} & \textbf{10B} & \textbf{10B Texts} & Partially \\
\hline
ImageNet-21K~\cite{deng2009imagenet}   & 22K Categories & 14M   & 14M Image-Level Labels             & Yes        \\
IG-1B~\cite{IN-1B}         & 17K Queries & 1B & 1B Hashtags & No        \\
JFT-3B~\cite{zhai2021scaling}      & 30K Categories & \textbf{3B} & 3B Noisy Image-Level Labels & No        \\
\textbf{GV-D$_\text{c}$-36M} & \textbf{115K Categories} & 36M & \textbf{36M Image-Level Labels} & Yes        \\ 
\hline
COCO~\cite{lin2014microsoft}    & 80 Categories & 118K  & 1M Bounding Boxes & Yes        \\
Object365~\cite{shao2019objects365}  & 365 Categories & 609K   & 10M Bounding Boxes & Yes        \\
Open Images~\cite{kuznetsova2020open}    & 600 Categories & 2M  & 15M Bounding Boxes & Yes        \\
\textbf{GV-D$_\text{d}$-3M}& \textbf{809 Categories} & \textbf{3M} & \textbf{25M Bounding Boxes} & Yes        \\ 
\hline
COCO-Stuff~\cite{caesar2018coco}& 182 Categories & 118K & Segmentation Masks             & Yes        \\
\textbf{GV-D$_\text{s}$-143K} & \textbf{334 Categories} & \textbf{143K} & Segmentation Masks & Yes        \\ 
\bottomrule            
\end{tabular}
\caption{\textbf{Summary of GV-D and other large-scale datasets for visual pretraining.} A large-scale database is a fundamental component of general vision pretraining. YFCC-100M, ImageNet-21K, COCO, Object365, OpenImages are instances of commonly used public datasets, while IG-1B, WIT, JFT-3B are proprietary ones that cannot be accessed by the community. 
We construct a novel data system \textbf{GV-D} with four subsets: 1) \textbf{GV-D-10B} consisting of 10 billion image-text pairs collected with 1.65 million queries; 2) \textbf{GV-D$_\text{c}$-36M} contains 36 million images with classification labels from our label system of 115K categories. Although GV-D$_\text{c}$-36M has fewer images than JFT-3B, it has the most clean labels. 3) \textbf{GV-D$_\text{d}$-3M} is composed of 3 million images with 25 million bounding boxes of 809 categories; 4) \textbf{GV-D$_\text{s}$-143K} with 143 thousand images and corresponding semantic segmentation masks of 334 categories.}
\label{tab:dataset_summary}
\end{small}
\end{table*}

\begin{table*}[h]
  \begin{small}
\centering
  \begin{tabular}{l|c|ccc}
  \toprule  
  Model    & Data Setting                      & Up-A  & Up-E  & Up-G  \\
  \midrule
  R50       & \multirow{2}{*}{10\%}             & 70.9  & 73.7  & 74.3  \\
  MN-B15          &                                   & 80.4  & 84.2  & 84.4  \\
  \bottomrule
  \end{tabular}

  \caption{\textbf{Average downstream classification accuracies of different upstream stages.} Classification precision monotonically increases when moving to later stages. Results are consistent at different model scales. \label{tab:mtb15 result of stage1-3}}
  \end{small}
\end{table*}

First, \emph{scaling up dataset size, broadening domains, and diversifying supervisory signals matter.}
In INTERN, GV-D acts as a foundational database for models to comprehensively ``study'' in the pretraining step.
Most existing datasets are constrained within a narrow computer vision task type, \eg,~ image classification, object detection, \etc. 
We argue that this is insufficient for general vision pretraining which requires wide coverage over various domains. Although some multi-modal or supervised pretraining approaches have used billion-scale data, the performances of resulting models on some structured tasks like object detection and segmentation are unsatisfactory. This indicates that simply increasing the size of a dataset with one single type of supervisory signals, such as paired texts in those multi-modal datasets, is still not good enough because the knowledge required by a vast majority of tasks may be missing.
In GV-D, as listed in Tab.~\ref{tab:dataset_summary}, besides multi-modal data, we include three additional subsets corresponding to three common vision tasks, each of which has the largest scale compared to publicly available datasets for the same type of task. Our label system with 115K categories, which is fully utilized in the classification set GV-D$_\text{c}$-36M, is more than four times larger than that of the ImageNet-21K~\cite{deng2009imagenet} dataset, covering a much wider range of hierarchically organized visual concepts. It significantly boosts the performance of more fine-grained tasks.
We also include billion-level image-text pairs in GV-D-10B, which further expands supervisory signals extensively and reduce negative effects associated with noisy data.
In summary, instead of being biased towards one supervision, GV-D provides rich data across multiple task types to achieve pretraining for general vision intelligence. Clear gains are observed in comprehensive evaluation on diverse tasks, which are discussed in Sec.~\ref{sec:superior_transfer}.

Second, \emph{multi-stage training brings consistent gains.}
As shown in Tab.~\ref{tab:mtb15 result of stage1-3}, INTERN learns fundamental skills with the \textit{amateur} pretraining stage. Then, after \textit{expert} pretraining with more specific and specialized knowledge, INTERN masters image classification with an average improvement of $+2.8\%$ for ResNet-50 and $+3.8\%$ for MN-B15 on 20 relevant datasets compared to the \textit{amateur} stage. Finally, based on the multiple expert models from the \textit{expert} stage, the \textit{generalist} stage mines a universal set of skills, obtaining further gains of $+0.6\%$ and $+0.2\%$. Meanwhile, the \textit{generalist} model possesses sufficient generality towards more, possibly unseen, tasks and circumstances, which is demonstrated in Sec.~\ref{subsec:core_results_generalizability}. The consistent gains at successive stages show that our \textit{amateur}-\textit{expert}-\textit{generalist} training pipeline effectively learns new knowledge and improves upon the previous stage. These results also imply that our continuous learning paradigm extends the ability of any single-stage pretraining, validating the design choice of combining them appropriately.

\begin{table*}
  \begin{adjustbox}{angle=90}
       \large
       \centering
       \resizebox{1.18\linewidth}{!}
       {
        \begin{tabular}{ll|lllllllllllllllllllll|lllll|l|l|ll}
          \hline
          \multicolumn{2}{l|}{\textbf{}}         & \multicolumn{21}{c|}{CLS}                                                                                                                                                                           & \multicolumn{6}{c|}{DET}                                                          & \multicolumn{1}{c|}{SEG} & \multicolumn{2}{c}{DEP} \\ \hline
          \multicolumn{2}{l|}{\rotatebox{-90}{10\% data}}      & \rotatebox{-90}{CIFAR-10} & \rotatebox{-90}{CIFAR-100} & \rotatebox{-90}{Food-101} & \rotatebox{-90}{Oxford-IIIT Pets} & \rotatebox{-90}{Oxford 102 Flower} & \rotatebox{-90}{SUN397} & \rotatebox{-90}{Stanford Cars} & \rotatebox{-90}{DTD}  & \rotatebox{-90}{Caltech-101} & \rotatebox{-90}{FGVC-Aircraft} & \rotatebox{-90}{SVHN} & \rotatebox{-90}{EuroSAT} & \rotatebox{-90}{RESISC45} & \rotatebox{-90}{Retinopathy} & \rotatebox{-90}{FER2013} & \rotatebox{-90}{UCF101} & \rotatebox{-90}{GTSRB} & \rotatebox{-90}{PatchCamelyon} & \rotatebox{-90}{ImageNet} & \rotatebox{-90}{Kinetics-700} & \rotatebox{-90}{CLS AVG $\uparrow$} & \rotatebox{-90}{VOC07+12 $\uparrow$}  & \multicolumn{3}{c}{\rotatebox{-90}{WIDER FACE $\uparrow$}}      & \multicolumn{2}{c|}{\rotatebox{-90}{CityPersons $\downarrow$}} & \rotatebox{-90}{VOC2012 $\uparrow$}      & \rotatebox{-90}{KITTI $\downarrow$}       & \rotatebox{-90}{NYUv2 $\downarrow$}      \\ \hline
          \multirow{3}{*}{Up-A}        & R50       & 92.4    & 73.5     & 75.8    & 85.7 & 94.6    & 57.9   & 52.7 & 65.0 & 88.5       & 28.7     & 61.4 & 93.8    & 82.9     & 73.8        & 55.0    & 71.1   & 75.1  & 82.9 & 71.9       & 35.2        & 70.9    & 76.3      & \multicolumn{3}{l}{90.3/88.3/70.7} & \multicolumn{2}{l|}{24.6/59.0}   & 62.5                     & 3.18       & 0.46       \\
                                     & MN-B4      & 96.1    & 82.9     & 84.3    & 89.8 & 98.3    & 66.0   & 61.4 & 66.8 & 92.8       & 32.5     & 60.4 & 92.7    & 85.8     & 75.6        & 56.5    & 76.9   & 74.4  & 84.3 & 77.2       & 39.4        & 74.7    & 74.9      & \multicolumn{3}{l}{89.3/87.6/71.4} & \multicolumn{2}{l|}{26.5/61.8}   & 65.7                     & 3.57       & 0.48       \\
                                     & MN-B15     &98.2   & 87.8     & 93.9    & 92.8 & 99.6    & 72.3   & 59.4 & 70.0 & 93.8       & 64.8     & 58.6 & 95.3    & 91.9     & 77.9        & 62.8    & 85.4   & 76.2  & 87.8 & 86.0       & 52.9        & 80.4    & 78.4      & \multicolumn{3}{l}{93.6/91.8/77.2} & \multicolumn{2}{l|}{17.7/49.5}   & 60.7                     & 2.42       & 0.38       \\ \hline
          \multirow{8}{*}{Up-E}        & C-R50   & 91.9    & 71.2     & 80.7    & 88.8 & 94.0    & 57.4   & 67.9 & 62.7 & 85.5       & 73.9     & 57.6 & 93.7    & 83.6     & 75.4        & 54.1    & 69.6   & 73.9  & 85.7 & 72.5       & 34.6        & 73.7    & 72.2      & \multicolumn{3}{l}{89.7/87.6/68.1} & \multicolumn{2}{l|}{22.4/58.3}   & 57.7                     & 3.21       & 0.50       \\
                                     & D-R50   & 86.4    & 57.3     & 53.9    & 31.4 & 44.0    & 39.8   & 8.6  & 44.6 & 72.5       & 15.8     & 64.2 & 89.1    & 72.8     & 73.6        & 46.6    & 57.4   & 67.5  & 81.7 & 45.0       & 25.2        & 53.9    & 87.7      & \multicolumn{3}{l}{93.8/92.0/75.5} & \multicolumn{2}{l|}{15.8/41.5}   & 62.3                     & 3.09       & 0.45       \\
                                     & S-R50   & 78.3    & 46.6     & 45.1    & 24.2 & 33.9    & 38.0   & 5.0  & 41.4 & 50.2       & 8.5      & 51.5 & 89.9    & 76.4     & 74.0        & 44.8    & 42.0   & 64.0  & 80.8 & 34.9       & 19.7        & 47.5    & 75.0      & \multicolumn{3}{l}{87.4/85.7/66.4} & \multicolumn{2}{l|}{19.6/53.3}   & 71.9                     & 3.12       & 0.45       \\
                                     & C-MN-B4  & 96.7    & 83.2     & 89.2    & 91.9 & 98.2    & 66.7   & 67.7 & 66.3 & 91.9       & 77.2     & 57.8 & 94.4    & 88.0     & 77.0        & 56.6    & 78.5   & 77.3  & 85.6 & 80.5       & 44.2        & 78.4    & 73.7      & \multicolumn{3}{l}{89.6/88.0/71.1} & \multicolumn{2}{l|}{30.3/65.0}   & 65.8                     & 3.54       & 0.46       \\
                                     & D-MN-B4  & 91.5    & 67.0     & 61.4    & 44.4 & 57.2    & 41.8   & 12.1 & 41.2 & 80.6       & 25.1     & 68.0 & 90.7    & 74.6     & 74.3        & 50.3    & 61.7   & 74.2  & 81.9 & 57.0       & 29.3        & 59.2    & 89.3      & \multicolumn{3}{l}{94.6/92.6/76.5} & \multicolumn{2}{l|}{14.0/43.8}   & 73.1                     & 3.05       & 0.40       \\
                                     & S-MN-B4  & 83.5    & 57.2     & 68.3    & 70.8 & 85.8    & 52.9   & 25.9 & 52.8 & 81.6       & 17.7     & 56.1 & 91.3    & 83.6     & 74.5        & 49.0    & 55.2   & 68.0  & 84.3 & 61.0       & 27.4        & 62.3    & 78.7      & \multicolumn{3}{l}{89.5/87.9/71.4} & \multicolumn{2}{l|}{19.4/53.0}   & 79.6                     & 3.06       & 0.41       \\
                                     & C-MN-B15 &98.7    & 90.1     & 94.7    & 95.1 & 99.7    & 75.7   & 74.9 & 73.6 & 94.4       & 91.8     & 66.7 & 96.2    & 92.8     & 77.6        & 62.3    & 87.7   & 83.3  & 87.5 & 87.2       & 54.7        & 84.2    & 80.4      & \multicolumn{3}{l}{93.2/91.4/75.7} & \multicolumn{2}{l|}{29.5/59.9}   & 70.6                     & 2.63       & 0.37       \\
                                     & D-MN-B15 & 92.2    & 67.9     & 69.0    & 33.9 & 59.5    & 45.4   & 13.8 & 46.3 & 82.0       & 26.6     & 65.4 & 90.1    & 79.1     & 76.0        & 53.2    & 63.7   & 74.4  & 83.3 & 62.2       & 33.7        & 60.9    & 89.4      & \multicolumn{3}{l}{95.8/94.4/80.1} & \multicolumn{2}{l|}{10.5/42.4}   & 77.2                     & 2.72       & 0.37       \\ \hline
          \multirow{3}{*}{Up-G}        & R50       & 92.9    & 73.7     & 81.1    & 88.9 & 94.0    & 58.6   & 68.6 & 63.0 & 86.1       & 74.0     & 57.9 & 94.4    & 84.0     & 75.7        & 54.3    & 70.8   & 74.3  & 85.9 & 72.6       & 34.8        & 74.3    & 87.7      & \multicolumn{3}{l}{93.9/92.2/77.0} & \multicolumn{2}{l|}{14.7/46.0}   & 66.2                     & 2.84       & 0.39       \\
                                     & MN-B4      & 96.7    & 83.9     & 89.2    & 92.1 & 98.2    & 66.7   & 67.7 & 66.5 & 91.9       & 77.2     & 57.8 & 94.4    & 88.0     & 77.0        & 57.1    & 79.0   & 77.7  & 86.0 & 80.5       & 44.2        & 78.6    & 89.1      & \multicolumn{3}{l}{94.9/92.8/76.5} & \multicolumn{2}{l|}{12.0/50.5}   & 72.2                     & 2.94       & 0.40       \\ 
                                     & MN-B15     & 98.7    & 90.4     & 94.5    & 95.4 & 99.7    & 74.4   & 75.4 & 74.2 & 94.5       & 91.8     & 66.7 & 96.3    & 92.7     & 77.9        & 63.1    & 88.0   & 83.6  & 88.0 & 87.1       & 54.7        & 84.4    & 89.8      & \multicolumn{3}{l}{95.9/94.2/78.8} & \multicolumn{2}{l|}{10.5/41.3}   & 77.3                     & 2.71       & 0.37       \\ \hline
          \multirow{3}{*}{ImageNet~\cite{russakovsky2015imagenet}}    & R50        & 88.6    & 65.8     & 59.3    & 88.3 & 64.6    & 43.4   & 17.9 & 56.0 & 83.5       & 23.3     & 60.8 & 93.1    & 80.1     & 71.5        & 44.6    & 57.8   & 71.5  & 83.7 & 74.3       & 28.1        & 62.8    & 69.5      & \multicolumn{3}{l}{87.8/84.2/65.7} & \multicolumn{2}{l|}{29.6/67.2}   & 58.0                     & 3.26        & 0.48        \\
                                     & R101       & 89.4    & 66.6     & 61.3    & 89.3 & 63.2    & 44.1   & 19.2 & 57.7 & 84.0       & 23.4     & 58.9 & 93.5    & 79.3     & 72.0        & 46.8    & 58.8   & 71.3  & 84.0 & 75.7       & 29.1        & 63.4    & 71.6      & \multicolumn{3}{l}{88.2/86.0/66.3} & \multicolumn{2}{l|}{30.1/68.4}   & 59.4                     & 3.24        & 0.48        \\
                                     & R152       & 89.1    & 65.9     & 60.1    & 89.3 & 63.0    & 44.2   & 19.1 & 57.8 & 83.7       & 23.4     & 57.4 & 93.1    & 79.1     & 74.5        & 45.5    & 61.5   & 73.1  & 84.4 & 76.9       & 30.2        & 63.6    & 73.0      & \multicolumn{3}{l}{90.1/88.4/71.2} & \multicolumn{2}{l|}{28.5/65.7}   & 61.0                     & 3.12        & 0.47        \\ \hline
          SwAV~\cite{caron2020unsupervised}                       & R50      & 89.7    & 65.9     & 65.3    & 76.1 & 69.4    & 47.0   & 14.5 & 62.6 & 74.8       & 25.2     & 64.8 & 95.5    & 85.2     & 76.7        & 51.8    & 61.6   & 79.1  & 87.2 & 64.7       & 29.7        & 64.3    & 69.2      & \multicolumn{3}{l}{88.7/86.4/66.9} & \multicolumn{2}{l|}{23.2/61.8}   & 56.8                     & 2.95        & 0.46        \\
          DeepClusterV2~\cite{caron2018deep}             & R50      & 89.6    & 66.1     & 66.7    & 77.7 & 70.1    & 47.3   & 13.3 & 62.0 & 74.4       & 23.6     & 65.5 & 95.4    & 86.0     & 76.5        & 51.7    & 61.8   & 79.2  & 86.8 & 66.8       & 30.3        & 64.5    & 70.0      & \multicolumn{3}{l}{88.9/86.5/66.7} & \multicolumn{2}{l|}{22.9/60.3}   & 56.8                     & 3.02        & 0.45        \\
          MoCo v2~\cite{chen2020improved}                    & R50      & 90.1    & 66.0     & 59.0    & 70.6 & 58.4    & 41.6   & 10.7 & 58.9 & 75.0       & 19.9     & 62.3 & 94.6    & 81.5     & 75.4        & 49.9    & 57.9   & 76.2  & 84.8 & 61.8       & 26.5        & 61.1    & 70.2      & \multicolumn{3}{l}{89.2/86.3/66.8} & \multicolumn{2}{l|}{23.2/58.1}   & 60.1                     & 3.13        & 0.46        \\ \hline
          \multirow{3}{*}{ViT~\cite{dosovitskiy2020image}}       & B/16      & 90.0    & 69.3     & 78.2    & 84.4 & 97.2    & 60.9   & 20.2 & 62.2 & 86.5       & 25.4     & 55.2 & 94.5    & 86.0     & 76.4        & 52.6    & 68.9   & 62.5  & 84.8 & 76.6       & 35.4        & 68.3    & -         & \multicolumn{3}{l}{-}              & \multicolumn{2}{l|}{-}            & -                        & -           & -           \\
                                     & L/16      & 96.1    & 83.2     & 85.9    & 89.4 & 98.3    & 67.9   & 27.2 & 66.6 & 93.1       & 28.5     & 59.8 & 95.4    & 87.1     & 73.6        & 56.6    & 80.6   & 72.3  & 85.1 & 82.2       & 42.6        & 73.6    & -         & \multicolumn{3}{l}{-}              & \multicolumn{2}{l|}{-}           & -                        & -           & -           \\
                                     & H/14      & 92.5    & 76.2     & 74.1    & 86.5 & 95.1    & 56.3   & 18.5 & 58.6 & 85.6       & 25.6     & 56.8 & 93.1    & 82.1     & 74.8        & 50.8    & 66.4   & 71.0  & 84.8 & 70.9       & 31.0        & 67.5    & -         & \multicolumn{3}{l}{-}              & \multicolumn{2}{l|}{-}           & -                        & -           & -           \\ \hline
          \multirow{4}{*}{CLIP~\cite{CLIP}}      & ViT-B /16 & 94.9    & 77.2     & 90.6    & 86.4 & 86.2    & 66.7   & 62.5 & 65.1 & 88.1       & 37.8     & 66.7 & 94.1    & 91.7     & 75.2        & 65.3    & 80.2   & 78.1  & 83.2 & 75.7       & 46.5        & 75.6    & -         & \multicolumn{3}{l}{-}              & \multicolumn{2}{l|}{-}           & -                        & -           & -           \\
                                     & R50      & 85.3    & 58.7     & 81.4    & 72.4 & 73.7    & 56.0   & 46.9 & 59.6 & 76.8       & 28.4     & 61.9 & 91.7    & 85.0     & 75.5        & 57.8    & 69.4   & 71.2  & 82.8 & 67.5       & 36.2        & 66.9    & 68.6      & \multicolumn{3}{l}{89.3/87.1/73.6} & \multicolumn{2}{l|}{21.3/57.3}   & 55.8                     & 3.15        & 0.48        \\
                                     & R50$\times$16   & 89.9    & 67.5     & 91.2    & 85.4 & 84.7    & 65.3   & 65.1 & 66.9 & 85.7       & 40.6     & 70.3 & 92.8    & 90.2     & 74.9        & 64.8    & 78.2   & 80.3  & 83.6 & 77.0       & 46.4        & 75.0    & 78.4      & \multicolumn{3}{l}{92.4/89.6/76.7} & \multicolumn{2}{l|}{19.1/52.6}   & 65.2                     & 2.91        & 0.42        \\
                                     & R101     & 88.3    & 63.8     & 85.4    & 80.5 & 78.3    & 60.4   & 56.3 & 62.0 & 81.8       & 31.4     & 61.1 & 91.9    & 87.4     & 75.8        & 59.7    & 74.1   & 74.0  & 82.5 & 71.0       & 39.9        & 70.3    & 71.7      & \multicolumn{3}{l}{90.4/87.0/71.0} & \multicolumn{2}{l|}{20.0/56.6}   & 60.4                     & 2.96        & 0.48        \\ \hline
          \multirow{3}{*}{BiT~\cite{bit}}       & M-R50     & 92.8    & 74.7     & 75.8    & 85.8 & 95.5    & 56.5   & 18.9 & 64.0 & 84.1       & 25.5     & 59.3 & 94.5    & 83.9     & 75.8        & 53.7    & 67.1   & 75.7  & 84.7 & 71.5       & 33.8        & 68.7    & -         & \multicolumn{3}{l}{-}              & \multicolumn{2}{l|}{-}           & -                        & -           & -           \\
                                     & S-R50     & 87.9    & 63.7     & 61.1    & 85.9 & 65.4    & 42.7   & 15.2 & 59.0 & 79.8       & 22.7     & 62.9 & 94.0    & 80.7     & 75.2        & 47.9    & 55.7   & 74.1  & 85.2 & 71.0       & 27.6        & 62.9    & -         & \multicolumn{3}{l}{-}              & \multicolumn{2}{l|}{-}           & -                        & -           & -           \\
                                     & M-R152$\times$4  & 96.5    & 83.2     & 82.3    & 87.7 & 96.1    & 63.8   & 17.3 & 65.7 & 88.0       & 19.0     & 54.0 & 94.4    & 84.0     & 76.3        & 55.7    & 77.2   & 72.7  & 82.8 & 79.6       & 39.9        & 70.8    & -         & \multicolumn{3}{l}{-}              & \multicolumn{2}{l|}{-}           & -                        & -           & -           \\ \hline
          \multirow{2}{*}{Instagram~\cite{IN-1B}} & 32$\times$8d     & 93.1    & 70.8     & 76.0    & 92.5 & 60.6    & 54.2   & 35.9 & 57.7 & 88.4       & 25.0     & 48.3 & 89.9    & 77.3     & 74.6        & 53.4    & 68.3   & 64.6  & 83.1 & 82.8       & 37.5        & 66.7    & -         & \multicolumn{3}{l}{-}              & \multicolumn{2}{l|}{-}           & -                        & -           & -           \\
                                     & 32$\times$48d    & 94.2    & 74.5     & 80.5    & 93.5 & 70.0    & 59.3   & 41.2 & 61.3 & 91.0       & 26.9     & 49.0 & 91.7    & 81.5     & 74.5        & 56.4    & 75.1   & 69.0  & 82.4 & 85.4       & 40.2        & 69.9    & -         & \multicolumn{3}{l}{-}              & \multicolumn{2}{l|}{-}           & -                        & -           & -           \\ \hline
          DetCo~\cite{detco}                      & R50      & -       & -        & -       & -    & -       & -      & -    & -    & -          & -        & -    & -       & -        & -           & -       & -      & -     & -    & -          & -           & -       & 68.0      & \multicolumn{3}{l}{89.9/85.5/66.2} & \multicolumn{2}{l|}{25.2/60.8}   & 57.4                     & 3.20        & 0.47        \\ \hline
          \multicolumn{2}{l|}{\rotatebox{-90}{100\% data}}      & \rotatebox{-90}{CIFAR-10} & \rotatebox{-90}{CIFAR-100} & \rotatebox{-90}{Food-101} & \rotatebox{-90}{Oxford-IIIT Pets} & \rotatebox{-90}{Oxford 102 Flower} & \rotatebox{-90}{SUN397} & \rotatebox{-90}{Stanford Cars} & \rotatebox{-90}{DTD}  & \rotatebox{-90}{Caltech-101} & \rotatebox{-90}{FGVC-Aircraft} & \rotatebox{-90}{SVHN} & \rotatebox{-90}{EuroSAT} & \rotatebox{-90}{RESISC45} & \rotatebox{-90}{Retinopathy} & \rotatebox{-90}{FER2013} & \rotatebox{-90}{UCF101} & \rotatebox{-90}{GTSRB} & \rotatebox{-90}{PatchCamelyon} & \rotatebox{-90}{ImageNet} & \rotatebox{-90}{Kinetics-700} & \rotatebox{-90}{CLS AVG} & \rotatebox{-90}{VOC07+12}  & \multicolumn{3}{c}{\rotatebox{-90}{WIDER FACE}}      & \multicolumn{2}{c|}{\rotatebox{-90}{CityPersons}} & \rotatebox{-90}{VOC2012}      & \rotatebox{-90}{KITTI}       & \rotatebox{-90}{NYUv2}      \\ \hline
          \multirow{1}{*}{Up-G} & MN-B15 & 99.0    & 92.5     & 95.5     & 96.4    & 99.7    & 83.8   & 93.5 & 85.4 & 97.7 & 96.2     & 72.6 & 97.9    & 96.0     & 79.2        & 67.8    & 91.6   & 88.5  & 87.6 & 88.4       & 59.2        & 88.4    &  90.7     & \multicolumn{3}{l}{96.4/94.7/80.8}             & \multicolumn{2}{l|}{10.6/41.3}             & 78.7        & 2.55              & 0.32                   \\ \hline
          \multirow{3}{*}{ImageNet}    & R50        & 91.8    & 74.5     & 71.3    & 92.4 & 90.8    & 60.5   & 49.9 & 72.3 & 90.8       & 48.5     & 67.0 & 95.8    & 88.1     & 74.9        & 54.0    & 68.1   & 80.0  & 82.5 & 74.3       & 32.4        & 73.0    & 79.5     & \multicolumn{3}{l}{89.1/86.8/68.6} & \multicolumn{2}{l|}{22.7/58.7}    & 66.0         & 3.09             & 0.43              \\
                                     & R101       & 93.0    & 77.2     & 72.7    & 92.3 & 90.4    & 60.8   & 50.1 & 71.6 & 91.9       & 47.0     & 65.9 & 95.8    & 86.8     & 76.0        & 54.7    & 69.8   & 79.2  & 83.3 & 75.8       & 33.7        & 73.4    & 82.2     & \multicolumn{3}{l}{90.1/87.1/71.0} & \multicolumn{2}{l|}{24.3/58.8}    & 68.0         & 3.27             & 0.45              \\
                                     & R152       & 93.5    & 78.0     & 73.7    & 93.0 & 89.6    & 61.6   & 52.8 & 71.9 & 92.1       & 48.4     & 64.2 & 95.8    & 87.6     & 75.0        & 54.7    & 71.4   & 78.7  & 82.9 & 77.1       & 34.8        & 73.8    & 82.7     & \multicolumn{3}{l}{90.9/88.9/72.2} & \multicolumn{2}{l|}{22.7/56.3}    & 67.8         & 3.06             & 0.43              \\ \cline{1-30}
          SwAV                       & R50      & 92.5    & 76.6     & 76.4    & 88.0 & 93.0    & 65.5   & 60.5 & 78.1 & 91.0       & 56.0     & 70.3 & 97.6    & 91.9     & 78.0        & 58.7    & 75.6   & 86.1  & 87.3 & 66.9       & 32.6        & 76.1    & 79.3     & \multicolumn{3}{l}{91.4/89.3/71.9} & \multicolumn{2}{l|}{19.3/53.5}    & 65.7         & 3.04             & 0.42              \\
          DeepClusterV2             & R50      & 92.8    & 76.1     & 76.0    & 89.6 & 93.9    & 63.3   & 58.6 & 78.7 & 91.3       & 51.8     & 71.0 & 97.5    & 91.7     & 77.7        & 58.4    & 75.6   & 85.4  & 87.1 & 69.3       & 34.7        & 76.0    & 79.2     & \multicolumn{3}{l}{91.5/88.6/71.3} & \multicolumn{2}{l|}{18.7/54.1}    & 65.4         & 2.95             & 0.45              \\
          MoCo v2                    & R50      & 93.4    & 76.3     & 72.2    & 84.4 & 90.7    & 60.2   & 48.3 & 75.1 & 89.9       & 51.1     & 69.4 & 96.9    & 90.1     & 77.2        & 58.1    & 71.8   & 83.0  & 85.6 & 69.1       & 35.3        & 73.9    & 79.1     & \multicolumn{3}{l}{92.5/89.9/76.9} & \multicolumn{2}{l|}{18.3/50.2}    & 66.9         & 3.09             & 0.43              \\ \hline
          \multirow{3}{*}{ViT}       & B/16      & 92.0    & 77.5     & 84.8    & 91.1 & 99.6    & 72.5   & 61.2 & 77.6 & 94.3       & 54.0     & 62.3 & 96.8    & 91.9     & 77.8        & 58.8    & 78.5   & 71.1  & 84.8 & 80.9       & 41.1        & 77.4    & -         & \multicolumn{3}{l}{-}              & \multicolumn{2}{l|}{-}            & -            & -                & -                  \\
                                     & L/16      & 96.8    & 87.3     & 89.7    & 92.8 & 99.7    & 76.7   & 71.4 & 79.7 & 96.8       & 59.0     & 67.6 & 97.3    & 92.7     & 73.6        & 63.4    & 86.0   & 81.9  & 85.9 & 82.2       & 42.6        & 81.1    & -         & \multicolumn{3}{l}{-}              & \multicolumn{2}{l|}{-}            & -            & -                & -                  \\
                                     & H/14      & 94.0    & 82.1     & 80.1    & 90.4 & 98.7    & 66.8   & 54.1 & 71.9 & 91.9       & 49.5     & 65.9 & 96.2    & 89.9     & 76.1        & 59.4    & 75.2   & 79.8  & 84.8 & 73.9       & 31.0        & 75.6    & -         & \multicolumn{3}{l}{-}              & \multicolumn{2}{l|}{-}            & -            & -                & -                  \\ \hline
          \multirow{4}{*}{CLIP}      & ViT-B /16 & 96.2    & 83.1     & 92.8    & 93.1 & 98.1    & 78.4   & 86.7 & 79.2 & 94.7       & 59.5     & 73.3 & 96.1    & 94.9     & 77.5        & 69.2    & 87.1   & 84.1  & 83.5 & 80.2       & 53.8        & 83.1    & -         & \multicolumn{3}{l}{-}              & \multicolumn{2}{l|}{-}            & -            & -                & -                  \\
                                     & R50      & 88.7    & 70.3     & 86.4    & 88.2 & 96.1    & 73.3   & 78.3 & 76.4 & 89.6       & 49.1     & 69.4 & 94.9    & 90.7     & 76.8        & 63.0    & 79.0   & 79.1  & 82.7 & 73.3       & 44.7        & 77.5    & 76.7     & \multicolumn{3}{l}{91.6/88.8/76.8} & \multicolumn{2}{l|}{18.3/52.3}    & 63.7         & 3.07             & 0.45              \\
                                     & R50$\times$16   & 92.2    & 74.9     & 93.3    & 93.5 & 98.3    & 79.2   & 88.7 & 79.1 & 93.7       & 62.7     & 75.8 & 95.6    & 93.5     & 77.2        & 69.0    & 86.1   & 85.9  & 83.5 & 81.5       & 53.9        & 82.9    & 83.6     & \multicolumn{3}{l}{92.5/89.9/76.9} & \multicolumn{2}{l|}{16.2/48.4}    & 68.7         & 2.83             & 0.39              \\
                                     & R101     & 91.1    & 73.5     & 88.9    & 91.0 & 96.4    & 75.1   & 84.0 & 76.3 & 92.0       & 50.7     & 66.2 & 94.9    & 91.7     & 76.9        & 65.1    & 82.1   & 80.1  & 82.8 & 75.7       & 47.5        & 79.1    & 79.5     & \multicolumn{3}{l}{90.8/86.7/74.6} & \multicolumn{2}{l|}{17.1/50.5}    & 66.5         & 3.01             & 0.45              \\ \hline
          \multirow{3}{*}{BiT}       & M-R50     & 94.9    & 82.2     & 83.3    & 91.5 & 99.4    & 69.9   & 59.0 & 77.3 & 93.9       & 55.6     & 65.4 & 96.9    & 91.1     & 76.9        & 60.0    & 76.5   & 82.8  & 82.5 & 76.7       & 40.4        & 77.8    & -         & \multicolumn{3}{l}{-}              & \multicolumn{2}{l|}{-}            & -            & -                & -                  \\
                                     & S-R50     & 91.7    & 74.8     & 72.5    & 92.3 & 92.0    & 61.1   & 53.5 & 72.4 & 91.2       & 52.5     & 69.8 & 96.8    & 89.0     & 76.3        & 55.9    & 69.1   & 81.9  & 83.0 & 75.2       & 35.0        & 74.3    & -         & \multicolumn{3}{l}{-}              & \multicolumn{2}{l|}{-}            & -            & -                & -                  \\
                                     & M-R152$\times$4  & 97.6    & 88.2     & 87.2    & 92.4 & 99.3    & 75.0   & 49.1 & 79.9 & 95.4       & 43.4     & 63.8 & 97.1    & 91.5     & 77.9        & 64.2    & 83.4   & 80.8  & 83.8 & 81.4       & 46.7        & 78.9    & -         & \multicolumn{3}{l}{-}              & \multicolumn{2}{l|}{-}            & -            & -                & -                  \\ \hline
          \multirow{2}{*}{Instagram} & 32$\times$8d     & 95.0    & 78.2     & 83.5    & 95.5 & 90.8    & 67.9   & 72.3 & 75.3 & 93.3       & 53.9     & 55.5 & 94.6    & 87.6     & 75.7        & 60.4    & 78.2   & 74.7  & 82.3 & 83.3       & 41.1        & 77.0    & -         & \multicolumn{3}{l}{-}              & \multicolumn{2}{l|}{-}            & -            & -                & -                  \\
                                     & 32$\times$48d    & 96.8    & 83.4     & 86.9    & 95.5 & 93.6    & 72.2   & 76.6 & 77.2 & 95.8       & 53.2     & 56.3 & 95.2    & 88.9     & 75.8        & 63.2    & 82.3   & 77.2  & 82.7 & 85.2       & 44.2        & 79.1    & -         & \multicolumn{3}{l}{-}              & \multicolumn{2}{l|}{-}            & -            & -                & -                  \\ \hline
          DetCo                      & R50      & -       & -        & -       & -    & -       & -      & -    & -    & -          & -        & -    & -       & -        & -           & -       & -      & -     & -    & -          & -           &         & 77.6     & \multicolumn{3}{l}{90.2/87.7/68.1} & \multicolumn{2}{l|}{17.3/51.2}    & 65.6         & 3.19             & 0.43              \\ \hline
          \end{tabular}
       }
  \end{adjustbox}
 \caption{\textbf{Linear probe performance of various pretrained models.} We evaluate on classification (CLS), object detection (DET), semantic segmentation (SEG) and depth estimation (DEP) tasks. We report AP50 for both PASCAL VOC~\cite{everingham2010pascal} and WIDER FACE~\cite{yang2016wider} with easy, medium and hard subsets, and MR$^{-2}$ for reasonable and heavy occlusion setups of CityPersons~\cite{zhang2017citypersons}. For semantic segmentation, mIoU is evaluated on PASCAL VOC 2012~\cite{everingham2010pascal}. KITTI~\cite{kitti} and NYUv2~\cite{NYUDEPTH} are depth estimation tasks with RMSE metric. $\uparrow$ means higher is better, and $\downarrow$ denotes the opposite.}
 \label{tab:whole_results} 
\end{table*}

\section{Dataset Creation for Gigantic-Scale Pretraining}
\label{GVB}

Large-scale database is crucial for computer vision pretraining systems, in which ImageNet-21K, YFCC-100M are instances of the commonly used public datasets while JFT, IG, and WIT are proprietary datasets that cannot be accessed by the society. 
Attributed to these datasets, visual representation learning has made significant progress.
Nonetheless, general vision pretraining requires more comprehensive datasets with an extensive spectrum of visual domains. Existing datasets, however, usually are limited in scope.
To address this, we construct a well-designed data system, GV-D, which consists of GV-D-10B with multi-modal data, as well as GV-D$_\text{c}$-36M\footnote{GV-D$_\text{c}$-36M is the version we used for our experiments. Currently its amount has reached 70 million.}, GV-D$_\text{d}$-3M and GV-D$_\text{s}$-143K with classification, detection, and segmentation annotations, respectively.

Based on the goal of obtaining wild data distribution, we collect images with 1.65 million queries from multiple sources such as web galleries, open-source datasets, etc. To reduce noise, we process raw data with both vision and language models. As a result, we collect a huge amount, quantitatively, 10 billion of image data.
We refer to this data system as GV-D-10B.

To efficiently obtain annotations for classification and detection tasks, we propose a hierarchical label system with a rigorously-defined taxonomy. 
The label system is a crucial component of our GV-D with more than 115K concepts from existing datasets, WikiData and WordNet. 
Powered by the hierarchical label system, we annotate 8.1 million images, including 10.5 million semantic labels and 2.7 million bounding boxes. Then, we construct two datasets as GV-D$_\text{c}$-36M and GV-D$_\text{d}$-3M. The construction consists of three main steps: 
1) inheriting public datasets including both their concepts and images. We utilize ten image classification datasets and three object detection datasets. Note that our GV-D$_\text{s}$-143K is also composed by inheriting four public datasets.
2) obtaining new raw
images. By searching query words from Flickr\footnote{\url{https://www.flickr.com}}, we collect raw data and form an unlabeled data pool. Those query words are sourced from the concepts of our label system.
For image classification, one query is simply one visual concept. 
For object detection, one query consists of two concepts, where one is a common concept, \eg \textit{dog}, and the other concept is either a semantic word referring to the scene, \eg \textit{street}, or also a common word, \eg \textit{ball}. To further enrich the results of searching, any given query word can be converted to its synonyms or its Chinese, Spanish, Dutch and Italian version.
3) building an active annotation pipeline. Active learning is a powerful technique to improve data efficiency for supervised learning methods, as it aims at selecting the smallest possible training set to reach a required performance. 

As shown in Tab.~\ref{tab:dataset_summary}, we compare our GV-D with other large-scale datasets for visual pretraining. Our datasets surpass the others in terms of the number of concepts, images and annotations.

\section{Towards Billion-Level Vision Model Design}

For years, convolutional neural networks (ConvNet) have dominated visual representations learning and demonstrated stable transferability on various downstream tasks, such as image classification, object detection and semantic segmentation.
Recently, Vision Transformer (ViT)~\cite{dosovitskiy2020image} achieves comparable performance on ImageNet-1k~\cite{russakovsky2015imagenet} with only vanilla transformer layers encoding image patch tokens by self-attention operators.
ViT also shows potentially higher capacity when trained with large-scale datasets (\eg ImageNet-21K, JFT-300M) than ConvNets.

Despite its merits on performance, some works~\cite{dai2021coatnet,d2021convit} point out that pure transformer layers may lack certain inductive biases and thus require more data and computing resources to counteract.
Besides, the computational cost of self-attention is superlinear with respect to the number of input tokens, which restricts applications to scenarios requiring high input resolution.
This suggests that we should hybridize convolution and transformer to balance two aspects - efficiency and effectiveness.

In this section, we introduce the \mtb{} family, a series of hybrid architectures with better generalization ability and higher model capacity. 
We firstly construct a unified search space, covering searchable configurations of both convolution and transformer. Then, we directly search on a large-scale dataset, ImageNet. We evaluate the searched \mtb{} on various vision tasks, \eg ImageNet classification, COCO object detection and ADE20K semantic segmentation. Our \mtb{} achieves higher performances under comparable resource constraints.

\subsection{Overview}

We aim to design an efficient but effective architecture family by hybridizing convolution and transformer.
This can help us better explore the respective roles of the two operators when combined.
Some recent works~\cite{wu2021cvt, yuan2021incorporating} attempt to incorporate convolution into self-attention or feed-forward network (FFN) sub-layers.
Others~\cite{dai2021coatnet, d2021convit} explore how to stack different types of blocks 
to form a complete network.
These design paradigms do lead to some high-performing architectures, but it is hard to figure out the optimal combination due to the subjective bias introduced by manual designs.

\begin{figure*}[!t]
    \centering
    \includegraphics[width=\textwidth]{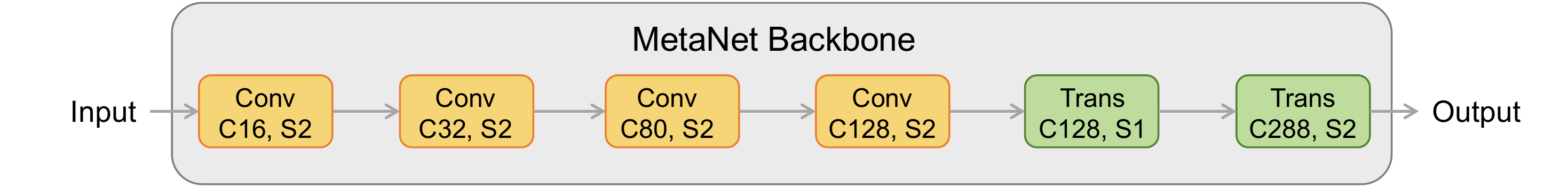} 
    \caption{\textbf{Searched \mtb{} architecture.} Conv and Trans represent convolution and transformer blocks respectively. C and S refer to the output channel number and stride of each stage.}
    \label{fig:backbone}
\end{figure*}

To better understand the design principle for hybrid blocks, we perform an architecture search for a combination of convolution and transformer operators, trying to automatically assemble the operators to create strong vision networks. While previous Neural Architecture Search (NAS) works 
mainly search for the best network size, we jointly search for the operators and network size in a unified search space. We show that our proposed unified architecture search achieves promising performance, resulting in a highly competitive hybrid network architecture as illustrated in Fig.~\ref{fig:backbone}.

\subsection{Searching for Hybridizing Convolution and Transformer}
\label{sec:operation}

In this section, we focus on how to search for the ``optimal'' combination of convolution and transformer. 
We introduce a unified search space that contains general operators (GOPs, including convolution and transformer), and then search for the best combination of those operators jointly. 
The general operator block is modeled as:
\begin{equation}
    \mathtt{y} = \mathtt{x + Operation(x)},
\end{equation}
where $\mathtt{Operation}$ can be the convolution or self-attention operator, and $\mathtt{x}$, $\mathtt{y}$ represent input and output features, respectively. For convolution, we place the convolution operation inside the bottleneck~\cite{sandler2018mobilenetv2}, which can be expressed as
\begin{equation}
    \mathtt{Operation(x)}\ =\ \mathtt{Proj_{ec\rightarrow c}(Conv(Proj_{c\rightarrow ec}(x)))}.
\end{equation}
The $\mathtt{Conv}$ operation can be either regular convolution or depth-wise convolution $\mathtt{DWConv}$~\cite{chollet2017xception}, and the $\mathtt{Proj}$ represents a linear projection. The subscript $\mathtt{ec\rightarrow c}$ denotes projecting $\mathtt{ec}$ channels to $\mathtt{c}$. For self-attention, operating on the large bottleneck feature map can be quite slow. Following previous works~\cite{dosovitskiy2020image, mixer}, we separate them from the bottleneck for computational efficiency, and the $\mathtt{Proj}$ is implemented inside the FFN~\cite{dosovitskiy2020image} sub-layer. 
This can be formulated as:
\begin{align}
    \mathtt{y}\ &\mathtt{ = y' + FFN(y')},\\
    \mathtt{y'}\ &\mathtt{ = x + SA(x)},\\
    \mathtt{FFN(y')}\ &\mathtt{ = Proj_{ec \rightarrow c}(Proj_{c\rightarrow ec}(y'))},
\end{align}
where $\mathtt{SA}$ can be either vanilla self-attention or local self-attention $\mathtt{LSA}$.

There are two main advantages of representing the different operators in a unified format and search space: 1) We can characterize all operators with the same set of configuration hyper-parameters except for the operation type, \eg expansion rate and channel size. As a result, the search space is much simplified, which in turn speeds up the search process. 2) Under the same network size configuration, different operator blocks has similar computational costs. The comparison between different operator combinations is fairer, which is crucial for NAS~\cite{tan2019mnasnet}.

\subsection{Unified Architecture Search}

\noindent\textbf{Unified Search Space.}
Prior art mainly focuses on the network size, and their approach achieves competitive results on various tasks. Compared with previous approaches, we jointly search the operator combination and network size. 
Our unified search space is as follows:
\begin{itemize}
    \item Operations $o$: convolution and transformer.
    \item Expansion ratio $e$: 2, 3, 4, 5, and 6.
    \item Relative repeat number $r$: -2, -1, 0, 1, and 2.
    \item Relative channel multiplier $c$: 0.5, 0.75, 1.0, 1.25, and 1.5.
\end{itemize}

In our unified search space, we search for a relative shift characterized by $r$ and $c$ based on an existing architecture, \eg EfficientNet~\cite{tan2019efficientnet}. We partition the desired model into 6 stages and search the above configuration per stage. Our overall search space size is 2.4$\times$10\textsuperscript{14}. As the search space is too big, it is unfeasible to run all possible architectures. To find the optimal one, we use reinforcement-learning-based NAS~\cite{fnas} to speed up the search process.

\noindent\textbf{Search Pipeline.}
We utilize reinforcement learning to search for the optimal hybrid architecture automatically. Concretely, we follow the previous work~\cite{fnas} and map an architecture in the search space to a list of tokens, which are determined by a sequence of actions generated by an RNN. The RNN is optimized using the PPO algorithm~\cite{ppo} by maximizing the expected reward. In our implementation, we simultaneously optimize accuracy and the theoretical computation cost (FLOPs). To handle the multi-objective optimization problem, we use a weighted product customized as~\cite{tan2019mnasnet} to approximate the Pareto optimal state. For one sampled architecture $m$, the reward is formulated as:
\begin{equation}
r(m) = a(m) \times \left[\frac{t}{f(m)}\right]^{\alpha},
\end{equation}
where functions $a(m)$ and $f(m)$ return the accuracy and FLOPs of $m$, $t$ is the target FLOPs, and $\alpha$ is a weight factor that balances the accuracy and computational cost.

During the search process, thousands of combinations of GOPs are trained on a proxy task with the same setting, which gives us a fair comparison among those combinations. When the RNN converges, the top-$k$ architectures with the highest reward will be trained with the full setting, and the top-performing one will be kept for model scaling and transferring to other tasks.

\noindent\textbf{Model Scaling.}
Following previous practice~\cite{tan2019efficientnet}, we search for a small model \mtbone{} and scale it up to \mtbfifteen{}. As pointed out in~\cite{tan2019efficientnet}, scaling a single dimension leads to saturated accuracy, and they propose compound scaling to scale depth, width, and resolution jointly. 
However, as the computation cost of self-attention grows quadratically with the input resolution, it is infeasible to scale resolution together with depth and width. For better efficiency, we keep the resolution fixed in the pretraining phase, and fine-tune the resulting models at the target resolution. Besides, we empirically find that the gradient of deep transformer-based architectures easily overflows when trained with mixed precision. 
To improve training stability, we scale the depth with a smaller coefficient. 
Based on the settings described above, we search the optimal scale factors for width and depth of different \mtb{}s, \eg \mtbfour{}, \mtbseven{} and \mtbfifteen{}, according to the convergence speed of the network within one training epoch.

\begin{table*}[t]
  \centering
  \small
    \begin{tabular}{lc|ccc|c}
    \toprule
    \multirow{2}{*}{Model} & \multirow{2}{*}{Family} & \multirow{2}{*}{\makecell{Input Size}} & \multirow{2}{*}{\makecell{\#FLOPs \\ (G)} } & \multirow{2}{*}{\makecell{\#Params \\ (M)}} & \multirow{2}{*}{\makecell{Top-1 \\ Acc.}} \\
    & & & & & \\
    \midrule
    EffNet-B2 \cite{tan2019efficientnet} & C     & 260   & 1.0     & 9.2   & 80.1 \\
    EffNetV2-B1\textsuperscript{$\ddagger$} \cite{tan2021efficientnetv2} & C     & 240   & 1.2   & 8.1  & 79.8 \\
    RegNetY-4GF \cite{radosavovic2020designing} & C     & 224   & 4.0     & 20.6  & 80.0 \\
    DeiT-Small \cite{deit} & T     & 224   & 4.3   & 22    & 79.8 \\
    XCiT-T24 \cite{xcit} & T & 224 & 2.3 & 12 & 79.4 \\
    PVT-Small \cite{wang2021pyramid} & T     & 224   & 3.8   & 24.5  & 79.8 \\
	CaiT-XXS36 \cite{cait} & T & 224 & 3.8 & 17.3 & 79.1 \\
    Ours, \textbf{\mtbone{}} & H     & 224   & 1.1  & 14    & \textbf{81.0} \\
    \midrule
    EffNet-B4 \cite{tan2019efficientnet} & C     & 380   & 4.2   & 19    & 82.9 \\
    RegNetY-16GF \cite{radosavovic2020designing} & C     & 224   & 16    & 84    & 82.9 \\
    ResNeSt-101 \cite{resnest} & C & 256 & 13 & 48 & 83.0 \\
    Swin-B \cite{liu2021swin} & T     & 224   & 15.4  & 88    & 83.3 \\
    CSwin-T \cite{cswin} & T & 224 & 4.3 & 23 & 82.7 \\
    DeepViT-L \cite{deepvit} & T & 224 & 12.5 & 55 & 83.1 \\
    XCiT-L24 \cite{xcit} & T & 224 & 36.1 & 189 & 82.9 \\
    CaiT-S36 \cite{cait} & T & 224 & 13.9 & 68.2 & 83.3 \\
    LV-ViT-S \cite{lvvit} & T & 224 & 6.6 & 26.2 & 83.3 \\
    T2T-ViT-24 \cite{t2t} & H & 224 & 15.0 & 64.1 & 82.6 \\
    CvT-21 \cite{d2021convit} & H     & 384   & 24.9  & 32    & 83.3 \\
    CoAtNet-1 \cite{dai2021coatnet} & H & 224 & 8.4 & 42 & 83.3 \\
    Ours, \textbf{\mtbfour{}} & H     & 256   & 4.6   & 29.8  & \textbf{83.4} \\
    \midrule
    EffNet-B6 \cite{tan2019efficientnet} & C     & 528   & 19   & 43    & 84.0 \\
    NFNet-F0 \cite{brock2021high} & C     & 256   & 12.4  & 71.5  & 83.6 \\
    ResNeSt-200 \cite{resnest} & C & 320 & 36 & 70 & 83.9 \\
    CSWin-B \cite{cswin} & T & 224 & 15 & 78 & 84.2 \\
    CaiT-M36 \cite{cait} & T & 224 & 53.7 & 270.9 & 83.8 \\
    LV-ViT-M \cite{lvvit} & T & 224 & 16 & 56 & 84.1 \\
    CoAtNet-2 \cite{dai2021coatnet} & H & 224 & 15.7 & 75 & 84.1 \\
    BoTNet-T7 \cite{botnet} & H & 256 & 19.3 & 79 & 84.2 \\
    Ours, \textbf{\mtbseven{}} & H     & 256   & \textbf{10.2}  & 78.5  & \textbf{84.2} \\
    \bottomrule
  \end{tabular}%
  \caption{\textbf{\mtb{} performance on ImageNet.} ``C'', ``T'' and ``H'' correspond to convolution-based, transformer-based and hybrid architectures. $\ddagger$: EfficientNetV2 is trained with progressive learning.}
  \label{tab:arch_imgnet}
\end{table*}%

\subsection{Results}

\noindent\textbf{Implementation Details.}
To find the optimal architecture in our search space, we directly search on ImageNet~\cite{russakovsky2015imagenet}. We reserve 50K images from the training set as a validation set. For each sampled architecture, we train it for 5 epochs. After that, we calculate the reward of the architecture with its FLOPs and the accuracy on the validation set. We set the target FLOPs $t$ and weight factor $\alpha$ in the reward function to 600M and -0.07, respectively. 
After the search process, we fully train the best $k=5$ architectures on ImageNet and select the top-performing one as the final \mtb{} architecture.

For regular ImageNet training, we mostly follow the training strategy in \cite{deit}. We pretrain all our \mtb{} models with 224$\times$224 resolution and fine-tune at the target resolution for training efficiency. Besides, we also transfer pretrained \mtb{}s to other tasks, \eg object detection on COCO and semantic segmentation on ADE20K.
For COCO object detection, we use the Mask R-CNN framework and compare the performance under 1x and 3x schedules. For ADE20K semantic segmentation, we apply the UperNet framework and report mIoU (\%) of different architectures under the same training setting.

\noindent\textbf{ImageNet Results.}
Tab.~\ref{tab:arch_imgnet} compares the image classification performance of our searched \mtb{} with other architectures, including convolution-based, transformer-based and hybrid architectures. Our models achieve better accuracy while being computationally efficient. 
More specifically, our \mtbone{} achieve 81.0 top-1 accuracy with 1.1 GFLOPs, outperforming EfficientNet-B2~\cite{tan2019efficientnet} with comparable FLOPs. Our \mtbfour{} achieves 83.4\% top-1 accuracy with 4.6 GFLOPs, which outperforms convolution-based EfficientNet-B4, transformer-based Swin-B, and the hybrid architecture CvT-21. For larger models, our \mtbseven{} achieves 84.2\% with 10.2 GFLOPs, surpassing NFNet-F0 and BoTNet-T7 with less FLOPs.

\noindent\textbf{Detection and Segmentation Results.} 
We evaluate \mtbone{} and \mtbfour{} by using them as the feature extractor in detection and segmentation frameworks. We compare our \mtb{} with other convolution- and transformer-based architectures.
As shown in Tab.~\ref{tab:arch_det_seg}, our searched \mtb{} consistently outperforms convolution-based ResNet~\cite{he2016deep}, transformer-based PVT~\cite{wang2021pyramid} and Swin-Transformer~\cite{liu2021swin}. In the object detection task, \mtbfour{} achieves 46.7 AP@box with 1x schedule and 48.2 AP@box with 3x schedule, which are 4.5 and 2.2 points better than Swin-T, respectively. For ADE20K semantic segmentation, we achieve 49.3\% mIoU with 56M parameters. Compared with Swin-T, our \mtb{} leads by 4.8\% mIoU with a similar number of parameters. All results demonstrate the strong generalization ability of our searched \mtb{} models.

\begin{table*}[t]
 \centering
 \small
  \begin{tabular}{l|c|cc|cc|cc}
\toprule
\multirow{2}{*}{Backbone} & \multirow{2}{*}{\makecell{\#Params (M) \\ Det/Seg}} & \multicolumn{2}{c|}{Mask R-CNN 1x} & \multicolumn{2}{c|}{Mask R-CNN 3x} & \multirow{2}{*}{\makecell{UperNet \\ mIoU (\%)}} \\ \cline{3-6}
 & & AP@box & AP@mask & AP@box & AP@mask &  \\ 
 \midrule
 ResNet-18~\cite{he2016deep}   & 31/ -  & 34.0 & 31.2 & 36.9 & 33.6 & -  \\
 ResNet-50~\cite{he2016deep}   & 44/ -  & 38.0 & 34.4 & 41.0 & 37.1 & -  \\
 PVT-Tiny~\cite{wang2021pyramid}   & 33/ -  & 36.7 & 35.1 & 39.8 & 37.4 & -  \\
 Ours, \textbf{\mtbone{}}       & 31/41  & \textbf{41.2} & \textbf{38.5} & \textbf{44.8} & \textbf{40.3} & \textbf{42.9} \\
 \midrule
 ResNet-101~\cite{he2016deep}  & 63/86  & 40.4 & 36.4 & 42.8 & 38.5 & 44.9  \\
 PVT-Small~\cite{wang2021pyramid}  & 44/ -  & 40.4 & 37.8 & 43.0 & 39.9 & -  \\
 Swin-T~\cite{liu2021swin}     & 48/60  & 42.2 & 39.1 & 46.0 & 41.6 & 44.5  \\ 
 Ours, \textbf{\mtbfour{}}       & 47/56  & \textbf{46.7} & \textbf{43.6} & \textbf{48.2} & \textbf{44.0} & \textbf{49.3} \\ 
 \bottomrule
\end{tabular}
\caption{Object detection, instance segmentation and semantic segmentation performance on the COCO val2017 and ADE20K val set.}
 \label{tab:arch_det_seg}
\end{table*}

\section{Pretraining Up-A Stage: Acquiring the Amateur from Multi-modal Supervisions}
\label{sec:stage1}
\begin{figure*}[t]
     \centering   
     \includegraphics[width=\linewidth]{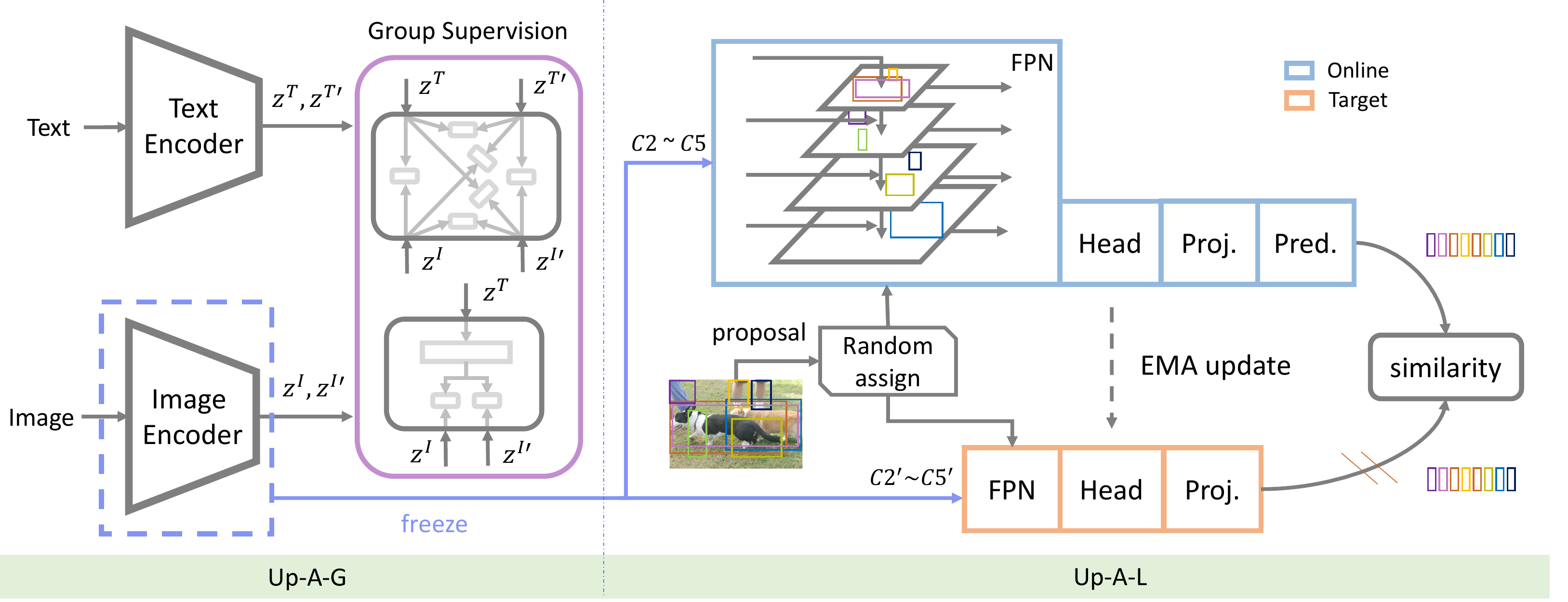}
     \caption{\textbf{Framework of Upstream-Amateur.} The Up-A stage has two pretraining phases: Upstream-Amateur for Global Representation (Up-A-G) and Upstream-Amateur for Local Representation (Up-A-L). Up-A-G (left) uses group-supervision functions for learning from richer supervision. $\{z^I,z^I{'}\}$ and $\{z^T,z^T{'}\}$ are embedding features of the augmented images and texts separately. To improve the performance of Up-A-G on dense prediction CV tasks, in the phase of Up-A-L (right), the well-trained vision-language models are adapted by the local self-supervision learning method.
     $\{C2 \small{\sim} C5\}$ and $\{C2' \small{\sim} C5'\}$ denote the feature maps of the augmented images from the frozen backbone network, in which the numeric subscripts represent the level of features. "Proj." and "Pred." are abbreviations of the projector and predictor. The online and target network are derived from BYOL~\cite{grill2020bootstrap}, which can help prevent models from collapsing.}
     \label{fig:up-a}
\end{figure*}

Recent large-scale multi-modal pretraining methods~\cite{CLIP,jia2021scaling} have demonstrated huge potential for learning high-quality visual representations. These works leverage image-text supervision, yet disregard other rich supervisory signals from image-image and text-text pairs. To fully exploit the advantage of the large-scale multi-modal data for acquiring an amateur model, we propose Upstream-Amateur (Up-A), a vision-language pretraining framework (see Fig.~\ref{fig:up-a}) which simultaneously mines intra-modal and cross-modal knowledge. 

In this stage, two pretraining phases, Upstream-Amateur for Global Representation (Up-A-G) and Upstream-Amateur for Local Representation (Up-A-L), are constructed sequentially. In the phase of Up-A-G (left branch in Fig.~\ref{fig:up-a}), we propose group-supervision functions for richer vision-language supervision. As for Up-A-L (right branch in Fig.~\ref{fig:up-a}), the well-trained vision-language models are adapted to be friendly to dense prediction tasks, in which the FPN~\cite{lin2017feature} and a head module are further adapted based on the pretrained multi-modal representation without any labeled data. The proposed Upstream-Amateur improves the downstream performance of multi-modal pretraining on both classification and detection tasks, and serves as a good starting point for further pretraining towards a general vision model.

\subsection{Method}
\label{subsec:stage1_approach}

\subsubsection{Upstream-Amateur for Global Representation}
\label{subsec:stage1_global}

Considering the semantically dense learning signals within image-text pairs, we propose the group-supervision functions, including intra-modal and cross-modal supervision (ICS) and similar-text supervision (STS).

For intra-modal and cross-modal supervision (ICS), supervision signals are mined from augmented images pairs, augmented text pairs and image-text pairs. Augmented images $\{x_i^I,x_i^{I}{'}\}$ and augmented texts $\{x_i^T,x_i^{T}{'}\}$ are fed to image encoder and text encoder respectively, generating normalized outputs $\{z_i^I, z_i^{I}{'}\}$ and $\{z_i^T,z_i^{T}{'}\}$.
The loss function is then constructed across augmented image-text pairs as Eq.~\ref{eq:ICS}, in which $\mathcal{L}_\text{CL}$ is the contrastive loss for multi-modal data, $\mathcal{L}_\text{CS}$ is the cosine similarity loss~\cite{chen2020simple}, $\mathcal{L}_\text{MLM}$ is Masked Language Modeling loss~\cite{devlin2018bert} and $\tau$ represents the temperature parameter. In this way, supervision signals are established within each modality and across different modalities, letting multi-view image-text pairs bring more invariant and robust information.

\begin{equation}
    \begin{aligned}
    \mathcal{L}_\text{ICS} = \frac{1}{N} \sum^N_{i=1} &  [\mathcal{L}_\text{CS}(z_i^I, z_i^{I}{'}) + \mathcal{L}_\text{MLM}(z_i^T, z_i^{T}{'}) ] 
     + \mathcal{L}_\text{CL}(z^I, z^T) + \mathcal{L}_\text{CL}(z^I, z_i^{T}{'})   + \mathcal{L}_\text{CL}(z_i^{I}{'}, z^T) + \mathcal{L}_\text{CL}(z_i^{I}{'}, z_i^{T}{'}), \\
    & \mathcal{L}_\text{CL}(x,y) = -\frac{1}{N} \sum^N_{i=1} \log\frac{\exp(x_i \cdot y_{+}/\tau)}{\sum_{j=1}^{N} \exp(x_i \cdot y_j/\tau)} -\frac{1}{N} \sum^N_{i=1} \log\frac{\exp(x_{+}\cdot y_i/\tau)}{\sum_{j=1}^{N} \exp(x_j \cdot y_i/\tau)}.
    \label{eq:ICS}
    \end{aligned}
\end{equation}

For similar-text supervision(STS), we maintain a first-in-first-out feature queue that records historical text features. The feature queue can represent the whole text data distribution. Then, we utilize the nearest-neighbor search in the feature queue according to the current $z_i^T$ to get the semantically similar text features. Finally, similar text features are applied to supervise augmented images $\{x_i^I,x_i^{I}{'}\}$ by the contrastive loss.

Each of the above supervisions is found to boost the performance of visual representation learning. Aggregating these supervisions lead to our final group-supervision $\mathcal{L}_\text{GS}$ as Eq.~\ref{eq:stage1}, in which $\alpha$ is a hyper-parameter.

\begin{equation}
    \mathcal{L}_\text{GS} =  (1-\alpha)\mathcal{L}_\text{ICS} + \alpha \mathcal{L}_\text{STS}.
    \label{eq:stage1}
\end{equation}

\subsubsection{Upstream-Amateur for Local Representation}

To guide the previous vision-language pretrained model to attend on more scale-invariance knowledge, we further pretrain FPN and part of a Faster R-CNN~\cite{ren2015faster} head module in a self-supervised manner. Inspired by SoCo~\cite{wei2021aligning}, we propose Upstream-Amateur for Local representation (Up-A-L). Up-A-L also utilizes the selective search method~\cite{uijlings2013selective} to generate proposals as local supervision. Yet our method aims to adapt a well-trained vision-language pretrained backbone to focus on local representations, while SoCo intends to learn object-level representations from scratch. In Up-A-L, the backbone part is frozen. The FPN and head modules function as transfer modules to map the proposals with the same position to similar representations. The whole pipeline is displayed in Fig.~\ref{fig:up-a}.

The frozen backbone is inherited from Up-A-G described in Sec.~\ref{subsec:stage1_global}, and processes the augmented images into multi-scale features $\{Ci, Ci'\}$, the $i^\text{th}$ of which represents feature maps from the $i^\text{th}$ stage of the backbone. Then, as shown in Fig.~\ref{fig:up-a}, $\{Ci, Ci'\}$ and proposals are fed to the online and target networks separately. The proposal features are cropped from FPN outputs through a random assignment mechanism, which assigns object proposals to $\{P2, P3, P4, P5\}$ randomly according to the probability distribution of $\{0.1, 0.2, 0.3, 0.4\}$. Next, proposal features are aligned to $7\times7$ shape by RoIAlign~\cite{maskrcnn}. Finally, after being embedded by the head, projector and predictor, the proposal features of the augmented images are fed into the consistency loss~\cite{grill2020bootstrap}. Attributed to Up-A-L, the pretrained model becomes more adapted to tasks requiring larger resolution and local representation, such as object detection.

\subsection{Experiments}
\label{subsec:stage1_results}

\subsubsection{Setting}
\noindent\textbf{Pretraining.}
We train our model from scratch on the GV-D-10B web dataset. During the pretraining process, the amount of training data gradually decreases to accelerate the convergence of the model. The input resolution of R50, MN-B4 and MN-B15 image encoders is $224\times224$, $256\times256$ and $256\times256$ respectively. We apply stochastic depth ratio of $0.2$ for \mtbfour{} and $0.6$ for \mtbfifteen{}. The maximum context length of the text encoder is $76$. We use SGD optimizer for the R50 model, and a mixed AdamW-SGD optimizer for the other two hybrid backbones. The learnable temperature parameter $\tau$ is initialized to $0.07$. The loss weight $\alpha$ is set to $0.5$.

For Up-A-L, we use a self-supervised learning setting as described in \cite{wei2021aligning} except that the resolution of the input image is set to $448\times 448$, the proposal number is set to 8, and the resolution of view3 is also set to $448\times 448$. Box jitter is not used because it is harmful under our setting. Since the backbone is frozen, we are able to use a large batch size of $1024$.

\noindent \textbf{Downstream Evaluation.} We mainly follow the evaluation setting of our benchmark which is elaborated in Sec.~\ref{subsec:eval_setting}. The only difference is that for fair comparison with CLIP~\cite{CLIP} on a subset of classification datasets, we only use the second strategy with the L-BFGS optimizer for evaluation.

\subsubsection{Training Strategy}

\noindent \textbf{Online Loss Monitoring and Low-Cost Auto-Resume.}
Due to a large amount of data noises, the training process is often unstable and easy to collapse. This problem is alleviated by determining whether the gradient of an iteration is going in the opposite direction of general training loss descent. Specifically, during training, if the loss of the current batch of data gradually increases and exceeds a set threshold of 0.5, it means that the batch will likely cause training abnormalities. At this time, the loss of the current batch will not be back-propagated, the current weight would be rolled back for 10 iterations, and then training continues with the data of the next iteration. In this way, we achieve an almost effortless local resuming, and avoid the huge potential cost of restarting the whole pretraining after a training crash.

\noindent \textbf{FP16-AdamW-SGD.}
When the image encode has transformer component, we find that using a mixed Adamw-SGD optimizer better avoids overfitting in the later stage of training. Specifically, we use AdamW as the optimizer of the image encoder, while applying momentum SGD for the text encoder. In addition, the learning rate of the parameter controlling the temperature is 0.1 times that of the other parameters of the model. The whole training process uses FP16 for accelerated training.

\begin{table*}[t]
\centering
\begin{small}
    \begin{tabular}{l|c|cccccccccccc}
    \toprule  
     \makebox[1.8cm][l]{Model} & \begin{turn}{90}Average\end{turn} & \begin{turn}{90}CIFAR-100\end{turn} & \begin{turn}{90}Food-101\end{turn} & \begin{turn}{90}Oxford-IIIT Pets\end{turn} & \begin{turn}{90}Oxford 102 Flower\end{turn} & \begin{turn}{90}Caltech-101\end{turn} & \begin{turn}{90}CIFAR-10\end{turn} & \begin{turn}{90}SUN397\end{turn} & \begin{turn}{90}Stanford Cars\end{turn} & \begin{turn}{90}FGVC-Aircraft\end{turn} & \begin{turn}{90}PatchCamelyon\end{turn} & \begin{turn}{90}DTD\end{turn} & \begin{turn}{90}ImageNet\end{turn} \\ \midrule   
        CLIP-R50~\cite{CLIP} & 82.1 & 70.3 & 86.4 & 88.2 & 96.1 & 89.6 & 88.7 & 73.3 & 78.3 & 49.1 & 82.7 & 76.4 & 73.3 \\ 
        Up-A-G R50 & 84.8 & 79.1 & 83.5 & 91.4 & \textbf{99.2} & 94.1 & 93.8 & 71.4 & 82.5 & 54.2 & 83.1 & \textbf{78.3} & 76.6 \\ 
        Up-A-G MN-B4 & \textbf{87.0} & \textbf{85.3} & \textbf{88.1} & \textbf{92.7} & 98.9 & \textbf{95.5} & \textbf{96.6} & \textbf{73.6} & \textbf{83.4} & \textbf{60.2} & \textbf{84.5} & 77.9 & \textbf{80.5} \\ \hline
        CLIP-VIT-L~\cite{CLIP} & 89.2 & 87.4 & \textbf{95.9} & 95.1 & 99.2 & 96.0 & 97.9 & 82.2 & \textbf{91.5} & 71.6 & 85.6 & 83.0 & 85.4 \\
        Up-A-G MN-B15 & \textbf{90.2} & \textbf{90.5} & 95.1 & \textbf{95.5} & \textbf{99.7} & \textbf{97.3} & \textbf{98.5} & \textbf{82.6} & 85.8 & \textbf{79.4} & \textbf{88.4} & \textbf{83.5} & \textbf{86.9} \\ 
    \bottomrule
    \end{tabular}
\caption{Performance comparison with the previous state-of-the-art method CLIP in various classification tasks with 100\% downstream training data. 
\label{tab:stage1-clip100}}
\end{small}
\end{table*}

\subsubsection{Results}

\noindent\textbf{Classification.} We report our linear probe performance on 11 downstream datasets which overlap with \cite{CLIP} in Tab.~\ref{tab:stage1-clip100}. The results with R50 models show that our Up-A-G pretraining scheme with group-supervision method significantly surpasses CLIP in terms of downstream classification performance. Furthermore, applying group-supervision functions to \mtb{} backbones leads to stronger results, which also proves the effectiveness of our searched network architecture.

\noindent\textbf{Detection.} We verify the effectiveness of Up-A-L in Tab.~\ref{tab:stage1-det}. All results are linear probe performance under the 10\% setting of VOC07+12. They show that with any of the three backbone architectures, Up-A-L can improve the performance of the Up-A-G pretrained model in the detection task. Moreover, we also conduct ablation experiments on Up-A-L pretraining resolution, proposal number and random assignment mechanism, as listed in Tab.~\ref{tab:Up-A-L-ab}. Note that ``r224-p4'' represents feeding $224\times 224$ images into the model and the proposal number is 4, which is the default setting of SoCo. Attributed to our two-phase design, we can fix the backbone during Up-A-L, and thus are able to apply the more computationally heavy ``r448-p8'' setting, which improves the detection performance by 0.9\% AP50. The proposed random assignment mechanism further strengthens the pretraining model by 0.4\% AP50.

\begin{table*}[h]
\centering
\begin{small}
\begin{tabular}{lcccc}
\toprule  
Model       & AP     &  AP50 & AP75 \\
\midrule   
ImageNet R50&  36.7  &  69.5      &  33.9 \\ \hline
Up-A-G R50      &  40.2  &  74.1          & 38.7 \\
Up-A-L R50    &  \textbf{41.7}  &  \textbf{76.3} & \textbf{40.4} \\\hline
Up-A-G MN-B4    &  39.8  &  74.2          & 38.0  \\
Up-A-L MN-B4  &  \textbf{40.5}   &  \textbf{74.9} & \textbf{38.6}   \\\hline
Up-A-G MN-B15    &  42.6  &  76.5          &  42.8   \\
Up-A-L MN-B15  &  \textbf{44.1}  &  \textbf{78.4} & \textbf{44.6}    \\
\bottomrule
\end{tabular}
\caption{Detection performance of Up-A models on VOC07+12 object detection dataset under the 10\% split setting.}
\label{tab:stage1-det}
\end{small}
\end{table*}

\begin{table*}[h]
\centering
\begin{small}
\begin{tabular}{lcc|cc}
\toprule  
r224-p4     & r448-p8   &  random assign & AP  & AP50  \\
\midrule   
\checkmark &            &                & 40.9  & 75.0 \\ 
           & \checkmark &                & 41.5  & 75.9 \\
           & \checkmark &  \checkmark    & \textbf{41.7} & \textbf{76.3} \\
\bottomrule
\end{tabular}
\caption{Ablation study on Up-A-L. ``r'' represents resolution, and ``p'' denotes proposal number.}
\label{tab:Up-A-L-ab}
\end{small}
\end{table*}

\noindent\textbf{Benchmark Performance.} After Up-A-G and Up-A-L pretraining, we evaluate the final \amateur{} on our downstream benchmark with 26 datasets. The results are shown in Tab.~\ref{tab:stage1-full}. When using R50, our model outperforms the ImageNet pretrained model on all evaluated datasets by large margins. When scaling up the model from MN-B4 to MN-B15, our method obtains steady performance gains.

\begin{table*}[!h]
\centering
\begin{small}
\begin{tabular}{l|ccccccc}
\toprule  
Model  &CLS AVG $\uparrow$ &  VOC07+12 $\uparrow$ & WIDER FACE $\uparrow$ & CityPersons $\downarrow$ & VOC2012 $\uparrow$ & KITTI $\downarrow$ & NYUv2 $\downarrow$ \\
\midrule   
ImageNet R50  & 62.8   &  69.5   &  88.6/84.2/60.9    & 29.6/65.8    & 58.03     & 3.258 & 0.478 \\
Up-A R50      & 70.9  &  76.3   &  90.3/88.3/70.7     & 24.6/59.0     &  62.54     & 3.181 & 0.456 \\
Up-A \mtbfour{}     & 74.7  &  74.9   &  89.3/87.6/71.4    & 26.5/61.8     &  65.71      & 3.565 & 0.482 \\
Up-A \mtbfifteen{}    & 80.4  &  78.4  &  93.6/91.8/77.2    &  17.7/49.5     & 60.68   & 2.423 & 0.383 \\
\bottomrule
\end{tabular}
\caption{Performance of Up-A models on downstream benchmark with 10\% training data.}
\label{tab:stage1-full}
\end{small}
\end{table*}

\section{Pretraining Up-E Stage: Building Multiple Experts from the Amateur}
\label{sec:stage2}

\begin{figure*}[ht]
    \centering
    \includegraphics[width=0.8\linewidth]{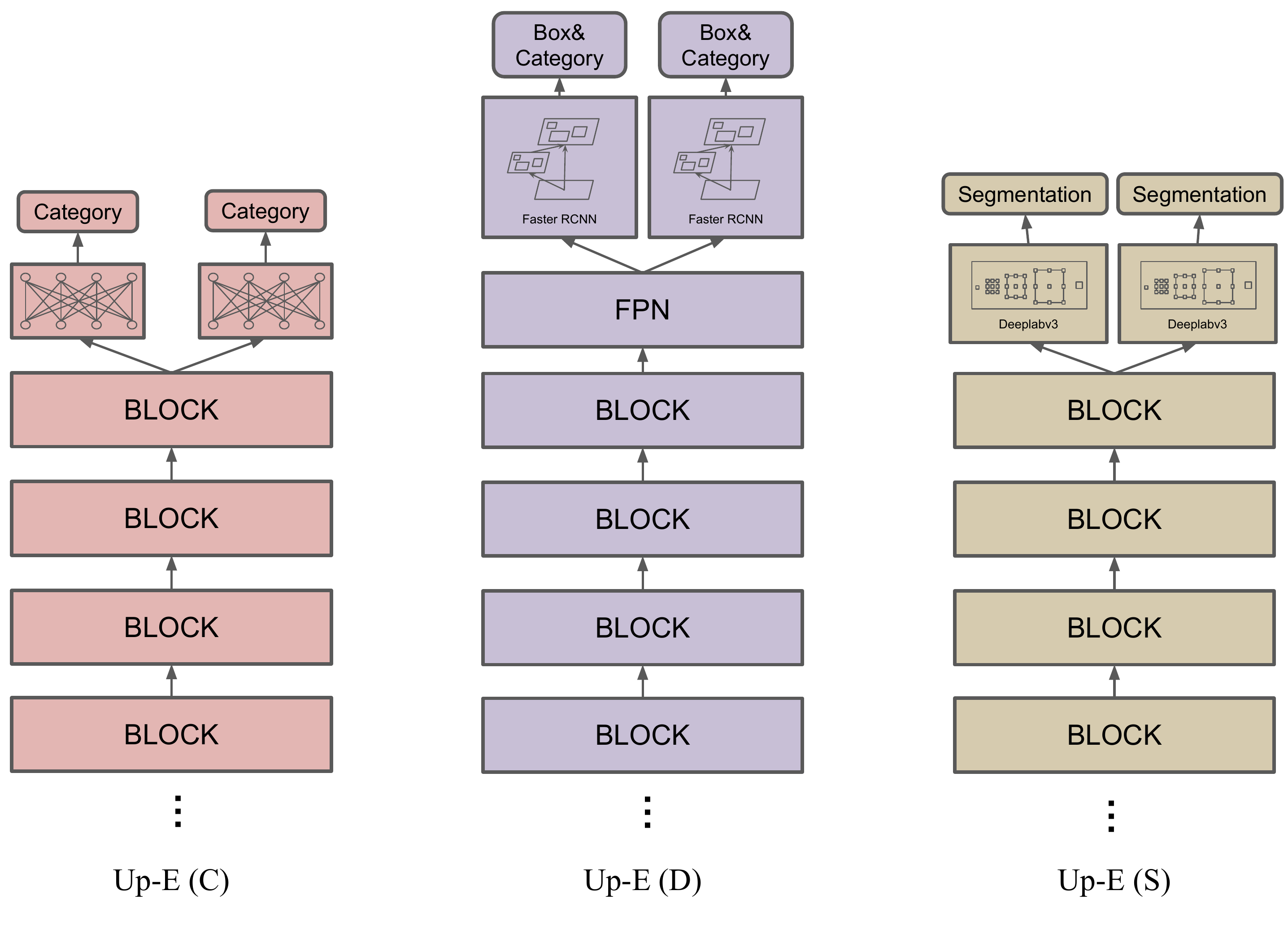}
     \caption{Overview of our \stagetwo{} architecture. The diagrams indicate our Up-E (C) , Up-E (D),  and Up-E (S) expert models. For Up-E (C), we put fully connected heads on top of the shared backbones for predicting the category. For Up-E (D), we first use a shared FPN to facilitate sharing between detection tasks, and then several Faster R-CNN heads are placed upon the FPN network for detecting objects. We use Deeplabv3 heads for our segmentation expert Up-E (S).}
     \label{fig:stage2-overview}
  \end{figure*}

The Up-A stage results in \amateur{} models which show good capabilities in general visual recognition problems. However, to fully master more specific tasks, such as detection and segmentation, more specialized pretraining within each task is still required, which motivates us to design this second pretraining stage, Up-E. 
Concretely, a package of task-specific networks, named {\em experts}, are trained based on the task-agnostic \amateur{} model from the previous stage. For each expert, we employ a simple multi-head design where each head is a dataset-specific sub-network that branches off a common, shared ``trunk'', as shown in Fig.~\ref{fig:stage2-overview}.
As such, we learn shared trunk parameters $\{\theta_t\}^\tau_{t=1}$ for $\tau$ expert models performing different tasks. When presented with a task-relevant downstream task, one can select an appropriate expert $\theta_t$. We argue that expert models trained with our approach can consolidate representations efficiently and achieve state-of-the-art performances on their respective tasks. In Sec.~\ref{subsec:stage2_method}, we further delineate our basic methodology for expert training. We then present our expert models \expertcls{}, \expertdet{}, and \expertseg{}, for image classification, object detection, and semantic segmentation, respectively, in the rest of this section.

\subsection{Method}
\label{subsec:stage2_method}

\noindent\textbf{Expert Definition.} Given a generic task type like classification or detection, we aim at creating an expert model with uniformly high performances on various datasets and benchmarks of the considered task. Generally, to learn the designated task, we have access to multiple data sources, each of which is annotated on its own label space. As is presented in Fig.~\ref{fig:stage2-overview}, we enforce models for different data sources to share most model parameters, essentially the ``backbone'' part for feature extraction, and have their own unique heads to produce final predictions which fit their own annotations. In this way, the learned shared parameters should be able to perform well on any of the datasets used during expert learning, reaching the goal of acquiring high-performing and generalizable representations. Our design is efficient and does not require any extra process on unifying annotations of different datasets. We conduct experiments on these three most basic computer vision tasks and all achieve state-of-the-art results, demonstrating the efficacy of our proposed model.

\noindent\textbf{Practical Implementation.} When training with multiple data sources, a training mini-batch would be a mixture of data from different sources which should be passed on to different dataset-specific heads correspondingly. To implement with normal forward and backward propagation, we distribute datasets to different computational devices, and the total resources allocated to a dataset is roughly proportional to the size of its training set. In this way, we ensure that one device only processes samples from the dataset assigned to it, and the other heads which do not correspond to that assigned dataset simply have zero gradient on that device. A consequently required technique is \emph{globally synchronized batch normalization}, \ie, synchronizing the statistics of batch normalization layers across all devices. This trick guarantees consistent performances on all datasets during evaluation.

\subsection{Image Classification Expert \expertcls{}}
\label{subsec:stage2_classification}

\subsubsection{Datasets}

We select large-scale datasets from various domains for training our classification expert \expertcls{}.

\noindent\textbf{ImageNet-21K}~\cite{deng2009imagenet} is a superset of the ImageNet's ILSVRC2012 variant~\cite{russakovsky2015imagenet}, and has $\sim$14.2M images belonging to 21,841 diverse classes. For simplicity, we ignore the hierarchical structure of the categories, and directly use this dataset for single-label classification.

\noindent\textbf{iNat2021}~\cite{van2021benchmarking} is an upgraded version of the previous iNat2017-2019 species classification datasets~\cite{van2018inaturalist}. It has $\sim$2.7M training images and 100k validation data, divided into 10k species spanning the entire tree of life.

\noindent\textbf{Herbarium 2021}~\cite{de2021herbarium} Half-Earth dataset contains more than 2.5M images of vascular plant specimens from 64,500 imbalanced classes.

\noindent\textbf{DF20}~\cite{picek2021danish} is a fine-grained and long-tailed dataset of $\sim$300k images collected from submissions to the Atlas of Danish Fungi.

\noindent\textbf{iWildCam 2020}~\cite{beery2020iwildcam} is constructed with more than 200k camera trap images of animals, and that year's dataset is supplemented with two new modalities - citizen scientists and remote sensing. Note that in our training process, we do not include the last modality of satellite imagery data.

\noindent\textbf{Tsinghua Dogs}~\cite{zou2020new} has its emphasis on the fine-grained classification of 130 dog breeds, and each breed has at least 200 data. We use the low-resolution version of this dataset.

\noindent\textbf{Places365}~\cite{zhou2017places} is a subset of the Places database, and the Challenge version contains more than 8M scene-centric training images. Note that we train and evaluate on small images (256$\times$256) of this database.

\noindent\textbf{iFashion}~\cite{guo2019imaterialist} is a database with more than 1M fashion images labeled on 228 fine-grained attribute-level classes. Note that this is a multi-label attribute recognition task.

\noindent\textbf{FoodX-251}~\cite{kaur2019foodx} has 251 fine-grained and visually similar food categories paired with 158k web images.

\noindent\textbf{CompCars}~\cite{yang2015large} (Comprehensive Cars) dataset is well-prepared for fine-grained car model classification. We only include the $\sim$137k web-nature data capturing the entire cars, and the associated car make labels with 163 classes.

Some of the datasets do not offer an official validation set, so we apply random splits by ourselves.
Quantitative details of the datasets are summarized in Tab.~\ref{tab:cls-datasets}.

\begin{table*}
\centering \small
\begin{tabular}{l|l|ccc}
\toprule
\multicolumn{2}{c|}{Dataset} & Train & Val. & \# Classes \\
\midrule
\multirow{11}*{\datanamecls}&ImageNet-21K~\cite{deng2009imagenet} & 14M & 50k$^*$ & 21,841 \\
&iNat2021~\cite{van2021benchmarking} & 2.7M & 100k & 10,000 \\
&Herbarium 2021~\cite{de2021herbarium} & 2.2M & 90k$^*$ & 64,500 \\
&DF20~\cite{picek2021danish} & 270k & 30k & 1,604 \\
&iWildCam 2020~\cite{beery2020iwildcam} & 140k & 16k$^*$ & 221 \\
&Tsinghua Dogs~\cite{zou2020new} & 65k & 5.2k & 130 \\
&Places365~\cite{zhou2017places} & 8.0M & 36k & 365 \\
&iFashion~\cite{guo2019imaterialist} & 1.0M & 10k & 228$^\dag$ \\
&FoodX-251~\cite{kaur2019foodx} & 120k & 12k & 251 \\
&CompCars~\cite{yang2015large} & 120k & 14k$^*$ & 163 \\
&Our Newly Collected & 7.5M & 340k & 34,054 \\
\bottomrule
\end{tabular}
\caption{Statistics of image classification datasets for learning \expertcls{}. We combine ten existing datasets and our newly collected part to build the whole dataset, \datanamecls{}. Note that $^*$ suggests that the validation set is obtained by our random split, and $^\dag$ denotes a multi-label classification task which is not included in our evaluation.}
\label{tab:cls-datasets}
\end{table*}

\subsubsection{Implementation Details}

We initialize the models by default with Up-A pretrained checkpoints obtained in Sec.~\ref{sec:stage1}. We apply synchronous SGD optimizer~\cite{goyal2017accurate} with Nesterov momentum~\cite{nesterov1983method} $0.9$ for optimizing the R50 \expertcls{} model, and AdamW optimizer~\cite{kingma2015adam,loshchilov2017decoupled} for training the two \mtb{} models. We fix the weight decay parameter to $10^{-8}$ in both optimizers. We train the two small models, R50 and \mtbfour{}, for $\sim$34 epochs. As for \mtbfifteen{}, we find it prone to overfitting in the classification task, so we halve the number of epochs. We decay the learning rate following a cosine schedule~\cite{loshchilov2016sgdr}. We include RandAugment~\cite{cubuk2020randaugment} and random erasing~\cite{zhong2020random} for data augmentation. We additionally apply stochastic depth~\cite{huang2016deep} to \mtb{} models.

To assess the generality and data efficiency of pretrained models, we provide average accuracy evaluated in the low-data (10\%) regime on our downstream classification benchmark with 20 datasets.

\begin{table*}[!t]
\centering
\subfloat[Up-A vs. \expertcls{}\label{tab:cls-ablation:1vs2}]{%
\tablestyle{4.5pt}{1.1}{\small}
\begin{tabular}{ll|c}
\toprule
 Model & Pretraining & CLS AVG \\
\midrule
 R50 & Up-A & 70.9 \\
 R50 & \expertcls{} & \textbf{73.7} \\ \hline
 \mtbfour{} & Up-A & 74.7 \\
 \mtbfour{} & \expertcls{} & \textbf{78.4} \\ \hline
 \mtbfifteen{} & Up-A & 80.4 \\
 \mtbfifteen{} & \expertcls{} & \textbf{84.2} \\
 \bottomrule
\end{tabular}}\quad
\subfloat[Data scheme\label{tab:cls-ablation:data}]{%
\tablestyle{4.5pt}{1.1}{\small}
\begin{tabular}{l|c}
\toprule
Data & CLS AVG\\
\midrule
ImageNet-21K & 68.8 \\ 
Unified Label Space & 72.9 \\
Natural (Default) & \underline{73.7}\\
Partially Merged & \textbf{73.9}\\ \bottomrule
\multicolumn{2}{c}{~}\\
\multicolumn{2}{c}{~}\\
\end{tabular}}\quad
\subfloat[Initialization\label{tab:cls-ablation:init}]{%
\tablestyle{4.5pt}{1.1}{\small}
\begin{tabular}{l|c}
\toprule
Initialization & CLS AVG \\
\midrule
ImageNet-1k & 73.3 \\
MOCO v2~\cite{chen2020improved} & 73.2 \\
SwAV~\cite{caron2020unsupervised} & 73.1 \\
Ours, Up-A & \textbf{73.7}\\ \bottomrule
\multicolumn{2}{c}{~}\\
\multicolumn{2}{c}{~}\\
\end{tabular}}
\caption{\label{tab:cls-ablations}Experiments and ablation studies on \expertcls{} learning. Results in (b) and (c) are obtained with R50 as the backbone.}
\vspace{-3mm}
\end{table*}

\subsubsection{Ablation Experiments}

As presented in Tab.~\ref{tab:cls-ablations}~(a), our \expertcls{} models steadily surpass their respective Up-A counterparts in terms of downstream classification performance. These results imply that extra supervision during expert training boosts the generalization ability for the classification task. The conclusion stays unchanged when switching the model architecture and scaling up the backbone network. This demonstrates the robustness of our expert learning approach.

\noindent\textbf{Importance of Multiple Data Sources.} Traditional supervised pretraining is generally performed with only one dataset. To prove that multiple datasets are essential for better generalizability, we compare \expertcls{} of our default setting with a baseline model pretrained on ImageNet-21K only. As shown in Tab.~\ref{tab:cls-ablations}~(b), our multi-dataset expert has significantly superior average performance than the baseline (+4.9) on downstream classification datasets. We conclude that leveraging multiple pretraining datasets is crucial for achieving strong downstream generality.

\noindent\textbf{Different Multi-Dataset Learning Schemes.} We further investigate how to make better use of multiple pretraining data sources. Our default method has been described in Sec.~\ref{subsec:stage2_method}, which is to use a shared backbone network and let each dataset have its own classification head, which is essentially a linear layer. Apart from this ``Natural'' approach, we also consider two variants: \textit{Unified Label Space} which merges label spaces of all datasets into a unified one including all of the more than 115K classes, and \textit{Partially Merged} which only merges semantically equivalent classes across different datasets while still preserving multiple label spaces. We list the results in Tab.~\ref{tab:cls-ablations}~(b). We find that completely unifying the pretraining label space has a negative impact on downstream performance (-0.8), while merging equivalent labels leads to a slightly better result (+0.2).

\noindent\textbf{Effect of Initialization.} In this set of experiments, we try different types of pretrained models and select the best one to serve as the initialization for expert training. We include supervised (ImageNet-1k) and self-supervised learning (MOCO v2 and SwAV), as well as our amateur model Up-A. In Tab.~\ref{tab:cls-ablations}~(c), we observe that initializing with our Up-A checkpoint gives the best performance at downstream, validating the effectiveness of our continuous learning paradigm.

\subsubsection{Remarks}

\noindent\textbf{Small Weight Decay.} When learning with multiple data sources, a naturally occurring phenomenon is that training converges much slower than the single-dataset case. One effective trick to mitigate this issue is using near-zero weight decay, which greatly accelerates optimization. As a result, many of our multi-dataset pretraining validation accuracies are close to their single-dataset counterparts if trained with the same number of epochs. For example, on our validation set of ImageNet-21K, our default R50 \expertcls{} has 38.94\% top-1 accuracy, while the same metric of the baseline trained solely on ImageNet-21K is only 0.05\% higher.

\noindent\textbf{Strong Augmentation.} Consistent with \cite{steiner2021train}, we find that applying relatively heavy data augmentation strategies is generally beneficial for both pretraining and downstream performances as long as training converges. Accredited to our optimizer with small weight decay, we can fully train models with RandAugment and random erasing added within our short 34-epoch schedule. Note that including mixup~\cite{zhang2018mixup} and CutMix~\cite{yun2019cutmix} may further boost the model's capabilities, yet more iterations are required for convergence.

\subsection{Object Detection Expert \expertdet{}}
\label{subsec:stage2_detection}

\subsubsection{Datasets}

We include three commonly-used large-scale object detection datasets - Open Images~\cite{kuznetsova2020open}, Objects365~\cite{shao2019objects365} and COCO~\cite{lin2014microsoft}, for pretraining our expert model \expertdet{}. We list details of these datasets as the following.

\noindent\textbf{COCO}~\cite{lin2014microsoft} We follow the standard split of this dataset. More specifically, we train on the train2017 subset with $\sim$118k images, and evaluate on the val2017 set of size 5k.

\noindent\textbf{Objects365}~\cite{shao2019objects365} We use the official training and validation sets, with $\sim$608k and 30k data respectively.

\noindent\textbf{Open Images}~\cite{kuznetsova2020open} We select the 2019 challenge version of this dataset. There are $\sim$1.7 million images as the training set, and another 40k data for validation.

We list details of used detection datasets in Tab.~\ref{tab:det-datasets}.

\begin{table*}
\centering \small
\begin{tabular}{l|l|ccc}
\toprule
\multicolumn{2}{c|}{Dataset} & Train & Val. & \# Classes \\
\midrule
\multirow{4}*{\datanamedet} & COCO~\cite{lin2014microsoft} & 118k & 5k & 80 \\
& Objects365~\cite{shao2019objects365} & 609k & 30k & 365 \\
& Open Images~\cite{kuznetsova2020open} & 1.74M & 41.6k & 500 \\ 
& Our Newly Collected  & 236k & N/A & 809 \\
\bottomrule
\end{tabular}
\caption{\label{tab:det-datasets} Statistics of object detection datasets for learning \expertdet{}. We combine three existing datasets and our newly collected part to build the whole dataset, \datanamedet{}.}
\end{table*}

\begin{table*}[!t]
\centering
\subfloat[Up-A-L vs. \expertdet{}\label{tab:det-ablation:1vs2}]{%
\tablestyle{4.5pt}{1.1}{\scriptsize}
\begin{tabular}{ll|ccc}
\toprule
 Model & Pretraining & VOC07+12 $\uparrow$ & WIDER FACE $\uparrow$ & CityPersons $\downarrow$ \\
\midrule
 R50 & Up-A-L & 76.3 & 90.3/88.3/70.7 & 24.6/59.0 \\
 R50 & \expertdet{} & \textbf{87.7} & \textbf{93.8}/\textbf{92.0}/\textbf{75.5} & \textbf{15.8}/\textbf{41.5} \\ \hline
 \mtbfour{} & Up-A-L & 74.9 & 89.3/87.6/71.4 & 26.5/61.8 \\
 \mtbfour{} & \expertdet{} & \textbf{89.3} & \textbf{94.6}/\textbf{92.6}/\textbf{76.5} & \textbf{14.0}/\textbf{43.8} \\ \hline
 \mtbfifteen{} & Up-A-L & 78.4 & 93.6/91.8/77.2 & 17.7/49.5 \\
 \mtbfifteen{} & \expertdet{} & \textbf{89.4} & \textbf{95.8}/\textbf{94.4}/\textbf{80.1} & \textbf{10.5}/\textbf{42.4} \\
 \bottomrule
\end{tabular}}\quad
\subfloat[Design choice\label{tab:det-ablation:data}]{%
\tablestyle{4.5pt}{1.1}{\scriptsize}
\begin{tabular}{l|ccc}
\toprule
Data & VOC07+12 $\uparrow$ & WIDER FACE $\uparrow$ & CityPersons $\downarrow$ \\
\midrule
Default & \textbf{87.7} & \textbf{93.8}/\textbf{92.0}/75.5 & 15.8/41.5 \\ \hline
Objects365 & 84.0 & 91.4/90.0/74.5 & 18.2/45.2 \\ \hline
Unified Label Space & 87.4 & 93.6/91.8/74.8 & 15.6/42.1 \\
Separate FPN & 87.6 & 93.5/91.7/\textbf{76.5} & \textbf{14.3}/\textbf{39.6}\\ \bottomrule
\multicolumn{2}{c}{~}\\
\multicolumn{2}{c}{~}\\
\end{tabular}}\quad
\subfloat[Initialization\label{tab:det-ablation:init}]{%
\tablestyle{4.5pt}{1.1}{\scriptsize}
\begin{tabular}{l|ccc|ccc}
\toprule
Initialization & COCO $\uparrow$ & Objects365 $\uparrow$ & Open Images $\uparrow$ & VOC07+12 $\uparrow$ & WIDER FACE $\uparrow$ & CityPersons $\downarrow$ \\
\midrule
ImageNet-1k & 46.1 & 25.5 & 63.1 & 87.3 & 93.6/91.9/76.6 & 16.6/44.0 \\
MOCO v2~\cite{chen2020improved} & 47.0 & 25.8 & 62.9 & 87.6 & 93.6/91.8/\textbf{76.6} & 15.8/41.8 \\
SwAV~\cite{caron2020unsupervised} & 44.4 & 24.7 & 62.5 & 86.7 & 92.5/90.2/72.7 & 16.4/45.2 \\
Ours, Up-A-L & \textbf{47.3} & \textbf{26.4} & \textbf{63.4} & \textbf{87.7} & \textbf{93.8}/\textbf{92.0}/75.5 & \textbf{15.8}/\textbf{41.5} \\ \bottomrule
\end{tabular}}
\caption{\label{tab:det-ablations} Experiments and ablation studies on \expertdet{} learning. Results in (b) and (c) are obtained with R50 as the backbone.}
\vspace{-3mm}
\end{table*}

\subsubsection{Implementation Details}

We choose standard Faster R-CNN~\cite{ren2015faster} as the head architecture of our object detection expert \expertdet{}. We further equip \expertdet{} with FPN~\cite{lin2017feature}. As displayed in Fig.~\ref{fig:stage2-overview}, parameters of the FPN are also shared across all datasets.

We use Up-A pretrained checkpoints by default as the initialization. We apply the same type of optimizer as classification expert training. The weight decay is set to $10^{-8}$. All models are trained for $\sim$18 epochs. We decay the learning rate by a factor of $0.1$ at 70\% and 90\% of the training iterations. Only random scaling and horizontal flip are used as the data augmentation. We additionally apply stochastic depth~\cite{huang2016deep} to \mtb{} models.

For validation on pretraining datasets, we report commonly-used metrics, \ie, mmAP for COCO and Objects365, and mAP at IoU $0.5$ for Open Images.
As for downstream evaluation, we provide results in the low-data (10\%) regime on detection datasets of our benchmark.

\subsubsection{Ablation Experiments}

Evaluation results in Tab.~\ref{tab:det-ablations}~(a) show that detection pretraining in the expert (Up-E) stage provides vast improvements on the previous amateur (Up-A) stage in terms of downstream object detection performance. Similar to the case in Tab.~\ref{tab:cls-ablations}~(a) for the classification task, the boost in performance is not affected by changes in the backbone architecture or the model scale.

\textbf{Single-Dataset vs. Multi-Dataset.} To validate our choice of using multiple data sources for expert learning, we compare our default R50 expert \expertdet{} with a baseline model pretrained only on the Objects365 dataset. In Tab.~\ref{tab:det-ablations}~(b), we find that on downstream detection tasks, our default setting outperforms the Objects365 baseline by large margins. This shows that leveraging multiple datasets achieves better generalization at downstream.

\textbf{Parameter Sharing Scheme.} As described in Sec.~\ref{subsec:stage2_method}, our expert method uses a shared backbone along with multiple dataset-specific heads. For the object detection task, a head would contain several parts, more specifically in our case, FPN and Faster R-CNN. Keeping the backbone part being shared, we still have three different sharing schemes concerning the head part: 1) \textit{Separate FPN}, where different datasets have entirely separate heads; 2) \textit{Default}, where the FPN part is shared across all datasets, and different datasets still have independent Faster R-CNN parameters; 3) \textit{Unified Label Space}, where all data sources are merged into one unified dataset, and only one head is required. Results in Tab.~\ref{tab:det-ablations}~(b) show that models corresponding to the three schemes have similar downstream performance, with the default setting and the one using separate heads being slightly better in some metrics.

\textbf{Effect of Initialization.} We also try different types of initialization for the object detection expert. In Tab.~\ref{tab:det-ablations}~(c), we report both pretraining validation and downstream evaluation metrics of the resulting models. We observe that initializing expert learning with our Up-A-L checkpoint gives the best performance at both upstream and downstream. This again demonstrates that our continuous learning paradigm, with amateur followed by the expert, is reasonable.

\subsubsection{Remarks}

Similar to what we find during classification expert training, setting a small weight decay also effectively accelerates the optimization process of the detection expert, and in turn boosts both pretraining and downstream performance. Combined with our multi-dataset setting, our expert models produce highly competitive results on common detection benchmarks. Notably, using the same R50, FPN and Faster R-CNN architecture, a vanilla single-dataset pretraining reported in \cite{shao2019objects365} has 22.5 mmAP on the validation set of Objects365, while our model reaches 26.4 (+3.9 mmAP) as shown in Tab.~\ref{tab:det-ablations}~(c).

\subsection{Semantic Segmentation Expert \expertseg{}}
\label{subsec:stage2_segmentation}

\subsubsection{Datasets}

We train our semantic segmentation expert \expertseg{} on four benchmark datasets: Cityscapes~\cite{cordts2016cityscapes}, ADE20k~\cite{zhou2017scene}, DOTA~\cite{xia2018dota,waqas2019isaid} and COCO-Stuff~\cite{caesar2018coco}, as listed in Tab.~\ref{tab:seg-datasets}. We select these datasets for their popularity, scales and diverse distribution.

\noindent\textbf{Cityscapes}~\cite{cordts2016cityscapes} is a large urban street scene dataset from the car perspective. These images all have a resolution of $2,048\times1,024$, in which each pixel is annotated with pre-defined $19$ classes.  In our experiments, we follow~\cite{chen2017rethinkingdeeplabv3} and crop $769\times769$ images for training and inference on the whole images.

\noindent\textbf{ADE20K}~\cite{zhou2017scene} is a widely-used semantic segmentation dataset of common scenes, covering a broad range of $150$ semantic categories. In our experiments, we crop images of size $513\times513$ for both training and inference.

\noindent\textbf{DOTA}~\cite{xia2018dota} consists of $2,806$ high-resolution remote sensing images. The image sizes range from $800\times800$ to $20,000\times20,000$ pixels. It contains $655,451$ instance annotations over $15$ foreground categories and one background class. In this work, we only use semantic masks for object segmentation. In our experiments, we crop images in the size of $513\times513$ for both training and inference.

\noindent\textbf{COCO-Stuff}~\cite{caesar2018coco} augments all 164k images of the popular COCO~\cite{lin2014microsoft} dataset. It covers $91$ stuff classes and $1$ ``unlabeled'' class. We use $513\times513$ images for both training and evaluation.

\begin{table*}[!t]
\centering
\begin{small}
\begin{tabular}{l|l|rrrc}
\toprule  
\multicolumn{2}{c|}{Dataset}           &         Train      &    Val.      & \#Classes  & Context \\
\midrule  
\multirow{4}*{\datanameseg} &
Cityscapes\cite{cordts2016cityscapes}  &          2,975     &    500       &      19    & Urban Street Scenes \\
& ADE20k\cite{zhou2017scene}             &          20,000    &    2,000     &      150   & Common Scenes \\
& DOTA\cite{xia2018dota}                 &          1,411     &    458       &      15    & Remote Sensing\\
& COCO-Stuff\cite{caesar2018coco}        &          118,000   &    5,000     &      182   & Common Objects \\

\bottomrule
\end{tabular}
\caption{\label{tab:seg-datasets} Statistics of semantic segmentation datasets composing \datanameseg{} for learning \expertseg{}.}
\end{small}
\end{table*}

\subsubsection{Implementation Details}
We use the standard DeepLabv3~\cite{chen2017rethinkingdeeplabv3} structure for our semantic segmentation expert model, and follow an open-source PyTorch implementation\footnote{\url{https://github.com/chenxi116/DeepLabv3.pytorch}}. 
We use the setting with output stride $16$, \ie modifying the stride of the last stage to $1$. For R50 models, we replace all subsequent layers with atrous convolutional layers with atrous rate set to $2$.

In all of our semantic segmentation experiments, we use consistent augmentation strategies that are used in \cite{chen2017rethinkingdeeplabv3}: random horizontal flipping and random resizing with a scale factor in range ($0.5$, $2.0$) are applied before cropping images with a fixed size during training. During inference, we do not use any multi-crop strategy for simplicity. We evaluate the performance of pretrained models using \textit{mIoU}, \ie, mean Intersection over Union between prediction and ground truth maps over all classes. As for optimizer, we use SGD with momentum $0.9$ and learning rate $0.01$ for R50, and AdamW with learning rate $6\times10^{-5}$, weight decay $0.01$ for \mtbfour{}. Both optimizers follow a polynomial learning rate decay: the initial learning rate is multiplied by $\left(1-\frac{\text{current\_iter}}{\text{max\_iter}}\right)^{p}$ with $p = 0.9$. 
Besides, we apply gradient clipping with a maximum norm of $4$ since we find the optimization can diverge without it.

\begin{table*}[h]
\centering
\begin{small}
\begin{tabular}{lll | cc  cc | c}
\toprule									
Model	& Initialization & Train Data &	Cityscapes	&	ADE20K	&	DOTA	&	COCO-Stuff	&   VOC2012 \\
\midrule								
R50 & ImageNet-1k & Cityscapes &	\textbf{0.759}	&	-	&	-	&	-	& 0.123 \\
R50	& Up-A & Cityscapes &	0.751	&	-	&	-	&	-	& \underline{0.155} \\
\hline									
R50 & ImageNet-1k & ADE20K &	-	&	0.400	&	-	&	-	& 0.469 \\
R50	& Up-A & ADE20K &	-	&	\underline{0.410}	&	-	&	-	& \underline{0.504} \\
\hline									
R50 & ImageNet-1k & DOTA &	-	&	-	&	0.602	&	-	& \underline{0.231} \\
R50	& Up-A & DOTA &	-	&	-	&	\underline{0.615}	&	-	& 0.208 \\
\hline									
R50 & ImageNet-1k & COCO-Stuff &	-	&	-	&	-	&	0.387	& \underline{0.677} \\
R50 & Up-A & COCO-Stuff &	-	&	-	&	-	&	\underline{0.388}	& 0.673 \\
\midrule
\midrule
R50 & ImageNet-1k & \datanameseg{} &	0.726	&	0.401	&	0.577	&	0.387	& 0.692 \\
R50	& Up-A & \datanameseg{} &	\underline{0.746}&	\underline{0.426}	&	\underline{0.580}	&	\underline{0.397}	& \underline{0.696} \\
\hline
\mtbfour{} & Up-A & \datanameseg{} &	0.737	&	\textbf{0.470}	&	\textbf{0.649}	&	\textbf{0.446}	& \textbf{0.778} \\
\bottomrule
\end{tabular}
\caption{\label{tab:seg-upstream} Experiments and ablation studies on \expertseg{} learning. The last two rows correspond to two \expertseg{} models under the default setting, with R50 and \mtbfour{} as the backbone, respectively. Note that \underline{underline} style denotes the best result within each box, and \textbf{bold} style denotes the best result in this table.}
\end{small}
\end{table*}

\subsubsection{Ablation Experiments}

\noindent\textbf{Pretraining Data.}
Tab.~\ref{tab:seg-upstream} compares our multi-dataset expert learning scheme with single-dataset baselines. We present both upstream validation and downstream transfer (VOC2012) results. Although our \expertseg{} R50 model fails to match its single-dataset counterpart in two of the pretraining datasets, the expert still has much better generalization ability at downstream than all the baselines. This implies that a shared representation across multiple segmentation datasets possesses strong generality.

\noindent\textbf{Initialization.} As shown in Tab.~\ref{tab:seg-upstream}, in most cases, leveraging Up-A models as the initialization for expert learning achieves better upstream and downstream results compared to using an ImageNet-1k supervised checkpoint. This again demonstrates the effectiveness of the visual representation learned with our Up-A stage.

\subsubsection{Remarks}
 
In our pretraining experiments, we find that adding more weight decay to the segmentation heads can improve the downstream result (see Tab.~\ref{tab:seg-downstream-r50-AW}), yet too much such weight decay instead hurts the transfer performance. We hypothesize that moderately increasing the weight decay of heads forces the shared backbone to learn more universal representation that works for all datasets, which is more beneficial for downstream transfer.

\begin{table*}[t]
\centering
\begin{tabular}{l | c}
\toprule
Model		&	VOC2012	\\
\midrule
Up-E (S) R50	&	0.696	\\
 + head weight decay $10^{-4}$	&0.703	\\
 + head weight decay $10^{-3}$	&\textbf{0.719}	\\
 + head weight decay $10^{-2}$	&	0.633	\\
\bottomrule													
\end{tabular}
\caption{\label{tab:seg-downstream-r50-AW} {Downstream performance of adding more weight decay to the semantic segmentation heads during expert learning. }}
\end{table*}

\def\expert{{\em expert}}
\def\generalist{{\em generalist}}

\section{Pretraining Up-G Stage: Making a Generalist from Experts}
\label{sec:stage3}

\begin{figure}[t]
	\centering
	\includegraphics[width=1.0\linewidth]{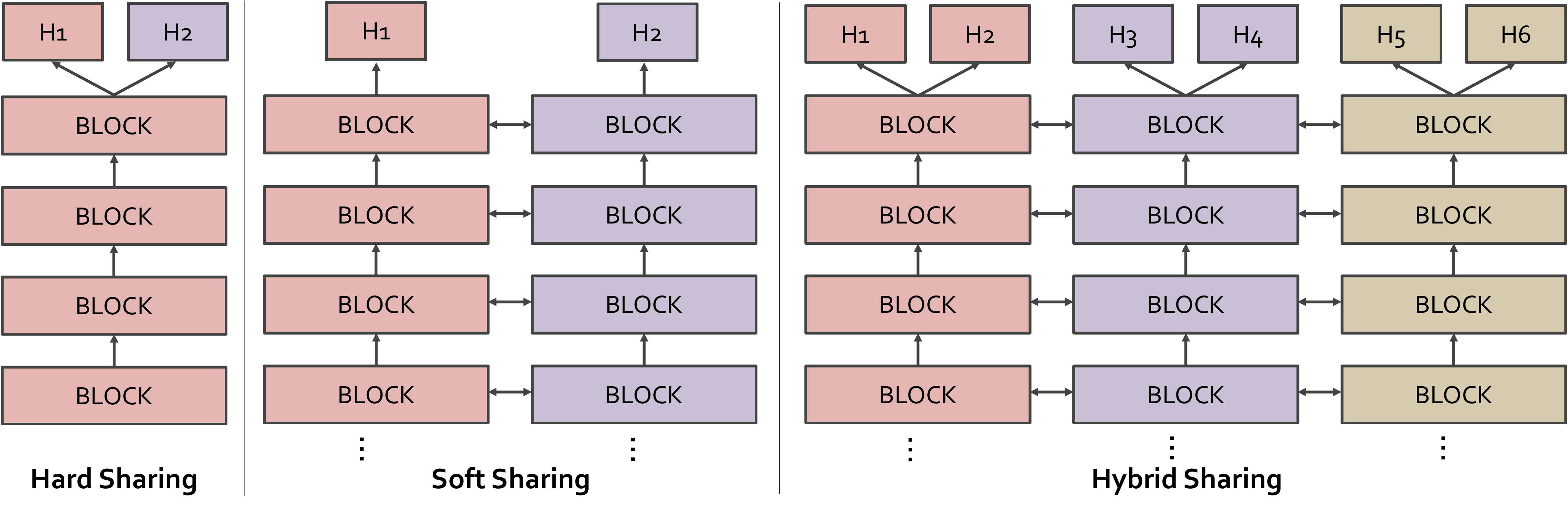}
	\caption{From soft parameter sharing paradigm to {\em hybrid} sharing. {\em Generalist} is easily extended from {\em experts} with the performance maintained.}
	\label{fig:stage3-relate-architecture}
\end{figure}

With well-performing {\em experts} in a wide spectrum of vision problems, we aim to build a unified model producing a general-purpose representation to achieve stronger and more generalizable performance in different tasks. Existing researches show that model (\expert{}) performance trained in multiple tasks could be further improved by fusing their features to a shared representation, as the knowledge captured by {\em experts} are mutually related~\cite{han2017heterogeneous,sermanet2013overfeat,maninis2019attentive, zhao2018modulation, guo2019depthwise,Simple_multi_dataset_detection}. Generally, the aforementioned multiple tasks refer to one vision problem (\eg classification) with different datasets (\eg ImageNet and CIFAR), or multiple vision problems (\eg classification and detection) with one dataset. In our framework, {\em experts} simultaneously specialize in multiple vision problems as well as multiple datasets due to their learning process.
How to integrate our {\em experts} to a unified model is a novel and more generalized setting compared to previous works on multi-task learning. 
Therefore, after the pretraining Up-E stage, we propose Up-G as the third pretraining stage to further unify the feature representation. We propose a new paradigm, named {\em hybrid} parameter sharing (Fig.~\ref{fig:stage3-relate-architecture} right), to develop a versatile model named {\em generalist} from {\em experts}. By leveraging both soft (Fig.~\ref{fig:stage3-relate-architecture} middle) and hard sharing (Fig.~\ref{fig:stage3-relate-architecture} left) methods, this paradigm can transfer information between experts without introducing task conflicts. To the best of our knowledge, this is a first attempt to systematically study large-scale multi-task learning in the setting with multiple vision problems and datasets.

The remaining of this section is organized as follows. Firstly, we illustrate our proposed paradigm with detailed analysis. We then conduct extensive experiments to verify our designs.

\subsection{Method}
\label{subsec:stage3_approach}

Our method aims to enhance \expert{} features by exchanging information learned from various tasks. We expect that each \expert{} broadens its vision from other tasks without compromising the performance on its own task. To reach this goal, we propose a new {\em hybrid} sharing paradigm to unify {\em experts} into a \generalist{}.

Our proposed paradigm involves both {\em soft} and {\em hard} sharing approaches. Specifically, we inherit the hard sharing approach employed in the previous Up-E stage, where in each \expert{} different datasets share one unified representation from a single backbone. We further incorporate soft sharing for feature transfer among {\em experts}. This is achieved by introducing {\em knowledge transfer modules} to conduct feature exchange at each stage of the {\em experts}. The purpose is to let the additional tasks assist one main task for more diverse knowledge.

\label{subsubsec:stage3_why}

With the hybrid sharing paradigm, we build \generalist{} from {\em experts}. In structure, \generalist{} is an interconnected version of all {\em experts}, we call the backbone of each \expert{} as a branch of \generalist{}. Further, we categorize each branch in \generalist{} into image-wise, patch-wise, and pixel-wise according to the task on which the corresponding \expert{} is trained.
We denote the outputs from four stages of an \expert{} by $ \{ C2, C3, C4, C5 \}$ and the corresponding feature maps from the FPN~\cite{lin2017feature} for patch-wise tasks as $\{P2, P3, P4, P5 \}$.

\subsubsection{Cross-Task Knowledge Transfer with Soft Sharing}
\label{sec:soft-sharing}
To build a set of generalizable representations, we connect all branches in \generalist{} using the knowledge transfer modules. From the viewpoint of each such module, there are a main branch corresponding to one task, and auxiliary branches of the other tasks. The non-linear knowledge transfer modules would receive features from auxiliary branches and fuse them into the main branch. More specifically, a knowledge transfer module at stage $d$ collects features of stages $1,2,\cdots,d$ from the auxiliary branch, and merge them into the main branch feature.
During training, we detach the gradients at the input side of the knowledge transfer modules to avoid cross-task interference among the {\em experts}.

\begin{figure}[t]
	\centering
	\includegraphics[width=0.6\linewidth]{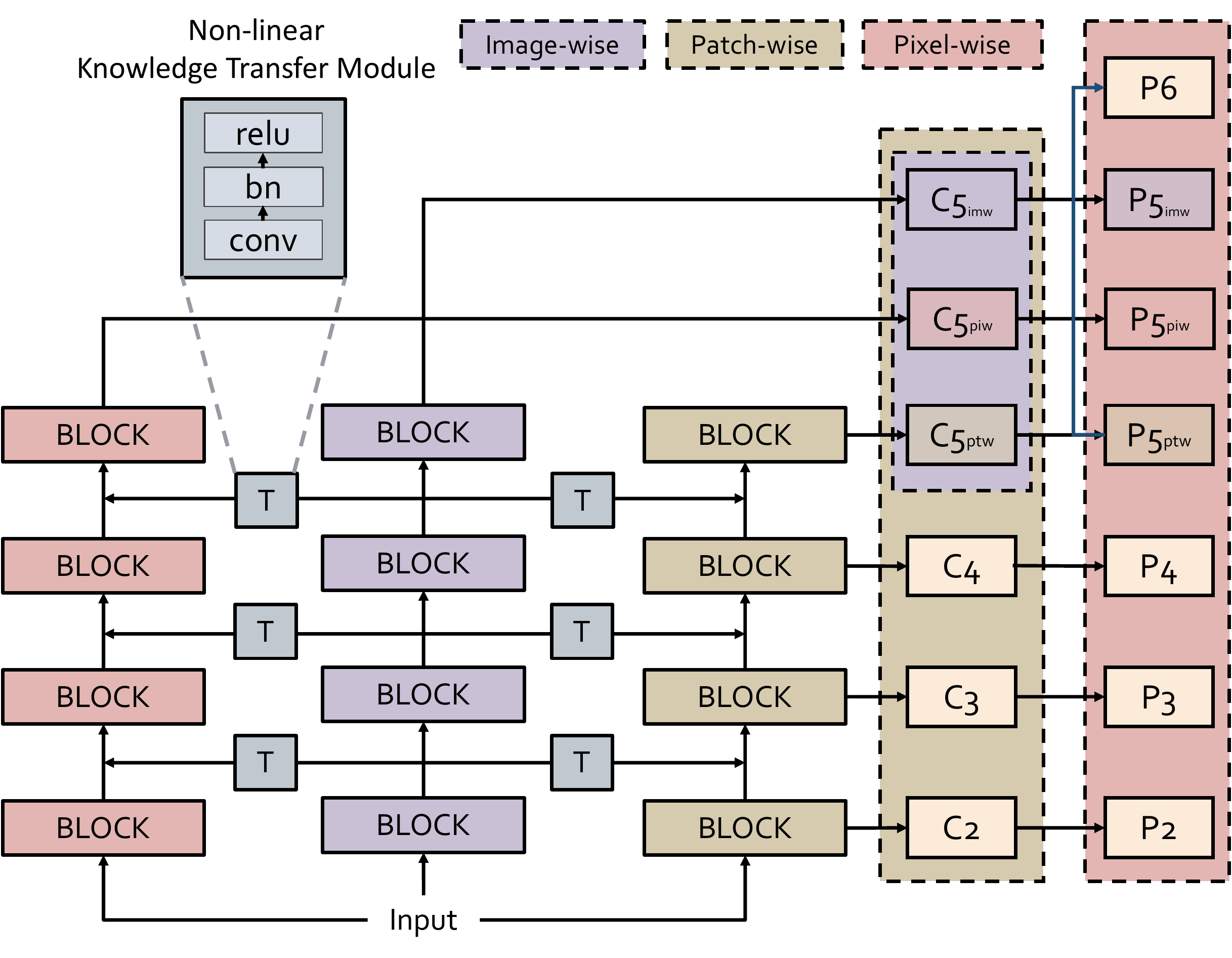}
	
	\caption{\textbf{Versatile representation learning.} {\em Generalist} extracts different types of unified representations to adapt to task requirements. Different types of $C5$ are concatenated in the feature dimension and generate the corresponding $P5$ by the FPN. Note that the knowledge transfer module also has cross-layer connections which are not visualized in this figure, passing auxiliary features from shallow layers to deep stages of the main branch.}
	\label{fig:stage3-unified-feature}
\end{figure}

\subsubsection{Versatile Representation Learning with Hard Sharing}
As shown in Fig.~\ref{fig:stage3-unified-feature}, to handle different tasks, the \generalist{} produces different types of unified representations, \eg $C5$ for image-wise tasks and lower-level features for patch-wise and pixel-wise ones. Specifically, the employed unified representations are designed as follows:

\begin{itemize}
\item Representation for image-wise prediction. Early researches~\cite{gu2018recent} reveal features extracted from the deep layers of a neural net contain more high-level representations than those from the shallow layers. In this regard, deep features cater to image-level prediction tasks like classification. From this perspective, we fuse $C5$ features from all branches of the \generalist{}, \ie $\{C5_\text{imw},C5_\text{ptw},C5_\text{piw}\}$, for image-level prediction. For experiments without a pixel-wise branch, we only integrate two $C5$ features, $\{C5_\text{imw},C5_\text{ptw}\}$.

\item Representation for patch-wise prediction. We utilize a feature pyramid design to cope with prediction tasks on boxes of different sizes, like detection.
We use $\{ P2, P3, P4, P5, P6 \}$ (cyan) after the FPN for such box-wise prediction. Note that $ C5_\text{ptw} $ still generates the $ P6 $ with a convolutional layer, while $\{C5_\text{imw},  C5_\text{ptw}, C5_\text{piw} \}$ produces $ \{ P5_\text{imw},  P5_\text{ptw}, P5_\text{piw} \} $, respectively.

\item Representation for pixel-wise prediction. In pixel-level tasks like segmentation and depth estimation, we assign $ \{ C2, C3, C4 \}$ of the pixel-wise branch and the fused version of $ C5 $ as the features for prediction. Note that if there is no pixel-wise branch, we use $\{C2,C3,C4\}$ of the patch-wise branch instead.
\end{itemize}

Different from the soft sharing strategy used in cross-task knowledge transfer, these final output representations from the \generalist{} are unified in a hard sharing manner. Similar to the Up-E stage, we employ the multi-head design on top of the unified feature maps, where each head is a dataset-specific and task-specific sub-network.

\subsection{Experiments}

\subsubsection{Implementation Details }
\label{subsubsec:stage3_model_training}

We use the same datasets as in Sec.~\ref{sec:stage2}. The difference is that all datasets are used simultaneously when learning the \generalist{}. Data preprocessing is consistent with Sec.~\ref{sec:stage2}. The architectures of the heads on top of \generalist{} features are also the same as the ones used for {\em expert} training, \eg Faster R-CNN~\cite{ren2015faster} for object detection and Deeplab v3~\cite{chen2017rethinkingdeeplabv3} for semantic segmentation.

To train our \generalist{}, we use SGD with momentum $0.9$ and learning rate $10^{-4}$ for R50, and AdamW~\cite{loshchilov2017decoupled} with learning rate $4\times10^{-6}$, for MN-B4 and MN-B15. The weight decay is set to $10^{-8}$ for both optimizers. We use different schedulers on learning rate for different branches. Same as Sec.~\ref{sec:stage2}, the learning rate of image-wise branches follows a cosine curve with the minimum learning rate being $0.0$, the optimizer of patch-wise tasks is applied with a ``multistep'' scheduler decaying the learning rate by a factor of $10\%$ when reaching 70\% and 90\% of total training steps, while pixel-wise tasks uses a polynomial decay on the learning rate with $0.9$ as the power. We apply $2,500$ linear warm-up~\cite{goyal2017accurate} steps when training the \generalist{}.

Batch normalization (BN) statistics are important for tasks with a small batch size, like object detection. We apply task-specific synchronized BN within each branch. As a result, parameters are still fully synchronized, while the BN statistics can be different for different tasks.

As mentioned in Sec.~\ref{sec:soft-sharing}, the auxiliary branches would not receive gradients from the corresponding knowledge transfer module to avoid task conflicts.

\subsubsection{Quantitative Evaluation}

\begin{table*}[ht]
    \renewcommand\arraystretch{1.3}
    \small
	\centering
	\resizebox{\linewidth}{!}{
	\begin{tabular}{@{}l|l|c|cll|c|cc@{}}
		\toprule
		\multicolumn{2}{c|}{\multirow{2}{*}{Model}} & \multicolumn{1}{c|}{CLS } & \multicolumn{3}{c|}{DET} & \multicolumn{1}{c|}{SEG} & \multicolumn{2}{c}{DEP} \\ \cline{3-9}
		\multicolumn{2}{c|}{} & \multicolumn{1}{l|}{CLS AVG~$\uparrow$} & \multicolumn{1}{l}{VOC07$+$12~$\uparrow$} & WIDER FACE~$\uparrow$ & CityPersons~$\downarrow$& \multicolumn{1}{l|}{VOC2012~$\uparrow$} & \multicolumn{1}{l}{KITTI~$\downarrow$} & \multicolumn{1}{l}{NYUv2~$\downarrow$} \\ \midrule
		\multirow{4}{*}{R50} & Up-E (C) & \underline{73.7}& 72.2& \multicolumn{1}{c}{89.7/87.6/68.1}& \multicolumn{1}{c|}{22.4/58.3} & 57.7& 3.214& 0.501\\
		\multicolumn{1}{c|}{} & Up-E (D) & 53.9& \underline{87.7}& \multicolumn{1}{c}{\underline{93.8/92.0/75.5}} & \multicolumn{1}{c|}{\underline{15.8/41.5}}& 62.3& 3.087& 0.449\\
		\multicolumn{1}{c|}{} & Up-E (S) & 47.5& 75.0   & \multicolumn{1}{c}{87.4/85.7/66.4}       & \multicolumn{1}{c|}{19.6/53.3}  & \underline{71.9} & 3.116& 0.454\\
		\multicolumn{1}{c|}{} & Up-G (C-D) & \textbf{74.3}& \textbf{87.7}& \multicolumn{1}{c}{\textbf{93.9/92.2/77.0}}& \multicolumn{1}{c|}{\textbf{14.7}/46.0} & 66.2& \textbf{2.835}& \textbf{0.391}\\ \midrule
		\multirow{4}{*}{MN-B4}
		& Up-E (C) & \underline{78.4} & 73.7    & \multicolumn{1}{c}{89.6/88.0/71.1}& \multicolumn{1}{c|}{30.2/65.0}  & 65.8   & 3.544& 0.459\\
		& Up-E (D) & 59.2    & \underline{89.3} & \multicolumn{1}{c}{\underline{94.6/92.6/76.5}}& \multicolumn{1}{c|}{\underline{14.0/43.8}}  & 73.1   & 3.048& 0.397\\
		& Up-E (S) & 62.3    & 78.7    & \multicolumn{1}{c}{89.5/87.9/71.4}& \multicolumn{1}{c|}{19.4/53.0}  & \underline{79.6}& 3.061& 0.414\\
		& Up-G (C-D) & \textbf{78.6}    & 89.1    & \multicolumn{1}{c}{\textbf{94.9/92.8/76.5}}& \multicolumn{1}{c|}{\textbf{12.0}/50.5}  & 72.2   & \textbf{2.944}& \textbf{0.397}\\ \midrule
		\multirow{3}{*}{MN-B15}  
		& Up-E (C) & \underline{84.2} & 80.4    & \multicolumn{1}{c}{93.2/91.4/75.7}& \multicolumn{1}{c|}{29.5/59.9}  & 70.6   & \underline{2.633}   & 0.369\\
		& Up-E (D) & 60.9    & \underline{89.4} & \multicolumn{1}{c}{\underline{95.8/94.4/80.1}}& \multicolumn{1}{c|}{\underline{10.5/42.4}} & 77.2 & 2.723 & 0.371\\
		& Up-G (C-D) & \textbf{84.4}    & \textbf{89.8}    & \multicolumn{1}{c}{\textbf{95.9/94.2/78.8}}& \multicolumn{1}{c|}{\textbf{10.5/41.3}}  & \textbf{77.3}   & 2.708& \textbf{0.365}\\ \bottomrule
	\end{tabular}}
	\caption{Comparison between \generalist{} and \expert{} models.}
	\label{tab:s3_compaired_to_s2}
\end{table*}

Tab.~\ref{tab:s3_compaired_to_s2} shows the performance of our \generalist{} (Up-G (C-D), incorporating the classification \expert{} Up-E (C) and the detection one Up-E (D)) on image classification (CLS), detection (DET), segmentation (SEG), and depth estimation (DEP) tasks. We observe that \generalist{} beats Up-E (C) and Up-E (D) or closely matches their performance in all tasks. It validates the effectiveness of our hybrid sharing paradigm that the \generalist{} can produce better representations than its \expert{} components in most cases. Exceptions exist as MN-B4 Up-E (D) shows higher than Up-G in SEG, and MN-B15 Up-E (C) yields lower error in KITTI than Up-G. These examples show that our proposed approach is still not a silver bullet. How to let the \generalist{} has uniformly stronger generalization ability than its sub-nets (\expert{}) in both seen and unseen tasks remains an open research topic.

On the segmentation task, Up-E (S) outperforms Up-G with both R50 and MN-B4 backbones. This can be mainly attributed to the fact that the segmentation expert is not merged into this version of Up-G. It suggests that although the \generalist{} is able to push up the upper limits of its contained {\em experts}, it still cannot surpass the performance an \expert{} who specializes in a new unseen task.

Interestingly, on depth estimation with R50 and MN-B4, Up-G gives better results than all {\em experts}. It reveals the great potential in the representation combining classification and detection features for depth estimation. We also notice that Up-E (D) works better than Up-E (S) on depth estimation. It is a bit counter-intuitive as we usually suppose semantic segmentation requires pixel-level understanding and is more similar to depth estimation, while object detection asks for a relatively coarse-level regional understanding. We conjecture that this phenomenon results from the distinctive annotation misalignment between semantic segmentation and depth estimation. Take a street as an example. It is labeled as the same class in all its pixels for semantic segmentation, but with spatially notable variant values for depth estimation as different pixels locate at different distances to the camera.

\label{subsubsec:stag3_visualization}
\begin{figure}[t]
	\centering
	\includegraphics[width=0.8\linewidth]{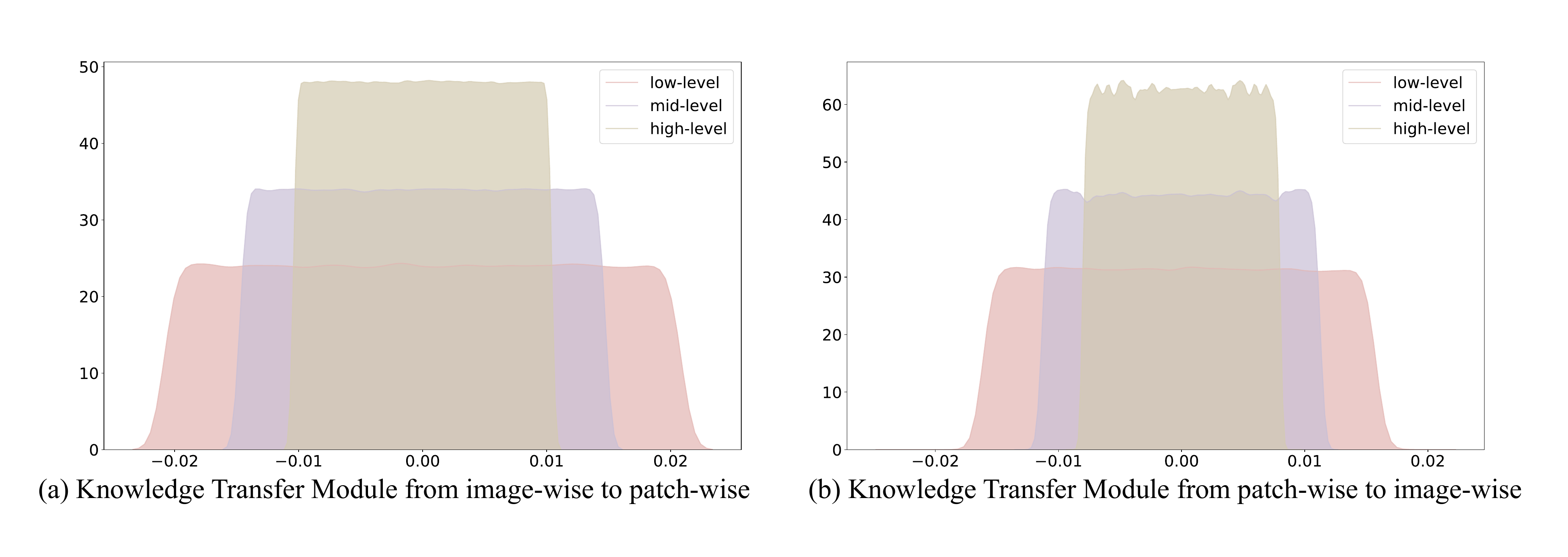}
	\caption{Visualization of response distribution from {\em knowledge transfer modules}. To compute the distribution, we collect responses from pretraining validation samples.}
	\label{fig:stage3-weight}
\end{figure}

\begin{table*}[h]
	\centering
	\subfloat[Position\label{tab:tab:s3-position}]{%
    \tablestyle{4.5pt}{1.1}{\small}
	\begin{tabular}{l|cc}
		\toprule
		Model  & CLS AVG & VOC07+12  \\ \midrule 
		Hard Sharing & 73.5 & 82.5 \\
		Same-level & 73.8 & 87.7 \\
		Low-level & 74.3 & 87.7   \\
		High-level & 74.0 & 87.4 \\
		Cross-level & \textbf{74.3} & \textbf{87.7} \\
		\bottomrule  
        \multicolumn{2}{c}{~}\\
        \multicolumn{2}{c}{~}\\
	\end{tabular}}\quad
	\subfloat[Choice of Architecture\label{tab:tab:s3-choice}]{%
    \tablestyle{4.5pt}{1.1}{\small}
	\begin{tabular}{l|cc}
		\toprule
		Method & CLS AVG & VOC07+12  \\ \midrule 
		Scalable & 73.8 & 87.1  \\ 
		Non-Linear & 73.8 & 87.7 \\ 
		Channel & 73.9 & 87.7 \\
		Attention & 73.7 & 87.2 \\
		Policy & 73.9 & 87.4 \\
		Gating & 73.5 & 87.0  \\ 
		NDDR & 72.4 & 86.9 \\
		\bottomrule
    \end{tabular}}
	\caption{\label{tab:s3-position-choice} Ablation studies on the position of knowledge transfer modules and the choice for their architecture.}
\end{table*}

\subsubsection{Ablation Studies}
\label{subsec:stage3-ablation}
\noindent\textbf{Position of Knowledge Transfer Modules.} 
We investigate how the knowledge transfer module works in our method, including its necessity and position. We find that compared with a pure hard-sharing baseline, our hybrid sharing scheme with knowledge transfer modules has clear advantage in downstream classification and detection performance.
We also conduct ablation experiments on different connection types: 1) ``Same-level'', which means knowledge transfer modules only merge features of all branches within the same stage without any cross-level connection; 2) ``Low-level'', in which knowledge transfer modules only interconnect at shallow stages; 3) ``High-level'', in which they are applied only at deep stages; 4) ``Cross-level'', which is the way described in Sec.~\ref{sec:soft-sharing} and used by previous experiments. From Tab.~\ref{tab:s3-position-choice} (a), we can see that the cross-level connection approach works best in terms of downstream transfer performance, and only connecting at low levels achieves the most similar performance. This empirically suggests that exchanging low-level features across different tasks can be beneficial for multi-task learning.

In addition, we visualize responses from knowledge transfer modules at different positions of the \generalist{}. By response we refer to the scale of a module output relative to the main branch feature value at the same position, so a response with a large absolute value means significant impact from the auxiliary branch on the main branch. We also denote the features from the shallow layers of the network as low-level features, and those from the deep layers as high-level features. Fig.~\ref{fig:stage3-weight} depicts the distribution of responses from knowledge transfer modules between the image-wise and patch-wise branches, from low-level to high-level. We observe that in both directions, the distributions are mostly centered around 0, yet the low-level responses would have a wider range than the high-level ones. This suggests that more information is transferred at shallow layers than deep layers, which also supports the experimental results in Tab.~\ref{tab:s3-position-choice} (a) that low-level knowledge transfer modules are more effective than high-level ones.

\noindent\textbf{Choices for Knowledge Transfer Module.}
\label{sec:choices_of_translayer}
We explore popular designs for our knowledge transfer modules, including NDDR~\cite{gao2019nddr}. The performances of different modules on downstream classification and detection tasks are listed in Tab.~\ref{tab:s3-position-choice} (b). We define ``Scalable''~\cite{ma2019snr}, ``Non-Linear'', ``Channel''~\cite{hu2018squeeze} and ``Attention''~\cite{liu2019end} as active transfer modules, and their weights are calculated by themselves from the input. We can observe that the results of these modules are relatively similar, and they all do well in classification transfer tasks. On the other hand, the weights of ``Policy''~\cite{devin2017learning}, ``Gating''~\cite{purushwalkam2019task} and ``NDDR'' are predicted by other modules. We find that the average performance of these three choices is not as good as previous designs. In other experiments, we select the ``Non-Linear'' design as the default setting.

\begin{table*}[]
    \renewcommand\arraystretch{1.2}
	\centering
	\footnotesize
	\begin{tabular}{@{}l|c|c|c|cc@{}}
		\toprule
		Model & \multicolumn{1}{l|}{CLS AVG $\uparrow$ } & \multicolumn{1}{l|}{VOC07$+$12 $\uparrow$} & \multicolumn{1}{l|}{VOC2012 $\uparrow$ } & \multicolumn{1}{l}{KITTI $\downarrow$} & \multicolumn{1}{l}{NYUv2 $\downarrow$} \\ \midrule
		Up-E (C) & 73.7 & 72.2 &  57.7& 3.214& 0.501\\
		Up-E (D) & 53.9& \underline{87.7}& 62.3& 3.087& 0.449\\
		Up-E (S) & 47.5& 75.0   &\underline{71.9} & 3.116& 0.454\\
		Up-G (C-D) & \underline{74.3} & \underline{87.7}   & 66.2 & \underline{2.835} & \underline{0.391}\\ \hline
		Up-G (C-D-S) & \textbf{74.3}& \textbf{87.7}& \textbf{73.8}& \textbf{2.797}& \textbf{0.391}\\ 
		\bottomrule  
	\end{tabular}
	\caption{Adding the pixel-wise \expert{} Up-E (S) to \generalist{}, resulting in Up-G (C-D-S). Note that the \underline{underline} style denotes the best result within each box and the \textbf{bold} style denotes the best result in this table.}
	\label{tab:s3-compare-s2-seg}
\end{table*}

\noindent\textbf{Expansibility of Generalist.} To demonstrate the extensibility of our paradigm, we add the pixel-wise \expert{} model Up-E (S) to the \generalist{}, which results in the same pipeline as depicted in Fig.~\ref{fig:stage3-unified-feature}. With our hybrid sharing scheme, adding such a new branch is straightforward, just connecting the new branch with extra knowledge transfer modules, and then adjust the final unified feature output. From results in Tab.~\ref{tab:s3-compare-s2-seg}, we can see a significant increase in downstream segmentation performance achieved by adding the pixel-wise branch. This is consistent with our intuition that the \generalist{} inherits expertise from its \expert{} components. Notably, the final \generalist{} still has strong generalization ability in other tasks, such as depth estimation.

\section{Down-A Stage: Transferring Knowledge to Various Downstream Tasks}
\label{sec:transfer}

Pretrained models in Up-G stage possess general capabilities in multiple tasks and domains, but when it comes to a specific downstream task, only part of the knowledge of pretrained models is needed, and the knowledge should also be adapted to the downstream task to achieve better performance, \ie transfer learning. Especially when only a few data are available for transferring, the problem should be paid more attention. 
Many of recent top-performing transfer learning methods~\cite{houlsby2019parameter,guo2019spottune,guo2020adafilter,ro2021autolr} indeed obtain prominent improvements.
However, these methods neither leverage the implicated information in the upstream pretraining nor consider the insufficiency of downstream data in few-shot scenarios.
In contrast, we propose a Multi-stage Fine-tuning (MF) method in this Down-A stage, hoping for alleviating the difficulty of transferring with few data.
By encoding upstream data in a generative model, \ie VQ-GAN~\cite{esser2021taming}, we are able to transfer pretrained model to multiple tasks and domains without using upstream data every time, which makes our method more general and extensible.

The remaining of this section is organized as follows. First, we introduce our proposed MF method and the supervision collapse problem. Then, we perform extensive experiments to validate our method.

\subsection{Multi-stage Fine-tuning}
\label{sec:irf}

\begin{figure}[t]
  \begin{center}
   \includegraphics[width=0.8\linewidth]{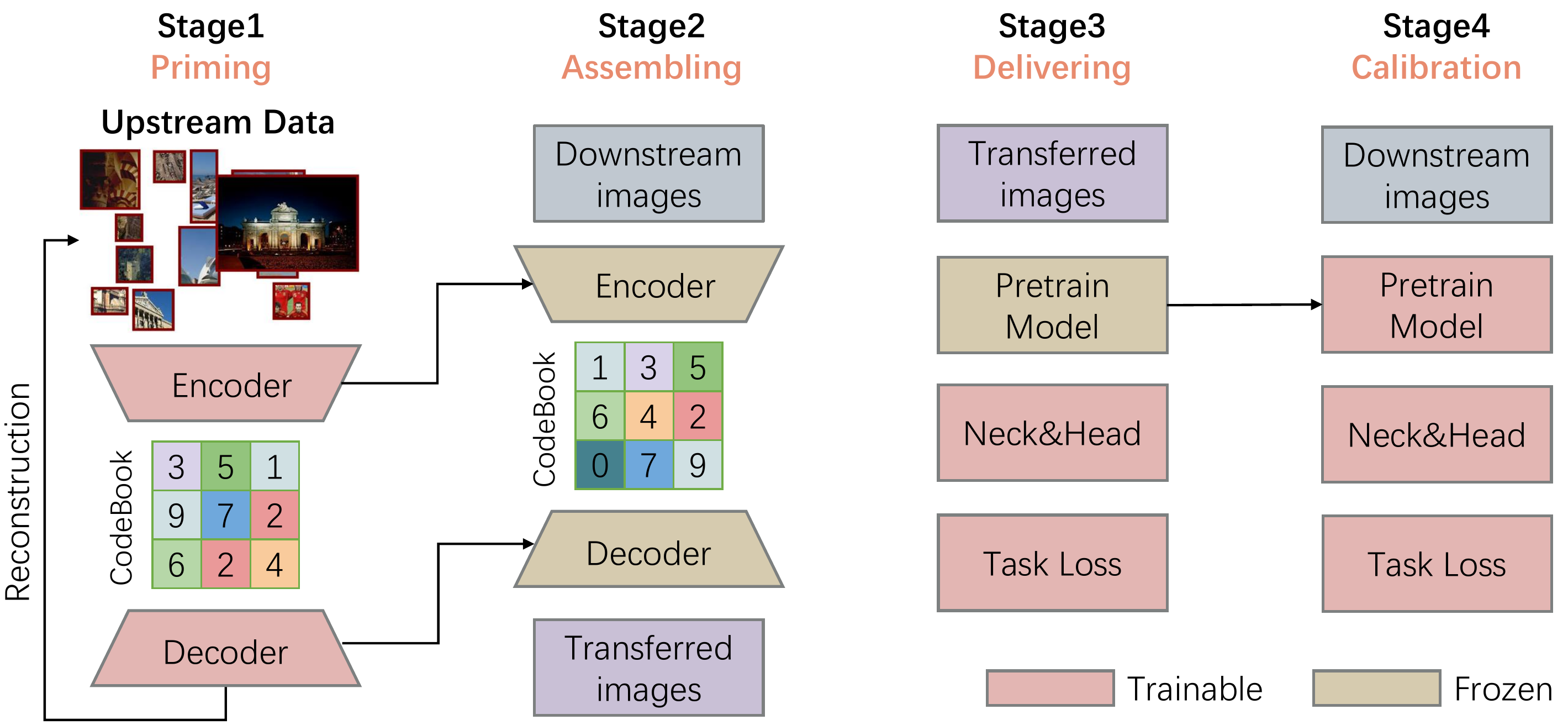}
    \end{center}
    \caption{Overview of Multi-stage Fine-tuning (MF). A VQ-GAN model is first trained with upstream data in stage 1. Then downstream data is reconstructed by the it in stage 2. After that, only the new added task specific parameters are trained by re-represented images in stage 3 and the whole model is fine-tuned by downstream data in stage 4.}
\label{fig:transfer-vp-pipeline}
\end{figure}

We use upstream data as the upstream information source by encoding their information into a VQ-GAN~\cite{esser2021taming} to decouple its dependency on the upstream model or training strategy.
When transferring, downstream data are first reconstructed by the VQ-GAN model and are used to train the newly added parameters while keeping pretrained parameters fixed.
The intuition behind this step is that the more similarity between downstream data and upstream data, the more reliable and informative feature the pretrained model can extract.
After that, original downstream data will be used to fine-tune the whole model for further global optimization.
As shown in Fig.~\ref{fig:transfer-vp-pipeline}, the Multi-stage Fine-tuning (MF) method can be roughly divided into four stages.

\textbf{Stage 1.} In this stage, we encode information in upstream data into a VQ-GAN, which will encode domain information of upstream data implicitly. The VQ-GAN is independent of the upstream pretrained model, so its architecture is not needed to keep the same as the pretrained model.
Besides, it only needs to be trained once, and can be reused by any downstream task.
So it both exploits upstream information and decouples the following stage with the pretrained model.

\textbf{Stage 2.} In this stage, we will reconstruct downstream data by the VQ-GAN trained in the first stage, which can be seen as a re-representation process. Note that the VQ-GAN is fixed in this stage, otherwise, it will lose the encoded upstream information.
We suppose a model can extract a reliable feature, \ie a feature that can express the image's information precisely and sufficiently, if an image is in a similar domain with its training data.
Then after the re-representation, features of transferred images extracted by the pretrained model will be more reliable and informative.

\textbf{Stage 3.} Different downstream tasks need different task-specific layers to fulfill their targets which is hard to be provided by the pretrained model, \eg classifier for classification tasks, neck and head for detection tasks, etc.
How to optimize them into a proper state for further optimization is important, especially in few-shot settings.
In this stage, we fix the pretrained model and only optimize the newly added layers with transferred images obtained in stage 2.
Using transferred images and fixing the pretrained model can keep the feature reliable after the extraction by pretrained model, so successive parts will get high-quality inputs and will be optimized into better local minima.

\textbf{Stage 4.} After the first three-stage, the model is already in a not bad state, but it still lacks global optimization. Besides, if we stopped at stage 3, when inference the model, the data should also be transferred, which is not proper for further usage. So we calibrate the domain into the downstream domain thoroughly in this stage. Original downstream images are used and all parameters are optimized in this stage.

\subsection{Supervision Collapse}

We suppose the features provided by the pretrained model have encoded sufficient information, \ie for any downstream task, most of the task-related features have already been extracted by the pretrained model.
All we need to do is to select task-related features from the whole features, like a linear probe.
Fine-tuning is a more widely applied way than linear probe in most realistic situations, it can also be seen as feature learning and selection in an end-to-end manner, which is expected to get higher performance than linear probe.
However, in few-data settings, the performances of fine-tuning are sometimes lower than the linear probe.
When fine-tuning the model in a few-data settings, the predictions can easily reach supervision signal by a few patterns, much more patterns will be collapsed.
The collapsed feature may be helpful for better generalization and the same phenomenon is also observed by~\cite{doersch2020crosstransformers}.

To alleviate the phenomenon, we use a similar method as~\cite{doersch2020crosstransformers}, but in a more general way, which we called sample-based contrastive learning.
Given a sample $x$, we first apply different data augmentations on it to generate two augmented samples $x_q$ and $x_k$, then $x_q$ is forwarding a normal updated network $\mathcal{N}$ to extract feature $f_q$ and $x_k$ is forwarding a momentum updated network by $\mathcal{N}$ to extract feature $f_k$. The resulting $f_q$, $f_k$ and historical feature $f_h$ are used to compute the contrastive loss.
At the same time, the task loss is also optimized normally.
In our method, for classification tasks, $x$ is the input image, $\mathcal{N}$ is the whole network; for instance-based tasks, like object detection, $x$ is the instance or patch, $\mathcal{N}$ is the backbone, neck and an extra appended network after $ROIAlign$ with a single $MLP$ layer; for pixel-based tasks, like semantic segmentation, $x$ is pixels from the same class, $\mathcal{N}$ is the backbone and a project head with two $1\times1$ convolutional layers with ReLU.

\subsection{Transfer Results}

In this part, we exhibit the experimental results of our decoupled transfer learning process. We conduct our approach on different downstream tasks, \ie classification, detection, segmentation, and compare it with other commonly used transfer methods.
All the results in this section are conducted with pretrained model R50 from pretraining stages and 10\% dataset setting to show the strength of our approach on few-shot transfer learning scenario.

As shown in Fig.~\ref{fig:transfer-vp-pipeline}, our transfer pipeline has 4 stages. For stages 1 and 2, we train VQ-GAN to encode all upstream data into a codebook following~\cite{esser2021taming}. For Stage 3, which will use images re-represented with our codebook, we freeze the model backbone and only update parameters for the task head of the model. We train for $5,000$ steps with a batch size of 64. The optimizer is SGD with Nesterov momentum and a momentum parameter of 0.9 in this stage. The initial learning rate is sampled log-uniformly from the range [1, 1e-3] and follows a multi-step decay schedule with weight decay 1e-5. For stage 4, all model parameters are fine-tuned with original downstream task data to calibrate the model precision. The initial learning rate is chosen from a grid of 5 logarithmically spaced values between ranges [1e-2, 1e-5], and the weight decay is similarly chosen from a grid of 4 logarithmically spaced values between 1e-3 and 1e-5. The other hyper-parameters are the same as those in stage 3.

\subsubsection{Downstream Tasks on Classification}

For the classification task, the model task head updated in stage 3 delivering is the last fully connected layer. As the results exhibited in Tab.~\ref{tab:cls-downstream-transfer}, our transfer method shows an evident performance improvement compared with the linear probe and fine-tune method. For an average accuracy on all classification tasks, the transfer method improves 4.2 on the precision result. 
As a result, in the few-shot setting of 10\% classification downstream dataset, our MF approach is more effectively and efficiently utilizing the pretrained model from our pretraining stages compared with the linear probe or fine-tuning method.

\begin{table*}[ht]
\centering
\begin{tabular}{l|ccc}
\toprule  
Dataset    &             Linear Probe &                       Fine-tune &    Transfer           \\
\midrule   
CIFAR-10~\cite{cifar}&      92.4 &          93.3	       &              95.2	     	 \\
CIFAR-100~\cite{cifar}           &       73.5 &     72.1        &              75.7      \\
Food-101~\cite{food}           &        75.8 &     73.6        &              75.7      \\
Oxford-IIIT Pets~\cite{pets}           &           85.7 &     83.7        &              83.9      \\
Oxford 102 Flower~\cite{flowers}           &        94.2 &     63.5        &              94.6      \\
SUN397~\cite{sun}           &         57.9 &     56.2        &              58.2      \\
Stanford Cars~\cite{stanfordcar}           &  52.7 &     50.9        &              52.9      \\
DTD~\cite{DTD}           &        65.0 &         47.8        &              63.3      \\
Caltech-101~\cite{Caltech101}           &    88.5 &      71.3        &              86.5      \\
FGVC-Aircraft~\cite{FGVC}           &      28.7 &      29.7        &              39.8      \\
SVHN~\cite{svhn}           &         61.4 &       94.5        &              95.1      \\
EuroSAT~\cite{eurosat}           &       93.8 &      96.4        &              96.0      \\
RESISC45~\cite{resisc45}           &      82.9 &      89.3        &              90.4      \\
Retinopathy~\cite{retinopathy}           &    73.8 &     78.1        &              79.3      \\
FER2013~\cite{fer2013}           &      55.0 &       58.7        &              59.4      \\
UCF101~\cite{ucf101}           &       71.1 &       70.9        &              68.3      \\
GTSRB~\cite{GTSRB}           &        75.1 &       92.1        &              95.5      \\
\midrule
Average           &     72.2 &    71.9        &              \textbf{76.4}      \\
\bottomrule
\end{tabular}
\caption{Transfer performance on classification downstream tasks with pretrained Up-A R50 model.}
\label{tab:cls-downstream-transfer}
\end{table*}

\subsubsection{Downstream tasks on Object Detection and Semantic Segmentation}

To further prove the generality of our approach, we evaluate our MF pipeline on object detection tasks and semantic segmentation tasks. For object detection downstream tasks, we use Mask-RCNN as our detection method following settings in ~\cite{maskrcnn} and DeepLabv3~\cite{chen2017rethinkingdeeplabv3} as the method for segmentation with pretraining Up-G as backbone model. Task head updated in stage 3 is the network after backbone part including FPN, RPN, cls/det head in detection and ASPP, cls head in segmentation. In stage 4, the whole network for detection and segmentation will be tuned with original downstream images. {Tab.~\ref{tab:downstream-transfer-different-stage}} exhibits the result.

\subsubsection{Transfer with Different Pretraining Stages}

In the whole framework, we have three pretraining stages. In this section, we show transfer results with R50 models trained from different pretraining stages and compare their performance on different downstream tasks. As conducted in the previous section, we evaluate the transfer performance on 20 classification tasks and show the average accuracy results in the first column of Tab.~\ref{tab:downstream-transfer-different-stage}. For object detection and semantic segmentation tasks, we choose PASCAL VOC to show the transfer performance on the second and the third column of Tab.~\ref{tab:downstream-transfer-different-stage} respectively.

\begin{table*}[ht]
\centering
\small
\begin{tabular}{l|ccc}
\toprule  
Model       &   CLS AVG &   VOC07+12 (AP50) &   VOC2012 (mIoU)    \\
\midrule   
Up-A        &   74.5        &   75.4	    &   65.9	        \\
Up-E (C)    &   75.6        &   -           &   -               \\
Up-E (D)    &   -           &   86.2        &   -               \\
Up-E (S)    &   -           &   -           &   74.1            \\
Up-G (C-D)  &   76.0        &   87.4        &   -               \\
\bottomrule
\end{tabular}
\caption{Transfer performance with R50 models from different pretraining stages. The first column is the transfer results averaged on 20 classification downstream tasks. The second column is the object detection results evaluated on PASCAL VOC~\cite{everingham2010pascal} and the last column is the pixel prediction segmentation task evaluated on VOC2012 with mIOU metric.}
\label{tab:downstream-transfer-different-stage}
\end{table*}

For results on classification tasks, Up-E increases 1.1 on precision compared with Up-A model and Up-G shows a further improvement of 0.4 on the final transfer performance. Object detection task and segmentation task also shows a consistent improvement through more pretraining stages. From the results, we can conclude that the pretrained model progressively performs better with more pretraining stages which demonstrates the effectiveness of our whole training pipeline and our transfer approach reserves the incremental consistency through different pretraining stages.

\subsubsection{Transfer with Different Model Architectures and Large Scale Models}

In this section, we show the performance of our transfer method on these models to indicate our transfer pipeline congruously plays a role in different model architectures as well as large scale models.

As exhibited in Tab.~\ref{tab:downstream-transfer-different-model}, with the architecture of \mtbfour{} and \mtbfifteen{}, the transfer method achieves consistent improvement in comparison with linear probe or fine-tuning. For \mtbfour{}, our transfer pipeline gains 1.3, 0.2, 1.5 on classification, detection and segmentation tasks. For a large scale model \mtbfifteen{} with one billion parameters, our transfer learning process also achieves an additional increment of 2.3 on classification downstream tasks and 0.6, 2.5 on detection and segmentation downstream tasks. On the whole, experimental results show the robustness and effectiveness of our approach with different model architecture and different model scales.

\begin{table*}[h]
\small
\centering
\begin{tabular}{l|l|ccc}
\toprule  
Model                   & Dataset               &   LP &   FT &    Transfer           \\
\midrule
\multirow{3}*{R50}      & CLS AVG               &   73.8    &   71.1      &                 75.6   \\
                        & VOC07+12 (AP50)       &   86.7 & 85.4 & 86.2 \\
                        & VOC2012 (mIoU)        &   71.8& 65.0 & 74.1 \\
\midrule   
\multirow{3}*{MN-B4}    & CLS AVG               &   78.4 &     77.8        &              79.7      \\
                        & VOC07+12 (AP50)       &   89.3 & 87.7 & 89.5 \\
                        & VOC2012 (mIoU)        &   77.3 & 75.2 & 78.8 \\
\midrule
\multirow{3}*{MN-B15}   & CLS AVG               &   84.2 &     85.6        &              86.5      \\
                        & VOC07+12 (AP50)       &   87.3 & 85.9 & 87.9 \\
                        & VOC2012 (mIoU)        &   73.8 & 73.0 & 76.3 \\
\bottomrule
\end{tabular}
\caption{Transfer performance with different architectures from pretrained Up-E models. LP and FT stands for results of linear probe and fine-tuning respectively.}
\label{tab:downstream-transfer-different-model}
\end{table*}

\subsection{Ablation Study}
\label{subsec:transfer-ablation}

In this section, we show the effectiveness of our Multi-stage Fine-tuning (MF) method and provide detailed experiments on each stage of the MF pipeline.

\textbf{Effectiveness of Stage 3.} We compare our method with not using the re-represented image, which means we use the original downstream task images. To make a fair comparison, a two-stage training process is applied: 1. Fixing the backbone model and performing linear probe on task head parameters. 2. Fine-tuning the whole network. The two-stage fine-tuning process is corresponding to stage 3 and stage 4 in our MF pipeline since there's no need for reconstruction of stage 1 and stage 2. We show results with classification downstream tasks in setting (a) of Tab.~\ref{tab:codebook_ablation}. Results show that delivering stage plays an important role in MF pipeline that using reconstruction images will bring an average improvement of 0.8 on accuracy on classification tasks.

We also compare the results with using an officially released VQ-GAN\cite{esser2021taming} model which is trained with ImageNet in the delivering stage in Tab.~\ref{tab:codebook_ablation} (b). Since our codebook used in delivering stage is trained with YFCC dataset which corresponds to upstream data used in training CLIP-R50 and the official released VQ-GAN is not matching with the CLIP-R50 upstream, our VQ-GAN codebook is supposed to perform better in encoding the the pretrained model in stage 3 information. As shown in Tab.~\ref{tab:codebook_ablation} (b), with the officially released VQ-GAN codebook, the avg performance decreased 0.7.

\begin{table*}[t]
    \centering
    \resizebox{0.9\linewidth}{!}{%
    \begin{tabular}{lll|cccccc|c}
    \toprule
    Setting & Delivering Data & Calibration Data & CIFAR-10 & CIFAR-100 & Food-101 & Oxford-IIIT Pets & Oxford 102 Flower & SUN397 & Average \\
    \midrule
    Fine-tune  & - & Original & 93.3& 72.1& 73.6 & 83.7 & 63.5 & 56.2 & 73.7\\
    Linear Probe  & Original &  - & 92.4& 73.5& 75.8 & 85.7& 94.2& 57.9 & 79.9\\
    \midrule
    (a)  & Original &  Original & 93.4& 74.1& 75.3& 81.6& 92.1& 57.5 & 79.5\\
    (b)  & Official Rec &  Original & 92.1& 73.4& 75.6& 85.9& 93.9 &57.1 & 79.6\\
    (c) & Rec & Rec & 94.8& 73.7& 72.9& 82.4& 92.6& 57.6 & 79.0\\
    (d) & Rec & Rec & 95.1& 75.3& 75.9& 82.6& 93.7& 58.5 & 80.2\\
    (e)  & - &  Rec + Original & 94.9& 74.1& 73.8& 81.4& 94.0 & 57.8 & 79.3\\
    (f)  & Rec + Original     &  Rec + Original & 93.6 & 73.6 & 75.7 & 83.9 & 94.1& 57.9 & 79.8\\
    \midrule
    Ours  & Rec &  Original &95.2 & 75.6 & 75.7 & 83.0 & 94.6 & 58.2 & 80.3\\
    \bottomrule
    \end{tabular}
    }
    \caption{Experiments to show effectiveness of delivering and calibration stage in our approach.}
    \label{tab:codebook_ablation}
\end{table*}

\begin{table*}[t]
\centering
\small
\begin{tabular}{l|cc}
\toprule  
Dataset   &  Original Dataset &     Rec Dataset   \\
\midrule   
CIFAR-10~\cite{cifar}&      2083 &          1829	   \\
CIFAR-100~\cite{cifar} &       1937      &    1896  \\
Food-101~\cite{food}           &        1956 &     1739\\
Oxford-IIIT Pets~\cite{pets}           &           1742 &     1655  \\
Oxford 102 Flower~\cite{flowers}           &        2757 &     2547  \\
SUN397~\cite{sun}           &         1127 &     1038  \\
\bottomrule
\end{tabular}
\caption{Similarity between upstream, reconstructed downstream, original downstream dataset. The similarity measure is calculated following~\cite{deng2021labels}.}
\label{tab:dataset-similarity-ablation}
\end{table*}

\textbf{Effectiveness of Stage 4.} In stage 4 of MF pipeline, we calibrate the model with original images to refine the model in the downstream domain. In this part, we show results of using re-represented images in the calibration stage in setting (c) of Tab.~\ref{tab:codebook_ablation}. Compared with ours, the performance decreased by 1.3, which shows the importance of calibration with original downstream data. Furthermore, in (d) of Tab.~\ref{tab:codebook_ablation}, we test the model trained from (c) on the re-represented downstream validation set, results show that the model is more suitable for the re-represented downstream domain which further indicates the role of calibration stage.

\textbf{Reconstructed Image vs Data Augmentation.} It is easy to consider a problem of whether using VQ-GAN to reconstruct the image is just a kind of data augment. To verify this, we mix up the re-represented and original downstream images in stage 3 and stage 4 of MF, (e) and (f) in Tab.~\ref{tab:codebook_ablation} corresponding to using mixed data with fine-tuning method and using mixed data with a two-stage fine-tuning process. Results of avg accuracy on classification tasks show that the mixture of data indeed brings some improvement to compare with a simple fine-tuning or linear probe, but it can not achieve our final results. In conclusion, re-representation plays a role in data augmentation but it also brings more in our MF pipeline.

\textbf{Reconstructed Image vs Original Image.} From our intuition in MF method, the reconstructed downstream images are more similar to upstream data and can extract more reliable features with the upstream data pretrained model. To further demonstrate this, we provide quantitative analysis on the similarity between upstream dataset and reconstructed downstream dataset as well as the similarity between upstream dataset and original downstream dataset. We use FD-score to measure the dataset similarity following the method described in~\cite{deng2021labels}. Results in Tab.~\ref{tab:dataset-similarity-ablation} exhibit the FD score between upstream dataset and reconstructed dataset is higher than the score between upstream dataset and original dataset, which means that our codebook successfully encodes the upstream information and re-represents the original downstream dataset into upstream data domain. With the reconstructed images, the pretrained model in stage 3 can extract more reliable features.

\textbf{Effectiveness of Sample-based Contrastive.} To alleviate supervision collapse, we apply the sample-based contrastive method. Tab.~\ref{tab:contrastive_ablation} (g) and (h) shows the result of whether applying sample-based contrastive in our MF pipeline. Furthermore, we apply experiments (c) to (f) to further excavate how it works. (c) and (d) tries to transfer from self-supervised pretrained models to see whether the 
improvements come from self-supervised learning, however, in our 10\% data setting, the model can not be trained well. Then, we use our Up-A model to concatenate with a self-supervised model. Results show that a self-supervised model will bring some extra information and those additional information maybe benefit the model, such as improvement on CIFAR-100 (73.5 to 74.5) or maybe harmful to the model, such as decrement on CIFAR-10 (92.4 to 91.9). Experiments (f) and (h) show that sample-based contrastive method does not work as a regularization method on the classifier as when we apply a linear probe on a transferred model, it can reach almost the same result as it was transferred before.

\begin{table*}[t]
    \centering
    \resizebox{1.0\linewidth}{!}{%
    \begin{tabular}{lll|cccccc|c}
    \toprule
    Exp. & Setting & Backbone & CIFAR-10 & CIFAR-100 & Food-101 & Oxford-IIIT Pets & Oxford 102 Flower & SUN397 & Average\\
    \midrule
    (a)  & Fine-tune & Up-A & 93.3& 72.1& 73.6 & 83.7 & 63.5 & 56.2 & 73.7\\
    (b)  & Linear Probe &  Up-A & 92.4& 73.5& 75.8 & 85.7& 94.2& 57.9 & 79.9\\
    \midrule
    (c)  & Linear Probe &  Self-supervised (Up-A Pretrained) & 58.3& 49.7& 29.1& 42.8& 39.7& 25.9 & 40.9\\
    (d)  & Linear Probe &  Self-supervised (Rand Init) & 19.6& 23.8& 12.9& 5.9& 9.6 &8.7 & 13.4\\
    (e) & Linear Probe & Up-A Concat Self-supervised (Up-A Pretrained) & 91.9& 74.5& 75.1& 85.9& 94.7& 58.5 & 80.0\\
    (f) & Linear Probe & Transferred Up-A & 94.8& 75.7& 75.2& 82.8& 94.2& 58.4 & 80.2\\
    \midrule
    (g)  & Ours w/o Sample-based Contrastive &  Up-A &94.6 & 75.0 & 75.4 & 82.5 & 94.5 & 58.1 & 80.1\\
    (h)  & Ours w/ Sample-based Contrastive &  Up-A &95.2 & 75.6 & 75.7 & 83.0 & 94.6 & 58.2 & 80.3\\
    \bottomrule
    \end{tabular}
    }
    \caption{Experiments to show effectiveness of applying sampled-based contrastive to alleviate supervision collapse in our approach.}
    \label{tab:contrastive_ablation}
\end{table*}

\begin{table*}[t]
    \centering
    \renewcommand\arraystretch{1.1}
    \resizebox{0.8\linewidth}{!}{
    \begin{tabular}{l|ccc|ccc|ccc}
    \toprule
    \multicolumn{1}{l|}{\multirow{2}*{Dataset}} & \multicolumn{3}{c|}{Up-A} & \multicolumn{3}{c|}{Up-E} & \multicolumn{3}{c}{Up-G}\\ \cline{2-10}
     & LP & FT & Transfer & LP & FT & Transfer & LP & FT & Transfer \\ \midrule  
    CIFAR-10~\cite{cifar} & 98.2 & 98.0 & 98.3 & 98.7 & 98.0 & 98.1 & 98.7  & 97.9 & 98.8 \\
    CIFAR-100~\cite{cifar} & 87.8 & 84.2 & 87.9 & 90.1 & 90.0 & 90.7 & 90.4  & 89.5 & 90.2 \\
    Food-101~\cite{food}  & 93.9 & 93.9 & 94.0 & 94.7 & 93.9 & 95.2 & 94.5  & 93.5 & 95.2 \\
    Oxford-IIIT Pets~\cite{pets}     & 92.8 & 92.2 & 93.9 & 95.1 & 94.9 & 95.5 & 95.4  & 94.6 & 95.5 \\
    Oxford 102 Flower~\cite{flowers}  & 99.6 & 98.3 & 99.6 & 99.7 & 99.0 & 99.6 & 99.7 & 97.5 & 99.4 \\
    SUN397~\cite{sun}   & 72.3 & 71.3 & 72.1 & 75.7 & 75.8 & 76.0 & 74.4  & 73.2 & 75.0 \\
    Stanford Cars~\cite{stanfordcar} & 59.4 & 58.2 & 59.5 & 74.9 & 77.2 & 81.0 & 75.4  & 74.1 & 76.1 \\
    DTD~\cite{DTD}      & 70.0 & 70.1 & 70.0 & 73.6 & 71.7 & 74.3 & 74.2  & 70.9 & 74.1 \\
    Caltech-101~\cite{Caltech101} & 93.8 & 94.7 & 93.8 & 94.4 & 94.3 & 95.3 & 94.5  & 92.6 & 94.4 \\
    AirCraft~\cite{FGVC} & 64.8 & 66.4 & 69.9 & 91.8 & 87.3 & 90.3 & 91.8  & 91.1 & 91.3\\
    SVHN~\cite{svhn}     & 58.6 & 95.4 & 95.7 & 66.7 & 95.5 & 95.6 & 66.7 & 66.9 & 67.5 \\
    EuroSAT~\cite{eurosat}  & 95.3 & 96.6 & 96.5 & 96.2 & 96.4 & 95.7 & 96.3 & 96.2 & 96.4 \\
    RESISC45~\cite{resisc45} & 91.9 & 91.9 & 92.5 & 92.8 & 92.9 & 92.5 & 92.7 & 92.2 & 92.8 \\
    Retinopathy~\cite{retinopathy} & 77.9 & 80.1 & 80.7 & 77.6 & 77.1 & 77.6 & 77.0 & 77.0 & 78.1 \\
    FER2013~\cite{fer2013}  & 62.8 & 57.7 & 64.8 & 62.3 & 61.9 & 63.3 & 63.1  & 62.5 & 63.0 \\
    UCF101~\cite{ucf101}   & 85.4 & 75.8 & 79.0 & 87.7 & 83.2 & 84.2 & 88  & 84.9 & 87.9 \\
    GTSRB~\cite{GTSRB}    & 76.2 & 94.2 & 94.3 & 83.3 & 95.8 & 96.0 & 83.6  & 83.5 & 83.7 \\
    PatchCamelyon ~\cite{patchcamelyon}     & 87.8 & 87.8 & 87.9 & 87.5 & 86.5 & 87.1 & 88.0  & 85.2 & 88.5 \\
    ImageNet ~\cite{krizhevsky2012imagenet} & 86.0 & 86.0 & 86.2 & 87.2 & 86.2 & 87.1 & 87.1 & 86.8  & 87.5 \\
    Kinetics-700 ~\cite{carreira2019short} & 52.9 & 32.9 & 53.1 & 54.7 & 54.0 & 54.1 & 54.7  & 54.3 & 54.9 \\
    \midrule
    CLS AVG & 81.2 & 82.0 & 82.4 & 84.2 & 84.6 & 85.5 & 84.4 & 83.2 & 84.5 \\ 
    \bottomrule
    \end{tabular}}
    \caption{Experimental results on \mtbfifteen{} model from pretraining Up-A, Up-E and Up-G with linear probe, fine-tuning and our transfer approach.}
    \label{tab:stage4-final-res}
\end{table*}

\subsection{Final Comparison}

In this section, we provide transfer results with our released model \mtbfifteen{}. We conducted extensive experimental results on \mtbfifteen{} model from pretraining Up-A, Up-E and Up-G with linear probe, fine-tuning and our transfer approach and the results are shown in Tab.~\ref{tab:stage4-final-res}.

\section{Benchmark and Evaluation}
\label{sec:overview}

We start this section by introducing all downstream tasks and their corresponding datasets. Following this is an overview of representative pretrained models included in our benchmark. After a revisit to details of the evaluation, we give the final experimental results at the end of this section.

\begin{table*}[ht]
   \centering
   \resizebox{0.9\textwidth}{!}{
   \begin{tabular}{lrrrrrrr}
   \toprule  
   \multicolumn{1}{l}{\multirow{2}*{Dataset}} & \multicolumn{1}{r}{\multirow{2}*{Classes}} &\multicolumn{1}{r}{\multirow{2}*{Train size}} &\multicolumn{1}{r}{\multirow{2}*{Test size}} &\multicolumn{1}{r}{\multirow{2}*{Evaluation metric}} & \multicolumn{3}{c}{ \quad Few-shot (10\%)} \\
   &&&&  & avg & min & max \\
   \midrule   
   CIFAR-10~\cite{cifar}               & 10  	  &  50,000	     &     10,000      & accuracy	& 500.0 & 500 & 500 \\
   CIFAR-100~\cite{cifar}              & 100     &  50,000      &     10,000      & accuracy   & 50.0 & 50 & 50 \\
   Food-101~\cite{food}	          & 101     &  75,750      &     25,250      & accuracy   & 75.0 & 75 & 75 	\\
   Oxford-IIIT Pets~\cite{pets}        & 37      &  3,680       &     3,669       & mean per class & 9.9 & 9 & 10 \\
   Oxford 102 Flower~\cite{flowers}  & 102    & 2,040     &     6,149       & mean per class & 2.0 & 2 & 2   \\
   SUN397~\cite{sun}                  & 397     & 19,850       &     19,850      & accuracy & 5.0 & 5 & 5  \\
   Stanford Cars~\cite{stanfordcar}   & 196     & 8,144        &     8,041       & accuracy & 3.7 & 2 & 6  \\
   DTD~\cite{DTD}                     & 47      & 3,760        &     1,880       & accuracy & 8.0 & 8 & 8  \\    
   Caltech-101~\cite{Caltech101}      & 102     & 3,060        &     6,085       & mean per class & 3.0  & 3  & 3 \\
   FGVC-Aircraft~\cite{FGVC}          & 100     & 6,667        &     3,333       & mean per class & 6.0 & 6  & 6 \\
   \midrule        
   SVHN~\cite{svhn}                   & 10      & 73,257       &     26,032      & accuracy & 732.0  & 465  & 1,386 \\
   EuroSAT~\cite{eurosat}             & 10      & 21,600       &     5,400       & accuracy & 215.6  & 159  & 243 \\
   RESISC45~\cite{resisc45}           & 45      & 25,200       &     6,300       & accuracy & 55.4  & 53  & 58 \\
   Retinopathy~\cite{retinopathy}     & 5       & 46,032       &     42,670      & accuracy & 920.4  & 94  & 3,394 \\
   FER2013~\cite{fer2013}             & 7       & 28,709       &     3,589       & accuracy & 409.7  & 43  & 721 \\ 
   UCF101~\cite{ucf101}              & 101     & 9,537        &     3,696       & accuracy & 9.0  & 7  & 12 \\
   GTSRB~\cite{GTSRB}                 & 43      & 39,209       &     12,630      & accuracy & 91.2  & 21  & 225 \\
   PatchCamelyon~\cite{patchcamelyon} & 2       & 294,912      &     32,768      & accuracy & 14,745.5  & 14,744  & 14,747 \\
   ImageNet~\cite{russakovsky2015imagenet}     & 1000    & 1,281,167    &     50,000      & accuracy & 128.1  & 73  & 130 \\
   Kinetics-700~\cite{carreira2019short} & 700     & 542,285      &     34,819      & accuracy & 77.0  & 39  & 100 \\
   \midrule
   VOC07+12~\cite{everingham2010pascal}   & 20      &   16551      &      4,952      & AP50   & 246.3  & 91  & 1,670   \\
   WIDER FACE~\cite{yang2016wider}             & 2       &   12,880     &      3,226      & AP50   & 1,288  & 1,288  & 1,288  \\
   CityPersons~\cite{cordts2016cityscapes} & -       &   2,975      &        500      & $\text{MR}^{-2}$     & 2,689  & 2,689  & 2,689  \\
   VOC2012~\cite{everingham2010pascal}   & -       &   10,582     &     1,449       & mIoU   & - & - & - \\
   NYUv2~\cite{NYUDEPTH}                 &   -     &   24,231     &       653       & RMSE   & - & - & - \\
   KITTI~\cite{kitti}                     &   -     &   23,157     &      696        & RMSE   & - & - & - \\
   \bottomrule
   \end{tabular}
   }
   \caption{\label{tab:down-dataset-details} \textbf{Details of dowstream classification datasets.} The rightmost three columns ``avg'', ``min'' and ``max'' refer to the average, minimum and maximum number of 10\% data in all classes of the dataset.}
\end{table*}

\subsection{Downstream Tasks and Datasets}
We select in total 26 downstream datasets consisting of 4 task types. 
The details, including specific tasks and their corresponding datasets, are provided as following:

\textbf{Classification.} There are 20 datasets used for the classification task. Specifically, it includes general object classification (CIFAR-10, CIFAR-100, ImageNet, Caltech-101, SVHN), fine-grained object classification (Food-101, Stanford Cars, FGVC-Aircraft, Oxford-IIIT Pets, Oxford 102 Flower), human-related classification (FER2013, UCF101, Kinetics-700), scene classification (SUN397), texture classification (DTD), traffic sign recognition (GTSRB), remote sensing image classification (EuroSAT, RESISC45) and medical image classification (PatchCamelyon, Retinopathy). For the two video datasets (UCF101 and Kinetics-700), we use the middle frame of each video sequence as the input image. Details of the datasets and the corresponding evaluation metrics are provided in Tab.~\ref{tab:down-dataset-details}. 

\textbf{Object Detection.} We consider VOC07+12 for object detection. This dataset is widely used as a downstream benchmark with 5k test images over 20 object categories. We use the 5k trainval images in VOC 2007 and 11.5k trainval images in VOC 2012 as the training set of VOC07+12. We report AP50 (average precision at IoU threshold of 50\%) as the metric for object detection.

\textbf{Pedestrian Detection.} Pedestrian detection is a popular topic in the computer vision community. We use CityPersons for evaluating the pedestrian detection task, which is a subset of Cityscapes and consists of only person annotations. There are 2,975 images for training, 500 and 1,575 images for validation and testing. We report $\text{MR}^{-2}$ for the reasonable and heavy occlusion subsets.

\textbf{Face Detection.} Besides object detection and pedestrian detection, we also incorporate the WIDER FACE dataset for face detection in our benchmark. WIDER FACE includes 393,703 faces with a high degree of variability in scale, pose and occlusion. When evaluating on this dataset, images are divided into three levels according to the detection difficulty: easy, medium, and hard. We report AP50 at these three levels.

\textbf{Semantic Segmentation.} The VOC2012 dataset is also included as the downstream semantic segmentation task. It contains 20 foreground object classes and one background class. We use the augmented dataset with 10,582 (trainaug) training images. The performance is measured in terms of pixel intersection-over-union averaged (mIOU) across the 21 classes.

\textbf{Depth Estimation.} We include two widely-used datasets, KITTI and NYUv2, to conduct evaluation for depth estimation. KITTI is one of the most popular datasets for mobile robotics and autonomous driving. Its RGB images have a resolution of around $1242\times376$, while the corresponding depth maps are of very low density. NYUv2 is a dataset that provides images and depth maps for different indoor scenes captured at a resolution of $640\times480$, where depth maps have an upper bound of 10 meters. We report RMSE for both of the datasets.

For each dataset, we evaluate 20 collected pretrained models with 100\% and 10\% training data respectively. To obtain the 10\% training data split, we randomly shuffle the training data instances of each class and fetch the first 10 percent, which keeps the distribution of data classes unchanged. The number of instances is rounded down if it is not divisible. 

To some extent, the 10\% split setting is similar to the $N$-way $K$-shot setting for few-shot and meta-learning. The difference is that our sampling methodology is based on percentage sampling, which takes into account class sizes of the training dataset, rather than a fixed number of samples in each class. It follows the same underlying data distribution of the 100\% datasets. Properties like long-tailed distribution can be well-kept, so it can better resemble real-world scenarios than the manually designed $N$-way $K$-shot style. For convenience, we also present 10\% setting as ``Few-shot'' in Tab.~\ref{tab:down-dataset-details}, and we provide some statistics on class distribution after our 10\% sampling.

\subsection{Models}
In combination with the tasks and datasets listed above, we collect the following representative models for evaluation. For fair comparison, we download released checkpoints rather than reproduce the models. The included models are as follows:
\begin{itemize}
    \item {\textit{ResNet}}: It contains three ImageNet-1k pretrained ResNet networks released by \cite{he2016deep}, R50, R101 and R152 respectively.
    \item {\textit{CLIP}}: They are provided by \cite{CLIP}, including CLIP-R50 and CLIP-R$50\times16$. Among them, CLIP-R50 is at the same parameter level as R50, and CLIP-R$50\times16$ is the largest released model we can obtain. 
    \item {\textit{CLIP-ViT}}: It is trained using the ViT-B/16~\cite{dosovitskiy2020image} architecture as the image encoder. The model is also released by \cite{CLIP}.\footnote{The model weights of CLIP-ViT-L series are not available publicly for now.}
    \item {\textit{Instagram-Pretrained ResNeXt}}: This includes two models: $32\times8$d and $32\times48$d, released by \cite{instance_1}. The models use ResNeXt as their architecture.
    \item {\textit{Big Transfer (BiT)}}: It is a series of  pretrained models released by \cite{bit}. We use BiT-S-R ($50\times1$), BiT-M-R ($50\times1$) and BiT-M-R ($152\times4$) for evaluation, where BiT-S and BiT-M represent models pretrained on ImageNet-1k and ImageNet-21K, respectively. BiT-M-R($152\times4$) is the largest model used in \cite{bit}. \footnote{The BiT-L series models pretrained on JFT-300M are not publicly available yet.}
    \item {\textit{Vision Transformer (ViT)}}: We also include Vision Transformer~\cite{dosovitskiy2020image} to our downstream benchmark, including ViT-B/16, ViT-L/16 and ViT-H/14. However, \cite{dosovitskiy2020image} does not directly provide PyTorch checkpoints, so we obtain weights via the Python library ``timm"~\cite{timm}.
    \item {\textit{SwAV~\cite{caron2020unsupervised}, DeepClusterV2~\cite{caron2018deep} and MoCo v2~\cite{chen2020improved}}}: We also incorporate R50 models trained with these three popular self-supervised learning methods.
    \item {\textit{Detco}~\cite{detco}}: It is designed for unsupervised detection tasks, and is also included in our downstream evaluation. We select the Detco model based on R50 architecture.
\end{itemize}

\subsection{Evaluation Details}
\label{subsec:eval_setting}

To evaluate pretrained models, we select {\em linear probe} as the main evaluation method, which keeps the backbone parameters fixed and only fine-tune task-specific heads with downstream training data. 
This is because fixing the backbone can more directly evaluate the task generalization capability of pretrained weights. In the rest of this section, we describe implementation details for evaluation on different tasks.

\noindent \textbf{Classification.}
For each classification dataset, we train a fully-connected layer over the extracted feature from the fixed backbone. We consider the following two strategies and report the best performance.
\begin{itemize}
    \item We use SGD with Nesterov momentum as the optimizer, with maximum 10,000 iterations and a batch size of 64 (100,000 maximum iterations for ImageNet, Kinetics-700 and PatchCamelyon due to their large training set sizes). We do not apply any data augmentation during training. As preprocessing, we resize input images to $r$ pixels along the shorter side and take a $r\times r$ center crop, where $r$ is based on the pretraining resolution, \eg $r=224$ for R50 models, $r=256$ for \mtbfour{} and \mtbfifteen{} models. We select the learning rate and weight decay by a logarithmically spaced grid search, where the learning rate is between $10^{-4}$ and $0.1$, while the weight decay is between $10^{-6}$ and $10^{-3}$. We also search with two momentum values $0.9$ and $0.99$.
    
    \item For a few models we find the SGD optimizer above hard to converge, so we also include the linear probe setting provided by \cite{CLIP}. In details, we use the L-BFGS optimizer with maximum 1,000 iterations and search for the best regularization strength in the logarithmic space.
\end{itemize} 
We analyze model performance using FPR@Recall as another stricter metric besides accuracy, since the former better reveals gaps among models when their accuracy values are too high to differ. Traditionally FPR@Recall is mainly used for binary classification, yet we apply the idea of one-vs-rest, which treats each class of a multi-class dataset as a binary classification problem and calculates the corresponding FPR@Recall. We report the average Recall@FPR over all classes as an additional metric for some top-performing models.

\noindent \textbf{Object Detection.}
For object detection, we evaluate models on PASCAL VOC07+12. We use the Faster R-CNN~\cite{ren2015faster} architecture and run 24,000 iterations with a batch size of 16. We use SGD as the optimizer with a learning rate between 0.001 and 0.1. The weight decay is set to 0.0001, and the momentum is set to 0.9. During evaluation, we follow the standard setting and resize the shorter and longer sides to maximum $800$ and $1,333$ pixels. The code is modified based on detectron2~\cite{wu2019detectron2}. Similar to Recall@FPR for classification, we also consider FPPI as the second metric for VOC07+12 object detection.

\noindent \textbf{Face Detection.}
Face detection experiments are conducted on WIDER FACE. We use RetinaFace~\cite{retinaface} as the framework based on a public code base\footnote{\url{https://github.com/biubug6/Pytorch\_Retinaface}}. Specifically, we train all the models for 80 epochs (800 epochs for 10\% setting) using a batch size of 32 with SGD, momentum of 0.9, weight decay of $5\times10^{-4}$ and searching the learning rate from $10^{-3}$ to $10^{-1}$.
During training, we randomly crop square patches from the original images and
resize these patches into $840\times840$. We also apply random horizontal flips and photo-metric color distortion as data augmentation.

\noindent \textbf{Pedestrian Detection.}
For pedestrian detection, we evaluate models on CityPersons. We run 60 epochs (600 epochs for 10\% setting) with a batch size of 8. The detection head architecture is Faster R-CNN. We use SGD as the optimizer with a learning rate between 0.0002 and 0.02. The weight decay and the momentum are set to 0.0001 and 0.9 respectively. We also use data augmentation of photo-metric distortion and random crop, and images scales are set to $1,216\times608$ and $2,048\times1,024$.

\noindent \textbf{Semantic Segmentation.}
For semantic segmentation, we evaluate pretrained models on PASCAL VOC2012. We run 30 epochs (300 epochs for 10\% setting) with a batch size of 16. The network architecture is based on DeepLabv3~\cite{chen2017rethinkingdeeplabv3}. We train the models using SGD as the optimizer with a learning rate between 0.0008 and 0.07. The weight decay is set to 0.0001, and the momentum is set to 0.9.

\noindent \textbf{Depth Estimation.}
We select NYUv2 and KITTI, two popular datasets, for depth estimation evaluation. The experiments are based on the framework of BTS~\cite{bts}. We run 50 epochs (500 epochs for 10\% setting) with a batch size of 16. We use Adam as the optimizer with a learning rate between 0.00000625 and 0.001. The weight decay is set to 0.001, and $\epsilon$ is set to 0.000001. Images are scaled to $544\times416$.

\subsection{Results}

\subsubsection{Overall Comparison}

\begin{figure*}
   \centering
    \includegraphics[width=0.944\linewidth]{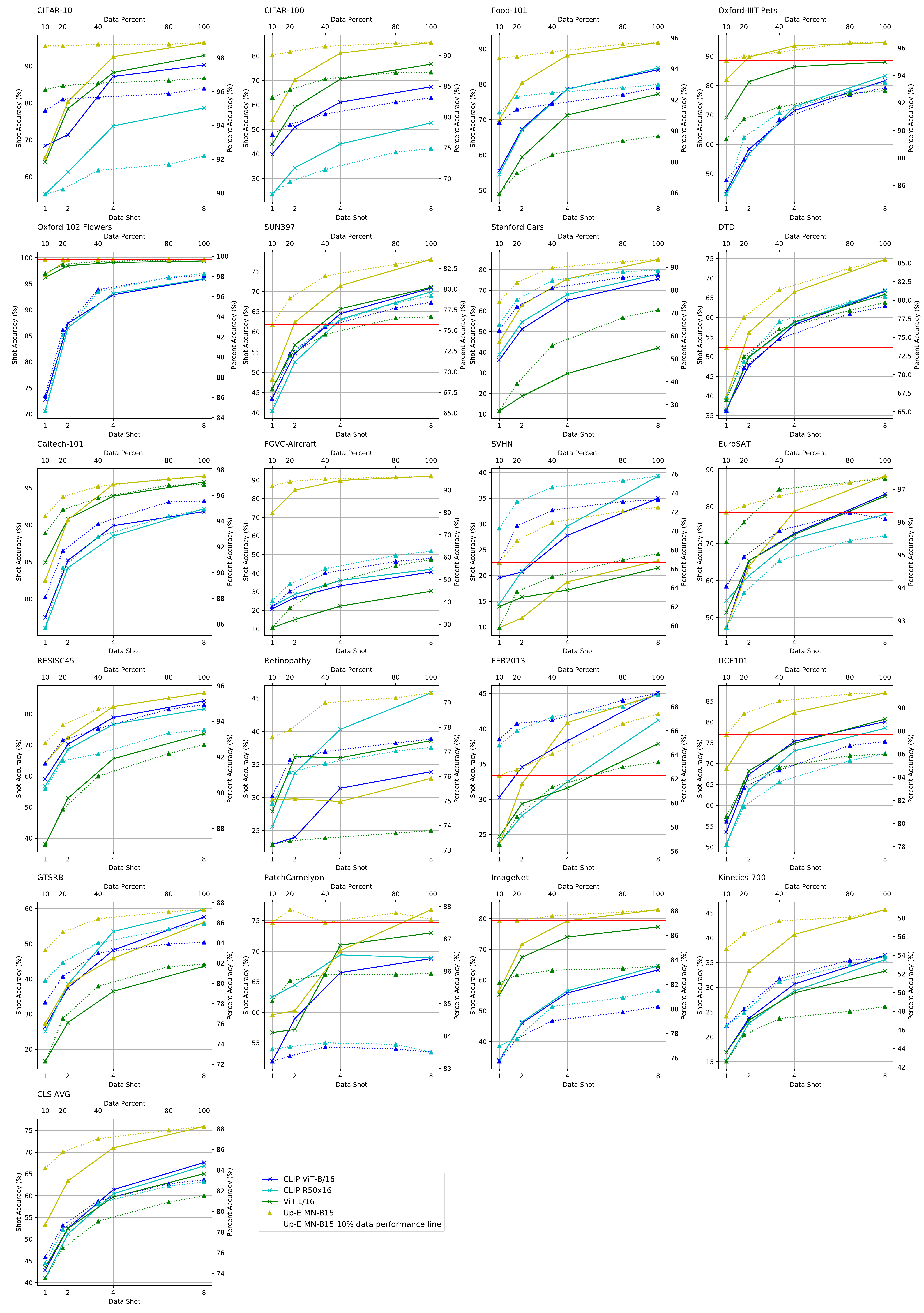}

    \caption{Linear probe performance with different shots and different percentages over 20 classification tasks. We use dashed line to represent percent performance and solid line to represent shot performance. Meanwhile, we highlight the comparison between our INTERN models using 10\% data and other models with 100\% data.}
    \label{fig:teng-percentage and shot}
 \end{figure*}

In this section, we make a comparison between our INTERN models and other collected public models on 26 linear evaluation tasks. Our models include Up-A, Up-E and Up-G equipped with R50, MN-B4 and MN-B15 architecture respectively. The linear probe results of 100\% and 10\% train data are provided in Tab.~\ref{tab:whole_results}. We note that our Up-G MN-B15, which is a multi-stage pretrained generalist model, achieves state-of-the-art results on most tasks, surpassing the best publicly available pretrained model, CLIP-R$50\times{16}$, by large margins. Take low-data (10\%) regime as an example, Up-G MN-B15 outperforms CLIP-R$50\times{16}$ with +9.4\% average accuracy on the classification evaluation suite, +11.4\% AP on VOC detection and +12.1\% mIoU on semantic segmentation. 

Our MN-B15 based models achieve the best performance on 18 classification datasets out of 20 in the classification evaluation suite (14 for Up-G MN-B15, 4 for Up-E MN-B4). We notice that our model obtains an obvious advantages on fine-grained datasets, Food-101, Oxford 102 Flowers, Oxford Pets and FGVC-Aircraft for instance. Besides low-data (10\%) regime, Up-G MN-B15 achieves the best performance in full-data (100\%) setting as well, with 88.4\% average classification suite accuracy, 90.7\% AP on VOC detection and 78.8\% mIoU on VOC semantic segmentation. For WIDER FACE detection and CityPersons detection, Up-G MN-B15 also achieves the state-of-the-art results. Specifically, Up-G MN-B15 has a obvious advantage on CityPersons detection, which is a difficult detection problem due to the data domain of monitoring, and surpasses CLIP-R$50\times{16}$ 8.6\% and 11.3\% in Reasonable and Heavy splits respectively.

In addition, it is worth noting that our INTERN models steadily outperform other publicly released similar-complexity rivals. Our R50 of Up-E outperforms the ImageNet pretrained on all metrics in both low-data and full-data settings. The results of R50 checkpoints from MoCo v2 and SwAV both show unsteady performances across variance tasks transferring, there is still a large gap between them and our models. Our proposed MN-B4, which is also a similar-complexity rival of R50, achieves the best performance in the same model complexity level. Up-G MN-B4 outperforms ImageNet pretrained R50 with +15.8\% average accuracy on classification suite, +19.6\% on VOC detection and +13.4\% mIoU on VOC semantic segmentation. The results prove that INTERN's good compatibility with different types of backbone networks.

 \begin{figure*}
   \centering
    \includegraphics[width=0.8\linewidth]{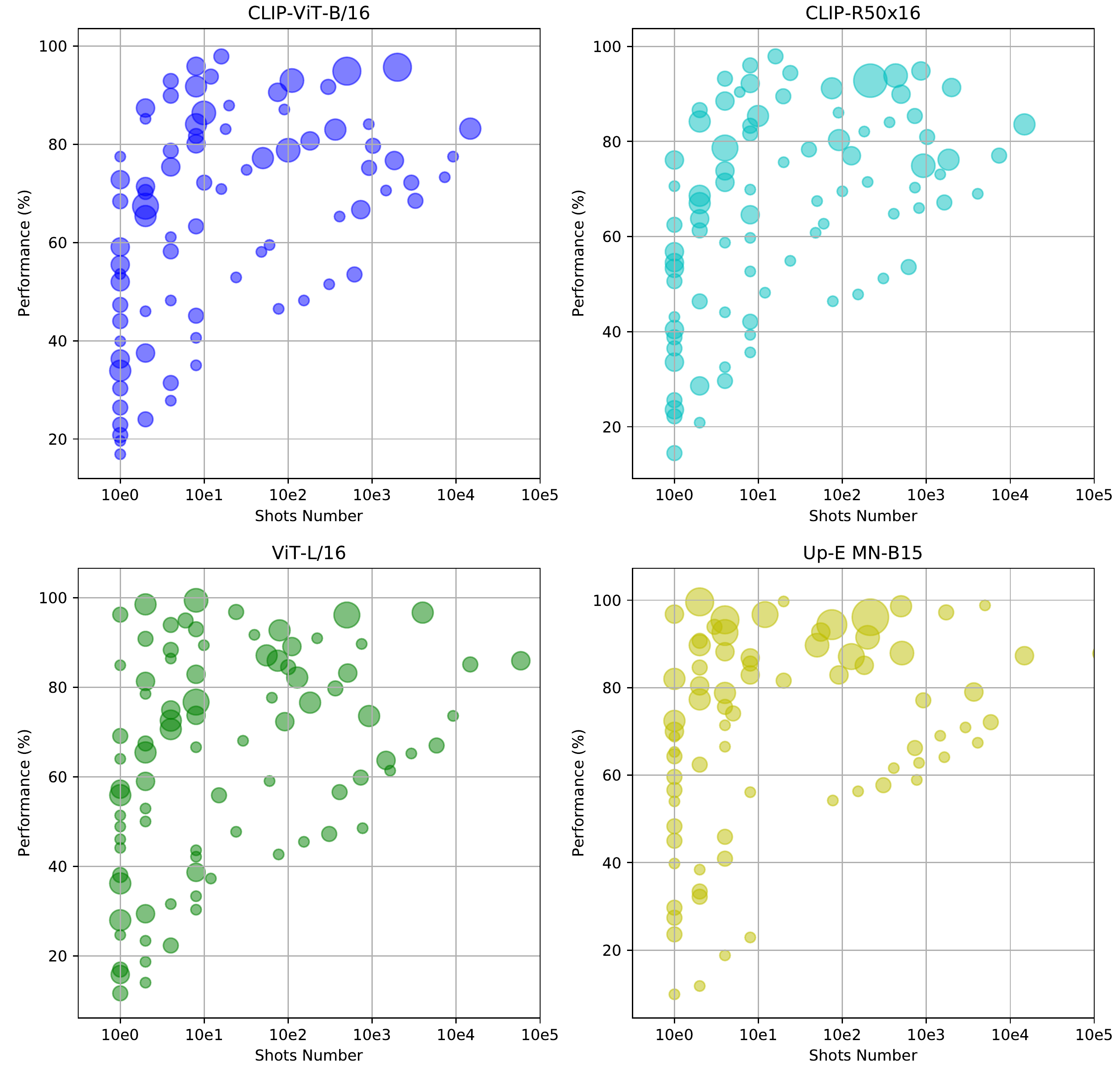}
    \caption{Performance of all dataset settings presented in shots way. We plot model performance over all classification datasets with 10\%, 20\%, 40\%, 80\% and 1, 2, 4, 8 shots. The size of points represents the density of their gathering.}
    \label{fig:teng-all-shots-crowd}
 \end{figure*}

\subsubsection{10\% vs 100\%}
We aim to train INTERN models with practical transferring abilities facing various downstream tasks with fewer data. Tab.~\ref{tab:whole_results} shows the comparison between INTERN models' low-data (10\%) setting results and other models' full-data (100\%) setting results. The performance of our Up-G MN-B15 using 10\% train data consistently surpasses CLIP-R$50\times{16}$ using 100\% data, with +1.5\% for classification suite, +6.2\% AP for VOC detection and +8.6\% mIoU for VOC semantic segmentation. In the model complexity level of R50, our Up-E MN-B4 also wins the 10\% vs. 100\% game for all downstream tasks compared with other R50 models.

\subsubsection{Different Data Percents and Shots}
To observe the trends of model performance as data volumes grow, we evaluate the Up-E MN-B15, our best classification expert model with 20\%, 40\%, and 80\% training data respectively. We also proceed with the evaluation following the traditional few-shot setting with 1, 2, 4, 8 shots. Besides our models, we conduct the same assessment on the three best-performing public released models: ViT-L/16, CLIP-ViT/B-16, and CLIP-R$50\times{16}$. The results of each dataset are provided in Fig.~\ref{fig:teng-percentage and shot}. 

Our Up-E MN-B15 achieves the best performance in both various percentages and few-shot settings. We notice that the performance curves of Up-E MN-B15 in the fine-grained dataset are even almost saturated with low data percents, \eg, Oxford 102 Flowers, FGVC-Aircraft, Food-101 and Oxford-IIIT Pets. In order to highlight our advantage in 100\% and 10\% comparison, we use a red horizontal line indicating the Up-E MN-B15  performance using only 10\% data, which defeats all the other three models with 100\% data in 11 datasets. The curves also show some other models' strengths, for instance, ViT-L/16 achieves comparable performance with our model in EuroSAT, and CLIP-ViT/B16 outperforms our model in SVHN and FER2013.

In addition to proving our Up-E MN-B15's advantages training with different train data size, we visualize per model performance in the form of data shots provided in Fig.~\ref{fig:teng-all-shots-crowd}. In detail, each dataset is presented with 5 points referring to 10\%, 20\%, 40\%, 80\% and 100\% respectively. Considering some points may overlap, we use different sizes of points to represent the density of their gathering. The results prove the universal improvements of Up-E MN-B15 in various classification datasets.

\begin{table*}
   \normalsize
   \centering
   \resizebox{\linewidth}{!}
   {
      \begin{tabular}{l|lcccccccccc}
      FPR                                                & Model         & \rotatebox{90}{CIFAR-10}                                                   & \rotatebox{90}{CIFAR-100}                                                   & \rotatebox{90}{Food-101}                                                   & \rotatebox{90}{Oxford-IIIT Pets}                                                  & \rotatebox{90}{Oxford 102 Flower}                                                   & \rotatebox{90}{SUN397}                                                    & \rotatebox{90}{Stanford Cars}                                                  & \rotatebox{90}{DTD}                                                   & \rotatebox{90}{Caltech-101}                                                   & \rotatebox{90}{FGVC-Aircraft}        \\     \hline                                    
      \multicolumn{1}{l|}{\multirow{4}{*}{Recall0.9}}    & CLIP-ViT-B/16 & {\color[HTML]{3166FF} 0.00}{\color[HTML]{000000} 253}                     & {\color[HTML]{3166FF} 0.0}{\color[HTML]{000000} 1371}                      & {\color[HTML]{3166FF} 0.00}{\color[HTML]{000000} 167}                     & {\color[HTML]{3166FF} 0.0}{\color[HTML]{000000} 1001}                 & {\color[HTML]{3166FF} 0.00}{\color[HTML]{000000} 578}                     & {\color[HTML]{3166FF} 0.00}{\color[HTML]{000000} 980}                    & {\color[HTML]{3166FF} 0.0}{\color[HTML]{000000} 2301}                  & {\color[HTML]{3166FF} 0.}{\color[HTML]{000000} 12865}                 & {\color[HTML]{3166FF} 0.00}{\color[HTML]{000000} 602}                     & {\color[HTML]{3166FF} 0.}{\color[HTML]{000000} 14368}    \\
      \multicolumn{1}{l|}{}                              & CLIP-R$50\times{16}$   & {\color[HTML]{3166FF} 0.0}{\color[HTML]{000000} 1522}                     & {\color[HTML]{3166FF} 0.0}{\color[HTML]{000000} 3277}                      & {\color[HTML]{3166FF} 0.00}{\color[HTML]{000000} 138}                     & {\color[HTML]{3166FF} 0.00}{\color[HTML]{000000} 949}                 & {\color[HTML]{3166FF} 0.00}{\color[HTML]{000000} 771}                     & {\color[HTML]{3166FF} 0.0}{\color[HTML]{000000} 1142}                    & {\color[HTML]{3166FF} 0.0}{\color[HTML]{000000} 1544}                  & {\color[HTML]{3166FF} 0.}{\color[HTML]{000000} 10326}                 & {\color[HTML]{3166FF} 0.00}{\color[HTML]{000000} 616}                     & {\color[HTML]{3166FF} 0.}{\color[HTML]{000000} 14361}    \\
      \multicolumn{1}{l|}{}                              & ViT-L/16       & {\color[HTML]{3166FF} 0.00}{\color[HTML]{000000} 164}                     & {\color[HTML]{3166FF} 0.00}{\color[HTML]{000000} 793}                      & {\color[HTML]{3166FF} 0.00}{\color[HTML]{000000} 420}                     & {\color[HTML]{3166FF} 0.00}{\color[HTML]{000000} 713}                 & {\color[HTML]{3166FF} 0.000}{\color[HTML]{000000} 20}                     & {\color[HTML]{3166FF} 0.00}{\color[HTML]{000000} 931}                    & {\color[HTML]{3166FF} 0.}{\color[HTML]{000000} 20980}                  & {\color[HTML]{3166FF} 0.}{\color[HTML]{000000} 13913}                 & {\color[HTML]{3166FF} 0.00}{\color[HTML]{000000} 630}                     & {\color[HTML]{3166FF} 0.}{\color[HTML]{000000} 28386}     \\
      \multicolumn{1}{l|}{}                              & Up-E MN-B15        & {\color[HTML]{3166FF} 0.00}{\color[HTML]{FE0000} 0}{\color[HTML]{000000} 18}                     & {\color[HTML]{3166FF} 0.00}{\color[HTML]{000000} 236}                      & {\color[HTML]{3166FF} 0.00}{\color[HTML]{FE0000} 0}{\color[HTML]{000000} 62}                     & {\color[HTML]{3166FF} 0.00}{\color[HTML]{000000} 244}                  & {\color[HTML]{3166FF} 0.000}{\color[HTML]{FE0000} 0}{\color[HTML]{000000} 3}                     & {\color[HTML]{3166FF} 0.00}{\color[HTML]{000000} 695}                    & {\color[HTML]{3166FF} 0.00}{\color[HTML]{000000} 991}                  & {\color[HTML]{3166FF} 0.}{\color[HTML]{FE0000} 0}{\color[HTML]{000000} 6192}                 & {\color[HTML]{3166FF} 0.00}{\color[HTML]{000000} 534}                     & {\color[HTML]{3166FF} 0.}{\color[HTML]{FE0000} 00}{\color[HTML]{000000} 154}    \\ \hline
      \multirow{4}{*}{Recall0.95}                        & CLIP-ViT-B/16 & {\color[HTML]{3166FF} 0.00}{\color[HTML]{000000} 693}                     & {\color[HTML]{3166FF} 0.0}{\color[HTML]{000000} 2669}                      & {\color[HTML]{3166FF} 0.00}{\color[HTML]{000000} 376}                     & {\color[HTML]{3166FF} 0.0}{\color[HTML]{000000} 1836}                  & {\color[HTML]{3166FF} 0.0}{\color[HTML]{000000} 1058}                     & {\color[HTML]{3166FF} 0.0}{\color[HTML]{000000} 1698}                    & {\color[HTML]{3166FF} 0.0}{\color[HTML]{000000} 4107}                  & {\color[HTML]{3166FF} 0.}{\color[HTML]{000000} 21574}                 & {\color[HTML]{3166FF} 0.0}{\color[HTML]{000000} 1208}                     & {\color[HTML]{3166FF} 0.}{\color[HTML]{000000} 19776}   \\
                                                         & CLIP-R$50\times{16}$   & {\color[HTML]{3166FF} 0.0}{\color[HTML]{000000} 3137}                     & {\color[HTML]{3166FF} 0.0}{\color[HTML]{000000} 5740}                      & {\color[HTML]{3166FF} 0.00}{\color[HTML]{000000} 325}                     & {\color[HTML]{3166FF} 0.0}{\color[HTML]{000000} 1840}                  & {\color[HTML]{3166FF} 0.0}{\color[HTML]{000000} 1517}                     & {\color[HTML]{3166FF} 0.0}{\color[HTML]{000000} 1972}                    & {\color[HTML]{3166FF} 0.0}{\color[HTML]{000000} 2814}                  & {\color[HTML]{3166FF} 0.}{\color[HTML]{000000} 18535}                 & {\color[HTML]{3166FF} 0.0}{\color[HTML]{000000} 1429}                     & {\color[HTML]{3166FF} 0.}{\color[HTML]{000000} 20867}    \\
                                                         & ViT-L/16       & {\color[HTML]{3166FF} 0.00}{\color[HTML]{000000} 488}                     & {\color[HTML]{3166FF} 0.0}{\color[HTML]{000000} 2136}                      & {\color[HTML]{3166FF} 0.00}{\color[HTML]{000000} 916}                     & {\color[HTML]{3166FF} 0.0}{\color[HTML]{000000} 1069}                  & {\color[HTML]{3166FF} 0.000}{\color[HTML]{000000} 32}                     & {\color[HTML]{3166FF} 0.0}{\color[HTML]{000000} 1591}                    & {\color[HTML]{3166FF} 0.}{\color[HTML]{000000} 31749}                  & {\color[HTML]{3166FF} 0.}{\color[HTML]{000000} 27484}                 & {\color[HTML]{3166FF} 0.00}{\color[HTML]{000000} 952}                     & {\color[HTML]{3166FF} 0.}{\color[HTML]{000000} 38714}     \\
                                                         & Up-E MN-B15        & {\color[HTML]{3166FF} 0.00}{\color[HTML]{FE0000} 0}{\color[HTML]{000000} 64}                      & {\color[HTML]{3166FF} 0.0}{\color[HTML]{FE0000} 0}{\color[HTML]{000000} 772}                      & {\color[HTML]{3166FF} 0.00}{\color[HTML]{000000} 146}                     & {\color[HTML]{3166FF} 0.0}{\color[HTML]{FE0000} 0}{\color[HTML]{000000} 413}                  & {\color[HTML]{3166FF} 0.000}{\color[HTML]{000000} 12}                     & {\color[HTML]{3166FF} 0.0}{\color[HTML]{000000} 1535}                    & {\color[HTML]{3166FF} 0.0}{\color[HTML]{000000} 2191}                  & {\color[HTML]{3166FF} 0.}{\color[HTML]{000000} 15545}                 & {\color[HTML]{3166FF} 0.00}{\color[HTML]{000000} 867}                     & {\color[HTML]{3166FF} 0.}{\color[HTML]{FE0000} 00}{\color[HTML]{000000} 382}   \\ \hline
      \multirow{4}{*}{Recall0.99}                        & CLIP-ViT-B/16 & {\color[HTML]{3166FF} 0.0}{\color[HTML]{000000} 3600}                     & {\color[HTML]{3166FF} 0.}{\color[HTML]{000000} 12250}                      & {\color[HTML]{3166FF} 0.0}{\color[HTML]{000000} 1840}                     & {\color[HTML]{3166FF} 0.0}{\color[HTML]{000000} 9685}                  & {\color[HTML]{3166FF} 0.0}{\color[HTML]{000000} 2661}                     & {\color[HTML]{3166FF} 0.0}{\color[HTML]{000000} 5386}                    & {\color[HTML]{3166FF} 0.0}{\color[HTML]{000000} 9222}                  & {\color[HTML]{3166FF} 0.}{\color[HTML]{000000} 29140}                 & {\color[HTML]{3166FF} 0.0}{\color[HTML]{000000} 3691}                     & {\color[HTML]{3166FF} 0.}{\color[HTML]{000000} 26433}     \\
                                                         & CLIP-R$50\times{16}$   & {\color[HTML]{3166FF} 0.}{\color[HTML]{000000} 10522}                     & {\color[HTML]{3166FF} 0.}{\color[HTML]{000000} 16073}                      & {\color[HTML]{3166FF} 0.0}{\color[HTML]{000000} 1739}                     & {\color[HTML]{3166FF} 0.}{\color[HTML]{000000} 11622}                  & {\color[HTML]{3166FF} 0.0}{\color[HTML]{000000} 3291}                     & {\color[HTML]{3166FF} 0.0}{\color[HTML]{000000} 6417}                    & {\color[HTML]{3166FF} 0.0}{\color[HTML]{000000} 6971}                  & {\color[HTML]{3166FF} 0.}{\color[HTML]{000000} 32819}                 & {\color[HTML]{3166FF} 0.0}{\color[HTML]{000000} 3633}                     & {\color[HTML]{3166FF} 0.}{\color[HTML]{000000} 29039}      \\
                                                         & ViT-L/16       & {\color[HTML]{3166FF} 0.0}{\color[HTML]{000000} 2162}                     & {\color[HTML]{3166FF} 0.}{\color[HTML]{000000} 12669}                      & {\color[HTML]{3166FF} 0.0}{\color[HTML]{000000} 3734}                    & {\color[HTML]{3166FF} 0.0}{\color[HTML]{000000} 5798}                  & {\color[HTML]{3166FF} 0.00}{\color[HTML]{000000} 334}                     & {\color[HTML]{3166FF} 0.0}{\color[HTML]{000000} 5836}                    & {\color[HTML]{3166FF} 0.}{\color[HTML]{000000} 48845}                  & {\color[HTML]{3166FF} 0.}{\color[HTML]{000000} 40641}                 & {\color[HTML]{3166FF} 0.0}{\color[HTML]{000000} 2259}                     &{\color[HTML]{3166FF} 0.}{\color[HTML]{000000} 49738}       \\
                                                         & Up-E MN-B15        & {\color[HTML]{3166FF} 0.0}{\color[HTML]{FE0000} 0}{\color[HTML]{000000} 429}                      & {\color[HTML]{3166FF} 0.}{\color[HTML]{FE0000} 0}{\color[HTML]{000000} 6588}                      & {\color[HTML]{3166FF} 0.0}{\color[HTML]{000000} 2102}                     & {\color[HTML]{3166FF} 0.0}{\color[HTML]{000000} 4020}                  & {\color[HTML]{3166FF} 0.00}{\color[HTML]{000000} 429}                     & {\color[HTML]{3166FF} 0.0}{\color[HTML]{000000} 7219}                    & {\color[HTML]{3166FF} 0.0}{\color[HTML]{000000} 6809}                  & {\color[HTML]{3166FF} 0.}{\color[HTML]{000000} 26787}                 & {\color[HTML]{3166FF} 0.0}{\color[HTML]{000000} 1945}                     & {\color[HTML]{3166FF} 0.}{\color[HTML]{FE0000} 0}{\color[HTML]{000000} 1188}   \\ \hline
      FPR                                                & Model           & \rotatebox{90}{SVHN}                                                   & \rotatebox{90}{EuroSAT}                                                   & \rotatebox{90}{RESISC45}                                                   & \rotatebox{90}{Retinopathy}                                                   & \rotatebox{90}{FER2013}                                                   & \rotatebox{90}{UCF101}                                                   & \rotatebox{90}{GTSRB}                                                   & \rotatebox{90}{PatchCamelyon}                                                   & \rotatebox{90}{ImageNet}                                                   & \rotatebox{90}{Kinetics-700} \\ \hline
      \multicolumn{1}{l|}{\multirow{5}{*}{Recall0.9}}    & CLIP-ViT-B/16  & {\color[HTML]{3166FF} 0.}{\color[HTML]{000000} 31943}                 & {\color[HTML]{3166FF} 0.00}{\color[HTML]{000000} 500}                     & {\color[HTML]{3166FF} 0.00}{\color[HTML]{000000} 228}                      & {\color[HTML]{3166FF} 0.}{\color[HTML]{000000} 51317}                         & {\color[HTML]{3166FF} 0.}{\color[HTML]{000000} 24283}                     & {\color[HTML]{3166FF} 0.0}{\color[HTML]{000000} 2323}                    & {\color[HTML]{3166FF} 0.0}{\color[HTML]{000000} 5201}                   & {\color[HTML]{3166FF} 0.}{\color[HTML]{000000} 25607}                  & {\color[HTML]{3166FF} 0.00}{\color[HTML]{000000} 217}                        & {\color[HTML]{3166FF} 0.0}{\color[HTML]{000000} 6234}                         \\
      \multicolumn{1}{l|}{}                              & CLIP-R$50\times{16}$      & {\color[HTML]{3166FF} 0.}{\color[HTML]{000000} 16834}                & {\color[HTML]{3166FF} 0.00}{\color[HTML]{000000} 758}                     & {\color[HTML]{3166FF} 0.00}{\color[HTML]{000000} 361}                      & {\color[HTML]{3166FF} 0.}{\color[HTML]{000000} 53310}                         & {\color[HTML]{3166FF} 0.}{\color[HTML]{000000} 26278}                     & {\color[HTML]{3166FF} 0.0}{\color[HTML]{000000} 2141}                    & {\color[HTML]{3166FF} 0.0}{\color[HTML]{000000} 4958}                   & {\color[HTML]{3166FF} 0.}{\color[HTML]{000000} 24402}                  & {\color[HTML]{3166FF} 0.0}{\color[HTML]{000000} 6205}                        & {\color[HTML]{3166FF} 0.00}{\color[HTML]{000000} 185}                         \\
      \multicolumn{1}{l|}{}                              & ViT-L/16         & {\color[HTML]{3166FF} 0.}{\color[HTML]{000000} 26261}                & {\color[HTML]{3166FF} 0.00}{\color[HTML]{000000} 258}                     & {\color[HTML]{3166FF} 0.00}{\color[HTML]{000000} 753}                      & {\color[HTML]{3166FF} 0.}{\color[HTML]{000000} 61230}                          & {\color[HTML]{3166FF} 0.}{\color[HTML]{000000} 37351}                     & {\color[HTML]{3166FF} 0.0}{\color[HTML]{000000} 2632}                    & {\color[HTML]{3166FF} 0.0}{\color[HTML]{000000} 5752}                   & {\color[HTML]{3166FF} 0.}{\color[HTML]{000000} 23110}                  & {\color[HTML]{3166FF} 0.00}{\color[HTML]{000000} 140}                        & {\color[HTML]{3166FF} 0.0}{\color[HTML]{000000} 8693}                         \\
      \multicolumn{1}{l|}{}                              & Up-E MN-B15         & {\color[HTML]{3166FF} 0.}{\color[HTML]{000000} 32460}                 & {\color[HTML]{3166FF} 0.00}{\color[HTML]{000000} 189}                     & {\color[HTML]{3166FF} 0.00}{\color[HTML]{000000} 187}                      & {\color[HTML]{3166FF} 0.}{\color[HTML]{000000} 41944}                         & {\color[HTML]{3166FF} 0.}{\color[HTML]{000000} 28696}                     & {\color[HTML]{3166FF} 0.0}{\color[HTML]{000000} 1294}                    & {\color[HTML]{3166FF} 0.0}{\color[HTML]{000000} 3693}                   & {\color[HTML]{3166FF} 0.}{\color[HTML]{000000} 13974}                  & {\color[HTML]{3166FF} 0.00}{\color[HTML]{000000} 102}                        & {\color[HTML]{3166FF} 0.0}{\color[HTML]{000000} 7334}                         \\ \hline
      \multirow{4}{*}{Recall0.95}                        & CLIP-ViT-B/16  & {\color[HTML]{3166FF} 0.}{\color[HTML]{000000} 45813}                 & {\color[HTML]{3166FF} 0.0}{\color[HTML]{000000} 1215}                     & {\color[HTML]{3166FF} 0.00}{\color[HTML]{000000} 605}                      & {\color[HTML]{3166FF} 0.}{\color[HTML]{000000} 61820}                         &{\color[HTML]{3166FF} 0.}{\color[HTML]{000000} 35222}                     & {\color[HTML]{3166FF} 0.0}{\color[HTML]{000000} 4175}                    & {\color[HTML]{3166FF} 0.0}{\color[HTML]{000000} 8205}                   & {\color[HTML]{3166FF} 0.}{\color[HTML]{000000} 37364}                  & {\color[HTML]{3166FF} 0.00}{\color[HTML]{000000} 472}                        & {\color[HTML]{3166FF} 0.}{\color[HTML]{000000} 10373}                         \\
                                                        & CLIP-R$50\times{16}$      & {\color[HTML]{3166FF} 0.}{\color[HTML]{000000} 27431}                  & {\color[HTML]{3166FF} 0.0}{\color[HTML]{000000} 1652}                     & {\color[HTML]{3166FF} 0.00}{\color[HTML]{000000} 809}                      & {\color[HTML]{3166FF} 0.}{\color[HTML]{000000} 65823}                         & {\color[HTML]{3166FF} 0.}{\color[HTML]{000000} 39969}                     & {\color[HTML]{3166FF} 0.0}{\color[HTML]{000000} 3794}                    & {\color[HTML]{3166FF} 0.0}{\color[HTML]{000000} 8374}                   & {\color[HTML]{3166FF} 0.}{\color[HTML]{000000} 36910}                  & {\color[HTML]{3166FF} 0.}{\color[HTML]{000000} 10475}                        & {\color[HTML]{3166FF} 0.00}{\color[HTML]{000000} 420}                         \\
                                                        & ViT-L/16        & {\color[HTML]{3166FF} 0.}{\color[HTML]{000000} 38720}                  & {\color[HTML]{3166FF} 0.00}{\color[HTML]{000000} 725}                     & {\color[HTML]{3166FF} 0.0}{\color[HTML]{000000} 1842}                      & {\color[HTML]{3166FF} 0.}{\color[HTML]{000000} 69800}                           & {\color[HTML]{3166FF} 0.}{\color[HTML]{000000} 49426}                     & {\color[HTML]{3166FF} 0.0}{\color[HTML]{000000} 4340}                    & {\color[HTML]{3166FF} 0.0}{\color[HTML]{000000} 8229}                   & {\color[HTML]{3166FF} 0.}{\color[HTML]{000000} 36018}                  & {\color[HTML]{3166FF} 0.00}{\color[HTML]{000000} 392}                        & {\color[HTML]{3166FF} 0.}{\color[HTML]{000000} 14069}                         \\
                                                        & Up-E MN-B15         & {\color[HTML]{3166FF} 0.}{\color[HTML]{000000} 47035}                  & {\color[HTML]{3166FF} 0.00}{\color[HTML]{000000} 455}                     & {\color[HTML]{3166FF} 0.00}{\color[HTML]{000000} 563}                      & {\color[HTML]{3166FF} 0.}{\color[HTML]{000000} 53326}                         & {\color[HTML]{3166FF} 0.}{\color[HTML]{000000} 40968}                     & {\color[HTML]{3166FF} 0.0}{\color[HTML]{000000} 2173}                    & {\color[HTML]{3166FF} 0.0}{\color[HTML]{000000} 5897}                   & {\color[HTML]{3166FF} 0.}{\color[HTML]{000000} 23596}                  & {\color[HTML]{3166FF} 0.00}{\color[HTML]{000000} 282}                        & {\color[HTML]{3166FF} 0.}{\color[HTML]{000000} 12205}                         \\ \hline
      \multirow{4}{*}{Recall0.99}                        & CLIP-ViT-B/16   & {\color[HTML]{3166FF} 0.}{\color[HTML]{000000} 75319}                & {\color[HTML]{3166FF} 0.0}{\color[HTML]{000000} 5321}                     & {\color[HTML]{3166FF} 0.0}{\color[HTML]{000000} 3278}                      & {\color[HTML]{3166FF} 0.}{\color[HTML]{000000} 84480}                         & {\color[HTML]{3166FF} 0.}{\color[HTML]{000000} 59664}                     & {\color[HTML]{3166FF} 0.0}{\color[HTML]{000000} 6580}                     & {\color[HTML]{3166FF} 0.}{\color[HTML]{000000} 16495}                   & {\color[HTML]{3166FF} 0.}{\color[HTML]{000000} 61677}                  & {\color[HTML]{3166FF} 0.0}{\color[HTML]{000000} 2736}                         &{\color[HTML]{3166FF} 0.}{\color[HTML]{000000} 25118}                         \\
                                                        & CLIP-R$50\times{16}$   & {\color[HTML]{3166FF} 0.}{\color[HTML]{000000} 53917}                   & {\color[HTML]{3166FF} 0.0}{\color[HTML]{000000} 7207}                     & {\color[HTML]{3166FF} 0.0}{\color[HTML]{000000} 3309}                      & {\color[HTML]{3166FF} 0.}{\color[HTML]{000000} 88887}                         & {\color[HTML]{3166FF} 0.}{\color[HTML]{000000} 71641}                     & {\color[HTML]{3166FF} 0.0}{\color[HTML]{000000} 6091}                     & {\color[HTML]{3166FF} 0.}{\color[HTML]{000000} 17286}                   & {\color[HTML]{3166FF} 0.}{\color[HTML]{000000} 65826}                  & {\color[HTML]{3166FF} 0.}{\color[HTML]{000000} 25973}                        & {\color[HTML]{3166FF} 0.0}{\color[HTML]{000000} 2539}                          \\
                                                        & ViT-L/16       & {\color[HTML]{3166FF} 0.}{\color[HTML]{000000} 66286}                    & {\color[HTML]{3166FF} 0.0}{\color[HTML]{000000} 4337}                     & {\color[HTML]{3166FF} 0.0}{\color[HTML]{000000} 7651}                      & {\color[HTML]{3166FF} 0.}{\color[HTML]{000000} 91020}                          & {\color[HTML]{3166FF} 0.}{\color[HTML]{000000} 71809}                     & {\color[HTML]{3166FF} 0.0}{\color[HTML]{000000} 7007}                    & {\color[HTML]{3166FF} 0.}{\color[HTML]{000000} 15103}                   & {\color[HTML]{3166FF} 0.}{\color[HTML]{000000} 63046}                  & {\color[HTML]{3166FF} 0.0}{\color[HTML]{000000} 3487}                        & {\color[HTML]{3166FF} 0.}{\color[HTML]{000000} 31944}                         \\
                                                        & Up-E MN-B15       & {\color[HTML]{3166FF} 0.}{\color[HTML]{000000} 76834}                   & {\color[HTML]{3166FF} 0.0}{\color[HTML]{000000} 2723}                     & {\color[HTML]{3166FF} 0.0}{\color[HTML]{000000} 2246}                      & {\color[HTML]{3166FF} 0.}{\color[HTML]{000000} 78422}                         & {\color[HTML]{3166FF} 0.}{\color[HTML]{000000} 61911}                     & {\color[HTML]{3166FF} 0.0}{\color[HTML]{000000} 3631}                    & {\color[HTML]{3166FF} 0.}{\color[HTML]{000000} 13290}                   & {\color[HTML]{3166FF} 0.}{\color[HTML]{000000} 51347}                  & {\color[HTML]{3166FF} 0.0}{\color[HTML]{000000} 1143}                        & {\color[HTML]{3166FF} 0.}{\color[HTML]{000000} 30051}                         \\ \hline
      \end{tabular}
   }
   \caption{\label{tab:teng-fpr} FPR@Recall performance of 4 models on 20 classification datasets. In order to highlight the comparison results, we color the zeros with blue and highlight the advantages of our Up-E MN-B15 with the red color.}
\end{table*}

\begin{table*}
   \small
   \centering
   \begin{tabular}{l|ccc}
      \hline
      Models        & FPR@Recall0.9 & FPR@Recall0.95 & FPR@Recall0.99 \\ \hline
      Clip-ViT-B/16 & 1.64          & 1.38                               & 0.95           \\
      CLIP-R$50\times{16}$   & 1.59          & 1.33                               & 0.91           \\
      ViT-L/16        & 1.64          & 1.38                               & 0.93           \\
      Up-E MN-B15        & 2.19          & 1.90                               & 1.32           \\ \hline
      \end{tabular}
   \caption{Average FPR@Recall over 20 classification datasets.}
   \label{tab:teng-fpr-brief} 
\end{table*}

\begin{table*}
   \small
   \centering
    \begin{tabular}{l|ccc}
    \hline
    Models      & MR@FPPI0.01     & MR@FPPI0.1 & MR@FPPI1 \\ \hline
    R152  & 0.65          & 0.33     & 0.14   \\
    MoCo v2     & 0.67          & 0.35     & 0.14   \\
    CLIP-R$50\times{16}$ & 0.63 & 0.28     & 0.10   \\
    Up-E MN-B15 & 0.39   & 0.09        & 0.01      \\ \hline
    \end{tabular}
   \caption{\label{tab:teng-fppi} MR@FPPI on VOC07+12 object detection.}
\end{table*}

 \begin{figure*}
   \centering
    \includegraphics[width=1\linewidth]{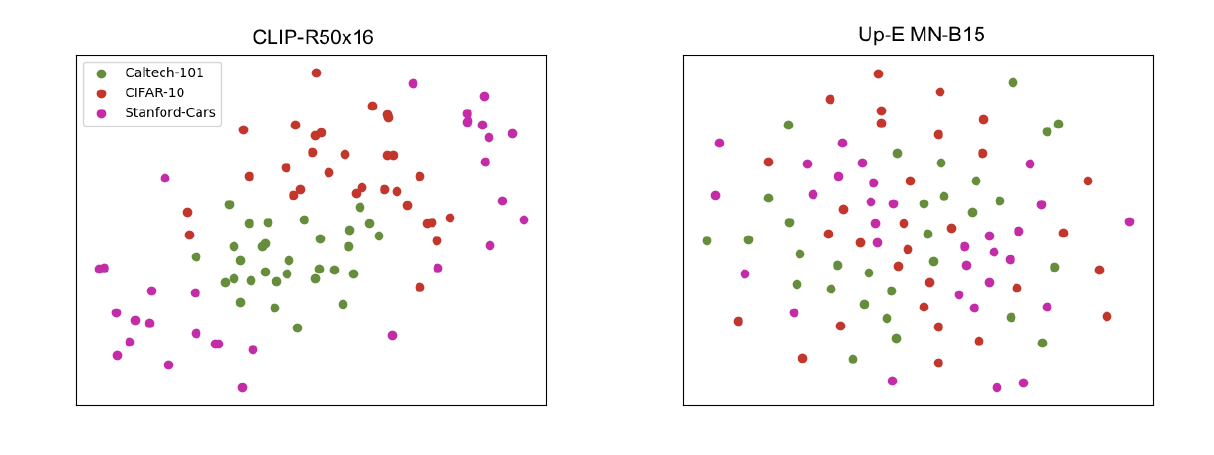}
    \caption{Visualization results of extract features with T-SNE. There are 35 samples categorized with \textit{car} for each dataset.}
    \label{fig:teng-tsne}
 \end{figure*}

\subsubsection{Comparison with Different Metrics}
For classification, we also provide results of FPR@Recall as another metric for model evaluation shown in Tab.~\ref{tab:teng-fpr}. We select 0.9, 0.95 and 0.99 as three thresholds of recall to evaluate the 20 classification datasets. In order to summarize, the negative logarithm with 10 based results of each number is calculated and averaged over 20 datasets shown in Tab.~\ref{tab:teng-fpr-brief}.
In the view of FPR@Recall, our Up-E MN-B15 surpasses CLIP-$50\times16$ by 30\% for all the three thresholds. Compared with traditional accuracy, the FPR@Recall with a high threshold can verify the confidence of a model's inference in classification tasks. The results prove our expert models' high stability achieved during the multi-stage pretraining. For detection, we provide MR@FPPI as another metric in the evaluation of VOC detection. The results are shown in Tab.~\ref{tab:teng-fppi}. The miss rate of CLIP-R$50\times16$ is double of our Up-E MN-B15 in Tab.~\ref{tab:teng-fppi} with FPPI of 0.01. The gap is further widened to ten times at the FPPI of 1.

\subsubsection{Visualization on Expert Robustness}

As an expert, it is required to deal with various data sources robustly and break through the barriers of different datasets. For the images sourced from other datasets while categorized into the same class, \eg, cars in Stanford Cars, CIFAR-10 and Caltech-101, we show that our INTERN expert models meet the goal of possessing sufficient robustness. 
We select 35 samples with the annotation of \textit{car} from Stanford Cars, CIFAR-10 and Caltech-101 respectively. Then we visualize the extracted features with T-SNE.
The comparison between CLIP-R$50\times16$ and Up-E MN-B15 are provided in Fig.~\ref{fig:teng-tsne}. We observe that the features extracted by our Up-E MN-B15 are more evenly dispersed than CLIP-R$50\times16$, the features from whom are obviously clustered around the different dataset sources. We attribute the visualization results to the robustness learned from the training process of the expert model, consisting of rich data sources corresponding to the expert task.

\subsubsection{Robust Representation for Generalizing to Unseen Tasks}

We aim to train our Up-G models to be proficient at all known tasks and have a fast adaptation ability to unseen tasks with few samples. As displayed in Tab.~\ref{tab:whole_results}, our Up-G models achieve better performance than most other pretrained models in the two unseen tasks, VOC segmentation and depth estimation, even in the 10\% vs. 100\% comparison. Take Up-G MN-B15 as an example, Up-G MN-B15 achieves 89.8\% mIoU on VOC segmentation and 2.71 RMSE on KITTI depth estimation using 10\% train data, surpasses CLIP-R$50\times16$ using 100\% train data with +6.2\% on VOC and +0.12\% on KITTI respectively. We note that any of the single expert model fails to achieve the generalizability, Up-E (C) and Up-E (D), for instance, lags behind the Up-G model in terms of VOC segmentation and KITTI depth estimation. Our pipeline effectively exploits information embedded in multiple tasks, leading to a more robust and generalizable model.

\section{Discussion}
\label{sec:Discussion}
Our new learning paradigm, INTERN, aims to systematically solve bottlenecks in the current field of general vision intelligence.
To this end, three fundamental bases of INTERN are proposed, \ie General Vision Dataset (GV-D), General Vision Architecture (GV-A) and General Vision Benchmark (GV-B), all of which together construct a general vision ecosystem. This ecosystem provides a platform for researchers and developers to study general vision openly and inclusively, significantly promoting research in the field of general vision.
Besides, our extensible upstream pretraining scheme, which is composed of three stages, \ie, Upstream-Amateur (Up-A), Upstream-Expert (Up-E), and Upstream-Generalist (Up-G), together with the flexible Downstream-Adaptation (Down-A) scheme, proves to be an effective learning pipeline to ease the training of the desired General Vision Model. 
Moreover, we are glad to see, facilitated with this learning pipeline, our trained model outperforms the previous state-of-the-art (CLIP~\cite{CLIP}), only using an order of magnitude fewer (10\%) downstream data.
Apart from the high performance of our models, we demonstrate that INTERN possesses easy extensibility with more tasks and excellent generalizability when facing unseen tasks.

Due to its ability to perform arbitrary scenario-agnostic vision tasks with fewer data, INTERN would have broad impacts on several aspects, including but not limited to the AI community, the industry and society. We analyze from different dimensions and list some of the potential impacts as follows. For the AI community, our multi-stage pretraining scheme, which is consistent with the natural process of human cognition, would provide a new perspective for developing more advanced and efficient pretraining options of General Vision Model. Other than that, our release of the general vision ecosystem will lower the threshold for researchers interested in this area. For example, the General Vision Benchmark (GV-B) can help examine the generalizability of their developed vision models in a relatively fair and comprehensive setting. 
For the AI industry, the AI companies typically over-specialize on a single task to develop AI applications, where each new AI application requires collecting and annotating a large amount of data and consumes a giant scale of computational resources. Thanks to the generality of our INTERN, AI companies can only focus on adapting pretrained models to some specific tasks, which not only reduces the training costs but also ensures the high precision of their models. For instance, in autonomous driving, our pretrained models can be integrated to help address open-world class recognition, further avoiding accidents due to the false recognition predicted by the autonomous driving system.
What is more, our INTERN would protect our environment to some extent for society. Specifically, the well-pretrained general vision models in INTERN can effectively transfer to various downstream tasks with significantly fewer data and annotations, which will substantially reduce the training consumption of downstream tasks, which in turn reduces carbon emissions significantly.

Undoubtedly, INTERN still has some limitations, and we will also improve it in future works.
In GV-A, we automatically search for the optimal combination of different operators; the operator design still relies on the human experience.
Some recent work~\cite{real2020automl} proposes to search for the basic operators from scratch. Building such a huge search space and using an efficient search method~\cite{fnas} may be a feasible way to find a better hybrid architecture.
In GV-B, the tasks we collect mainly focus on mainstream perception tasks. On the one hand, the types of perception tasks are limited to some widely-used image-based ones. Some video-based perception tasks, \eg action localization, as well as some low-level tasks, \eg image super-resolution and image inpainting are not included in the GV-B. On the other hand, only considering perception tasks cannot be the long-term goal for a complete general vision pretraining. Tasks for evaluating the model's cognitive and reasoning abilities should be included in future work. Moreover, the similarities and differences between general high-level and general low-level vision are worth exploring.

In the pretraining scheme, the generality of the Up-A stage on instance level tasks is a more concerning question. For learning vision-language representation, Up-A-G trained by global contrastive learning with rich supervisory signals and Up-A-L trained by weak instance contrastive learning have achieved performance improvements in classification and detection tasks. However, for more dense and fine-grained tasks, the Up-A stage still requires many attempts, such as pixel-level contrastive signals and fine-grained loss functions. 
In the Up-E stage, we introduce three types of experts and yield exciting transfer performance to various downstream tasks, as shown in Tab.~\ref{tab:consolidation}. It is promising to add more experts such as generative experts like GAN, a multi-modality expert like image captioning and visual question answering, as well as prediction experts like motion prediction in our framework to facilitate more generalizable visual representations. 

Nevertheless, our INTERN still faces some potential risks. Regarding the versatile abilities and availability of INTERN, others may leverage them in surveillance applications and military use, in which INTERN has the potential to be misused. We made strict protocols for global researchers and engineers to employ our work safely and lawfully. 
Besides, the images in GV-D may involve privacy issues, such as license plate numbers, faces, and documents. How to protect the privacy security under general vision will become an important issue. We will also explore federal learning or other methods to exploit our collected data without compromising privacy.

\section{Related Work}
\label{sec:related_work}

\noindent\textbf{General Vision Model Design.}
A host of ConvNets have been proposed to push forward the state-of-the-art computer vision approaches such as \cite{resnest,tan2019efficientnet}. Despite the numerous CNN models, their basic operations, convolution, are the same. Recently, \cite{dosovitskiy2020image} proposed a pure Transformer based image classification model Visual Transformer (ViT), which achieves impressive performance on ImageNet benchmark. DeiT \cite{deit} reveals that well-trained ViT can obtain a better performance-speed trade-off than ConvNets. Visual Transformer (VT) decouples semantic concepts of the input image into different patches and relates them densely through Transformer encoder blocks. BoTNet \cite{botnet} proposes a conceptual redefinition that the successive bottleneck blocks with self-attention mechanism can be regarded as Bottleneck Transformer blocks, although the short-connection form is different. With the same operations, CeiT \cite{yuan2021incorporating} also redesigns a patch-to-tokens scheme and attaches a Layer-wise Class token Attention (LCA) at the top of the Transformer to aggregate multi-level representations. Based on the equivalence between point-wise convolution and position-wise FFN, LocalViT \cite{localvit} extends this convolutional version FFN to an inverted residual block to build a depth-wise convolutional framework. To generalize variable-resolution inputs using the inherent positional information of convolution, ResT~\cite{rest} assumes that there is a correlation between positional encodings and inputs. Following the observation that the borders of convolution with zero paddings can encode the absolute positional information, CPVT \cite{cpvt} re-places positional encoding with a series of convolutions. CvT~\cite{wu2021cvt} propose to incorporate self-attention and convolution by generating $\mathtt{Q}$, $\mathtt{K}$, and $\mathtt{V}$ in self-attention with convolution. ConViT~\cite{d2021convit} tries to unify convolution and self-attention with gated positional self-attention and it is more sample-efficient than self-attention. 
Previous works use a different form to combine those operators and get promising results, but it requires tedious manual design, lacking effective guidelines. Similar to Network In Network (NIN) \cite{nin} series, Han et al. \cite{tnt} leverage a Transformer-iN-Transformer (TNT) model to aggregate both patch-and pixel-level representations. Similar to a 2D TSM, Liu et al. \cite{liu2021swin} present a Shifted windows (Swin) Transformer that utilizes a shifted window along the spatial dimension to model global and boundary features. As a local-global separate Transformer, Twins \cite{chu2021twins} replaces the complicated design of Swin Transformer with a Spatially Separable Self-Attention mechanism (SSSA). UniNet \cite{liu2021uninet} proposes to jointly search the combination of General Operators (GOP) and the DownSampling Module (DSM), and achieves a better computation-accuracy trade-off.

Vision Outlooker (VOLO)~\cite{yuan2021volo} uses outlook attention to focus on finer-level features, instead of other attention-based modules. The paradigm of hierarchical Transformer is first introduced by Tokens-to-Token ViT (T2T-ViT) \cite{t2t}. In T2T-ViT, a layer-wise T2T transformation is used to aggregate neighboring tokens into one single token. Similar to the shrinking strategy of PVT \cite{cpvt}, Pooling-based Vision Transformer (PiT) \cite{pit} and Convolutional vision Transformer (CvT) \cite{wu2021cvt} utilize pooling and convolution to perform token embedding, respectively. From the structure aspect, CaiT \cite{cait} present efficient class-attention in image Transformers, including two stages. 1) Multiple self-attention stages without class token. In each layer, a learned diagonal matrix initialized by small values is exploited to update the channel weights dynamically, thereby offering a degree of freedom for channel adjustment. 2) Last few class-attention stages with frozen patch embeddings. Deep Transformer suffers from attention collapse and over-smoothing problems, but still largely preserves the diversity of attention map between different heads. Based on this observation, Zhou et al. \cite{deepvit} propose Deep Vision Transformer (DeepViT) that aggregates cross-head attention maps and re-generates a new one by using a linear layer to increase cross-layer feature diversity. 

\noindent\textbf{Large-scale Multi-modal Pretraining.}
Contrastive learning, as a pretext task of self-supervised training, has achieved remarkable success in visual representation learning. MoCo~\cite{he2020momentum} achieves contrastive-learning based training by building a dynamic dictionary with a queue and a moving-averaged encoder. SimCLR~\cite{chen2020simple} selects the optimal data augmentation combination method, adds a non-linear mapping after the encoder, discards the memory bank and adopts a larger batch size to calculate the contrastive learning loss. MoCo v2~\cite{chen2020improved} draws on the idea of SimCLR~\cite{chen2020simple} and improves on the basis of MoCo~\cite{he2020momentum}. BYOL~\cite{grill2020bootstrap} adds the predictor module to the projector, and learns the mapping from online encoder to target encoder through predictor, only considering the positive sample pairs. Besides, it uses the target network stop-gradient for training. SimSiam~\cite{chen2021exploring} removes the target encoder updated by momentum on the basis of BYOL~\cite{grill2020bootstrap}. In addition, with the popularity of vision Transformer, MoCo v3~\cite{chen2021empirical} applies contrastive learning to the Vision Transformer network, and analyzes the characteristics of the Vision Transformer structure when training a contrastive learning model. DINO~\cite{caron2021emerging} designs an unlabeled self-supervised method with knowledge distillation by analyzing the characteristics of ViT~\cite{dosovitskiy2020image} features. In addition to the good performance, the learned features in DINO have strong interpretability.

Multi-modal learning has made significant progress on image-text tasks recently, especially on large scale pretraining tasks, with amazing results. Their paradigm is mainly to use two encoders to respectively encode image and text information, and use contrastive learning for training. CLIP~\cite{CLIP} is a milestone work. It uses 400 million image-text data crawled on the Internet for training, and has achieved remarkable results in image-text retrieval tasks, zero-shot image recognition tasks and pretraining tasks. ALIGN~\cite{jia2021scaling} scales up the image-text data to 1 billion on the basis of CLIP~\cite{CLIP}. During training, the text encoder generates the label weights and uses a larger batchsize to achieve better results than CLIP~\cite{CLIP}. WenLan~\cite{huo2021wenlan} and EfficientCLIP~\cite{wang2021efficientclip} train Chinese pairs data. EfficientCLIP~\cite{wang2021efficientclip} uses Ensemble Confident Learning to make good use of single-modal text data during multi-modal training. With fewer training resources, it achieves the state-of-the-art performance on Chinese cross-modal retrieval tasks. CLIP2Video~\cite{fang2021clip2video} is based on the semantic space constructed by the CLIP~\cite{CLIP} model to capture motions at fine temporal video frames, re-align the tokens of video clips and phrases and enhance the multi-modal correlation to transfer the image-language pretraining model to video-text retrieval, achieving state-of-the-art performance on major text-to-video and video-to-text retrieval benchmarks.

\noindent\textbf{Classification.} In classification, 
\cite{bilen2017universal,rebuffi2017learning,guo2019depthwise,li2021representation,guo2019spottune,simpson2008refining,strezoski2019many,wang2021multi,xiao2020multi,newell2019feature,liang2018dynamicseg,lambert2020mseg} introduce architectures to handle multiple tasks using data from multiple domains. Among them, \cite{bilen2017universal} proposes to absorb different domains in a single neural network by tuning certain parameters in batch and instance normalization layers throughout the architecture, however, \cite{bilen2017universal} does not study the learned representation on downstream tasks.
Similarly, \cite{rebuffi2017learning} introduces multivalent neural network architectures for multiple-domain learning. This method performs well on ten different visual classification problems. \cite{guo2019depthwise} resembles these works and makes use of depthwise separable convolutions in order to build more efficient multi-task networks. Recently, \cite{li2021representation} consolidates knowledge from various task-specific teacher models into a single student model. Our \stagetwo{} is similar to these works, on the difference that our \expertcls{} model is trained at scale on eleven datasets from ImageNet-21K\cite{deng2009imagenet}, iNat2021\cite{van2021benchmarking} to Places365\cite{zhou2017places}, which is one of the largest scale settings by far, to the best of our knowledge. Our model is also trained on the proposed \datanamecls{} dataset, the number of label of the dataset reaches 115K, which is by the time the largest taxonomy in classification datasets.

\noindent\textbf{Object Detection.} There are many works \cite{zhao2020object,Towards_universal,Simple_multi_dataset_detection} that manage to train a unified model on multiple object detection tasks. \cite{zhao2020object} manually merges the taxonomies of 3 datasets: COCO, Pascal VOC and SUN-RGBD and train with pseudo-labels generated by dataset specific networks. This work takes much effort on unifying taxonomies and is not feasible when adding additional dataset.
\cite{Towards_universal} train a universal object detector capable of operating over multiple domains, and gain robustness by joining diverse sources of supervision.
\cite{Simple_multi_dataset_detection} built a unified object detector on 3 datasets: COCO, Objects365, and OpenImages and observes that a single detector performs as well as dataset-specific models on each individual dataset and generalizes better to new unseen domains. This work is similar to ours, but we propose a novel object detection dataset GV-D$_\text{d}$-3M and validate the unified detector not only on detection task in new domain but also other types of downstream tasks, \ie classification and segmentation as well as depth estimation. 

\noindent\textbf{Semantic Segmentation.}
 \cite{bevandic2021multi,liang2018dynamicseg,lambert2020mseg,shi2021multi} try to train a single semantic segmentation model on various datasets. \cite{liang2018dynamicseg} explicitly integrates a semantic concept hierarchy into the dynamic network optimization and trains four datasets within a unified segmentation model, which shows the superior performance on each dataset involved.
 \cite{lambert2020mseg} manually unifies the taxonomies of seven semantic segmentation datasets from different domains and produces compatible annotations across datasets by relabeling object masks by heavy manual efforts.
 \cite{shi2021multi} proposes a multi-dataset pretraining framework for semantic segmentation, making use of the off-the-shelf annotations to construct general pretrained models. Our work resembles these works, on the difference that we evaluate transfer performance on a wide range of tasks including classification, object detection and pixel-wise prediction tasks, including semantic segmentation and depth estimation. 
 
\noindent\textbf{Feature Fusion.}
In Sec.~\ref{sec:stage2}, \expert{} models can be regarded as an encoder sharing hard parameters, which branches to a specific task header. Different from \expert{}, in \generalist{}, each task has reached its own optimization, and the characteristics need to interact between processing tasks in a specific sharing mechanism. This mechanism is similar to the soft parameter sharing in multi-task learning described in cross stitch network~\cite{misra2016cross}, which uses a linear combination of the activations in every layer of the task-specific networks as a means for soft feature fusion. For instance, NDDR-CNN~\cite{gao2019nddr} adopts dimension reduction method for feature fusion between multi-task learning among multiple branches. In the process of feature fusion, MTAN~\cite{liu2019end} uses an attention mechanism to extract the shared feature branch. Simon~\cite{vandenhende2019branched}  leverages the employed tasks' affinities to construct branched multi-task networks. It lets the deeper network gradually grow more task-specific in a given budget. FAIR~\cite{purushwalkam2019task} proposes a gating function conditioned on the task to produce features, which represent the compatibility between the input image and the concept under consideration. 

\noindent\textbf{Transfer Learning.}
Transfer learning aims at improving performance on target domains by transferring the knowledge from source domains or source foundation models. To achieve this, some of the recent top-performance transfer learning methods rely on the modification of pretrained model's architecture~\cite{ramachandran2016unsupervised,houlsby2019parameter}, some try to find important features and maintain them or only update part of weights more or less heuristically~\cite{kirkpatrick2017overcoming,chronopoulou2019embarrassingly,howard2018universal,guo2019spottune,guo2020adafilter,ro2021autolr}, and others try to get more supervision signals~\cite{phang2018sentence,yogatama2019learning} which rely on human design for different downstream tasks. All these methods indeed obtained prominent improvements, but their universality is not satisfactory.

Another term commonly used in the transfer learning area is domain adaptation.
Prior attempts in domain adaptation focus on finding an effective distribution matching strategy with the knowledge from source domain and target domain, which can be defined as closed set domain adaptation~\cite{long2015learning,courty2016optimal} or partial domain adaptation\cite{cao2018partial, zhang2018importance}. Compared with those approaches, our method has no dependency on source domain data. Works in \cite{kundu2020universal, feng2020kd3a, ahmed2021unsupervised, liang2020we} discuss the problem of source domain free adaptation which also assumes no prior knowledge on the source domain. However, these works do not give consideration to a unified transfer pipeline, thus they are hard to be applied to industrial production compared with our Down-A.

\section{Conclusion}
\label{sec:conclusion}

INTERN proves to be an effective paradigm towards general vision intelligence. We hope our proposed INTERN will serve as a firm foundation for further general vision studies in the community.
To the industry and society, strongly generalizable models from INTERN are expected to dramatically reduce data demands, thus facilitating applications of AI technologies.
That being said, there are still many opportunities for improvement in the future. It is possible to incorporate new modalities, including but not limited to videos and audio. The final generalist model could potentially generalize to a far broader range of visual tasks beyond currently considered well-solved ones. We are also interested in boosting the data efficiency of upstream pretraining, making the whole learning process more economical and affordable.
Moreover, future researches may leverage learned strong visual representation to approach more complex cognitive understanding tasks.

\section*{Acknowledgements}

We would like to thank the entire SenseTime infrastructure and HPC teams for providing the service to train large-scale models. We would also like to thank \textbf{Ziwei Liu} for his thoughtful discussion on the design of the generalist pretraining stage, \textbf{Yuanhan Zhang} for discussion in the label system and support in main experiments for dataset creation. We are also grateful to \textbf{Yichun Zhou and Zexin He} for their support in the active learning method and experiments for annotation pipeline construction. During constructing the label system, \textbf{Mukai Li and Yubo Ma} gave useful guidance by providing some widely-used knowledge graph. \textbf{Qian Liu, Meihong Sun, Xiaoyu Yu} took responsibility for data annotation and quality control with the human annotation team. \textbf{Xiao Luo, Baohong Lv and Xianshun Ren} crawled raw data from the Flickr platform. \textbf{Ming Yi} provided sufficient storage space and stable storage I/O.
As for the experiments part, We'd like to express our gratitude to \textbf{Weichao Luo} for providing the codebase for GPU memory optimization, such as Half-precision floating point format (FP16) and Zero Redundancy Optimizer (ZeRO), \textbf{Zhengyang Liang} for supporting the codebase for face detection evaluation and the models evaluation in the benchmark. We also appreciate the contribution of \textbf{Bei Gan and Dongyang Liu} for helping the codebase construction for benchmark evaluation, \textbf{Junqin Huang} and \textbf{Yubo Fu} for implementing transfer learning algorithms and performing experiments on multiple downstream tasks. Besides, \textbf{Lichen Zhao and Feng Liang} gave great supports in the multi-modal pretraining stage. Many thanks are also given to the explorations of \textbf{Yufeng Cui and Zexin He} on hyperparameter-searching for benchmark settings.

Finally, we would like to thank \textbf{Lu Sheng, Wei Li, Yi Wang, Xianzheng Ma and Hongjie Zhang} for finalizing the drafts of the paper. \textbf{Huipeng Deng} gave help in correcting grammar mistakes. And \textbf{Xianzheng Ma} took charge of reviewing the whole paper.

The project is supported by the Shanghai Committee of Science and Technology, China (Grant No. 21DZ1100100 and 20DZ1100800).

\clearpage

\section*{Contributions}

\textbf{Qinghong Sun, Yuming Huang and Zhenfei Yin} constructed the \emph{General Vision Data}, including crawling and cleaning strategy design, label system implementation and systematically studied the effect of large-scale data.

\textbf{Jihao Liu, Shuo Qin, Guanglu Song and Yu Liu} searched the neural architectures and proposed the MetaNet family for \emph{General Vision Architecture}.

\textbf{Jianing Teng and Chengyu Wang} implemented the \emph{General Vision Benchmark}, including collecting downstream datasets and conducting transfer learning experiments. 

\textbf{Fenggang Liu, Huan Peng and Yangguang Li} developed \emph{Upstream-Amateur} and pre-processed the 10B data for efficient contrastive learning.

\textbf{Siyu Chen, Gengshi Huang and Kun Wang} implemented the pre-training for classification, object detection and semantic segmentation in \emph{Upstream-Expert} stage.

\textbf{Yinan He and Kun Wang} improved the fusion strategy of generalist and conducted the massive experiments.

\textbf{Mengya Gao, Yujie Wang, Yichao Wu and Ding Liang} developed the downstream adaptation approach and led the analysis of transfer learning.

\textbf{Conghui He} organized the workers for labelling and kept a quality of annotation.

\textbf{Fengwei Yu} led the team for providing computing service and implementation optimization for training large-scale models.

\textbf{Jing Shao and Yu Qiao} overall designed and led the research, and were responsible for research progress.

\textbf{Junjie Yan, Dahua Lin and Xiaogang Wang} thought deeply about the general vision model and suggested the key research route.

{\small
\bibliographystyle{ieee_fullname}
\bibliography{egbib}
}

\end{document}